\documentclass[table]{ai2style/ai2}
\usepackage{amssymb}
\usepackage{multirow}
\usepackage{bigdelim}
\usepackage{todonotes}
\usepackage{longtable}
\usepackage{tabularray}
\usepackage{wrapfig}
\usepackage[most]{tcolorbox} 
\usepackage{url}
\usepackage{xspace}
\usepackage{svg}
\usepackage[absolute]{textpos} %

\usepackage{fdsymbol}   %

\usepackage[utf8]{inputenc} %
\usepackage[T1]{fontenc}    %
\usepackage{url}            %
\usepackage{booktabs}       %
\usepackage{amsfonts}       %
\usepackage{nicefrac}       %
\usepackage{microtype}      %
\usepackage[table]{xcolor}         %
\usepackage{amsmath}
\usepackage[most]{tcolorbox}
\usepackage{csquotes}

\usepackage{siunitx}
\usepackage{graphicx}
\usepackage{arydshln}
\usepackage{wrapfig}
\usepackage{enumitem}
\usepackage{soul} %

\usepackage{multirow}
\usepackage{xspace}
\usepackage{adjustbox}
\usepackage{pifont}
\usepackage{caption}
\usepackage{makecell}
\usepackage{subcaption}
\usepackage{bold-extra}
\usepackage{url}

\usepackage{hyperref}
\definecolor{linkcolor}{RGB}{0, 0, 128}
\hypersetup{
     colorlinks   = true,
     citecolor    = linkcolor,
     linkcolor    = linkcolor,
     urlcolor     = linkcolor,
}
\usepackage{pifont}%
\newcommand{\cmark}{\ding{51}}%
\newcommand{\xmark}{\ding{55}}%
\usepackage{listings}

\setlist[itemize]{leftmargin=*,itemsep=0em,parsep=0.3em,topsep=0.3em}

\definecolor{maroon}{HTML}{F26035}
\definecolor{yellow}{HTML}{FDBC42}
\definecolor{lavender}{HTML}{734f96}
\definecolor{darkergrey}{HTML}{444444}
\definecolor{midgrey}{HTML}{e6eded}
\definecolor{ai2pink}{HTML}{f0529c}%
\definecolor{ai2midpink}{HTML}{fad3e5}
\definecolor{ai2lightpink}{HTML}{fbecf3}
\definecolor{ai2midwhite}{HTML}{f2e5d9}
\definecolor{ai2offwhite}{HTML}{fbf4ee}
\definecolor{ai2green}{HTML}{0fcb8c}
\definecolor{ai2lightgreen}{HTML}{e7f9f3}
\definecolor{ai2darkgreen}{HTML}{105257}
\definecolor{ai2purple}{HTML}{B932EB}
\definecolor{ai2lightpurple}{HTML}{f7e8fc}
\definecolor{neutralEight}{HTML}{343434}
\definecolor{neutralFive}{HTML}{838383}
\definecolor{neutralThree}{HTML}{bebebe}
\definecolor{neutralOne}{HTML}{dedede}
\definecolor{lightgrey}{HTML}{fafcfc}

\usepackage{tikz}
\newcommand{\cblock}[3]{
  \hspace{-1.5mm}
  \begin{tikzpicture}
    [
    node/.style={square, minimum size=10mm, thick, line width=0pt},
    ]
    \node[fill={rgb,255:red,#1;green,#2;blue,#3}] () [] {};
  \end{tikzpicture}%
}

\newcommand{\norm}[1]{\left\lVert#1\right\rVert}

\definecolor{maroon}{HTML}{F26035}
\definecolor{yellow}{HTML}{FDBC42}
\definecolor{darkred}{RGB}{156, 39, 33}
\definecolor{darkblue}{RGB}{31, 90, 153}
\definecolor{forestgreen}{rgb}{0.13, 0.55, 0.13}
\definecolor{olmoDarkBlue}{HTML}{012e59}
\definecolor{olmoBlue}{HTML}{265ed4}
\definecolor{olmoLightBlue}{HTML}{012e59}
\definecolor{olmoTeal}{HTML}{00d5ff}
\definecolor{olmoYellow}{HTML}{ffbb00}
\definecolor{olmoOrange}{HTML}{ff9100}

\newcommand{\tulu}{T\"ulu 3\xspace}
\newcommand{\olmo}{\textsc{OLMo}\xspace}
\newcommand{\olmozero}{\textsc{OLMo 1}\xspace}
\newcommand{\olmoapril}{\textsc{OLMo-0424}\xspace}
\newcommand{\olmotoo}{\textsc{OLMo 2}\xspace}
\newcommand{\olmotooinstruct}{\textsc{OLMo 2-Instruct}\xspace}
\newcommand{\dolma}{\textsc{Dolma}\xspace}

\newcommand{\olmomix}{\textsc{OLMo 2 Mix 1124}\xspace}
\newcommand{\dolminos}{\textsc{Dolmino Mix 1124}\xspace}

\newcommand{\diveStability}{Pretraining Stability}
\newcommand{\diveAnnealing}{Mid-training Recipe}
\newcommand{\diveInfra}{Infrastructure as a Research Catalyst}
\newcommand{\divePost}{Post-training Pipeline}

\newcommand{\huggingface}{\raisebox{-1.5pt}{\includegraphics[height=1.05em]{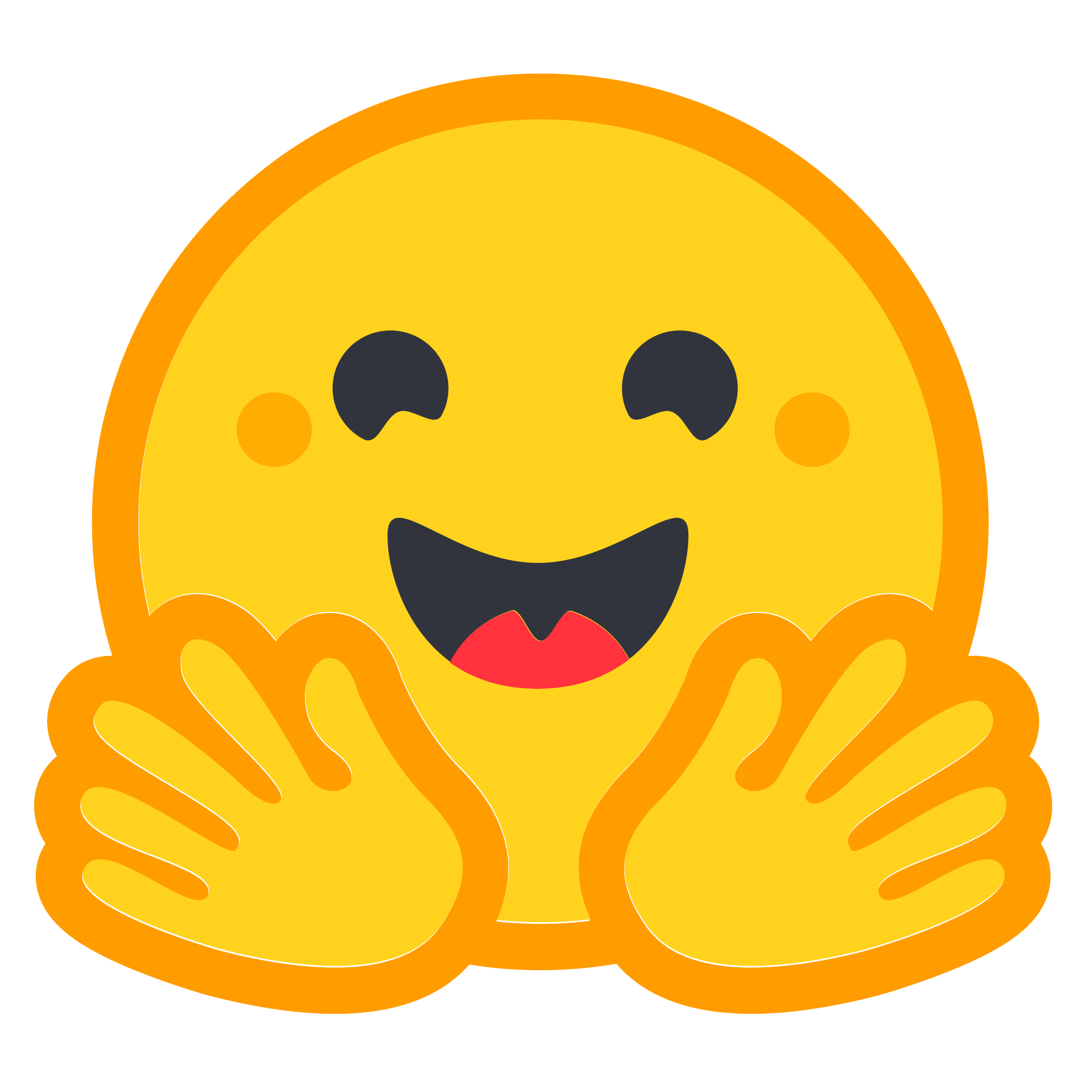}}\xspace}
\newcommand{\emailLogo}{\raisebox{-1.5pt}{\includegraphics[height=1.05em]{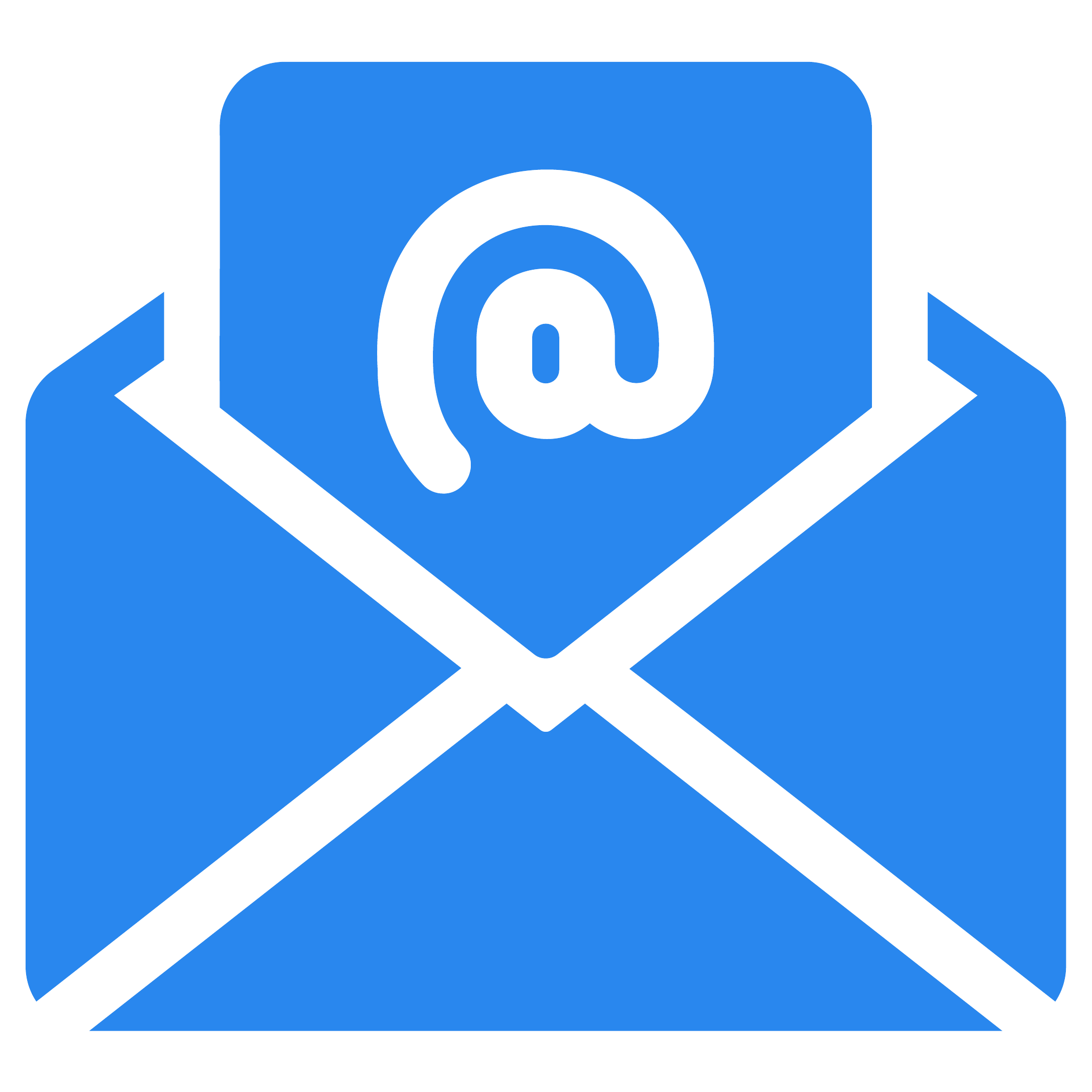}}\xspace}
\newcommand{\hfdataset}{\raisebox{-1.5pt}{\includegraphics[height=1.05em]{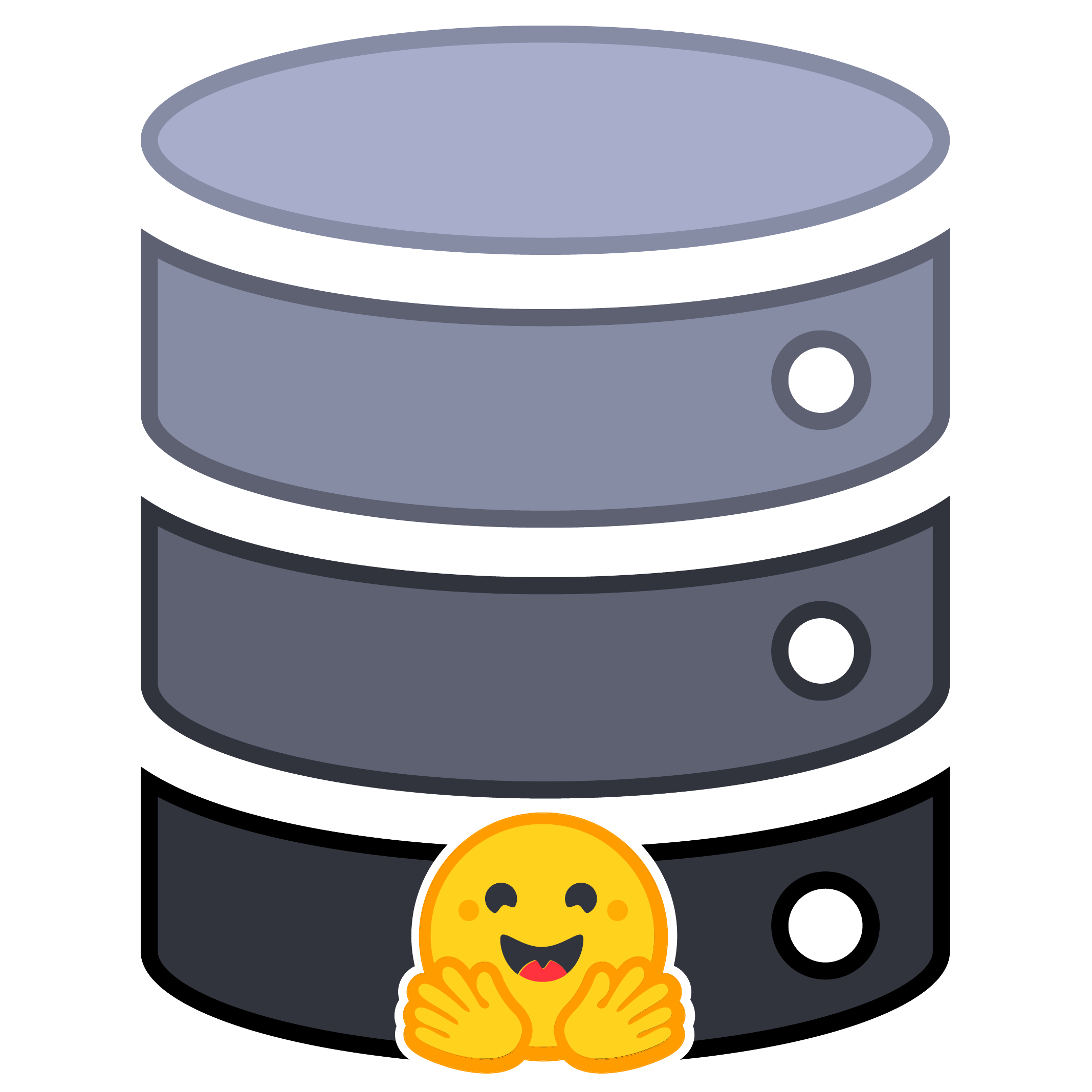}}\xspace}
\newcommand{\github}{\raisebox{-1.5pt}{\includegraphics[height=1.05em]{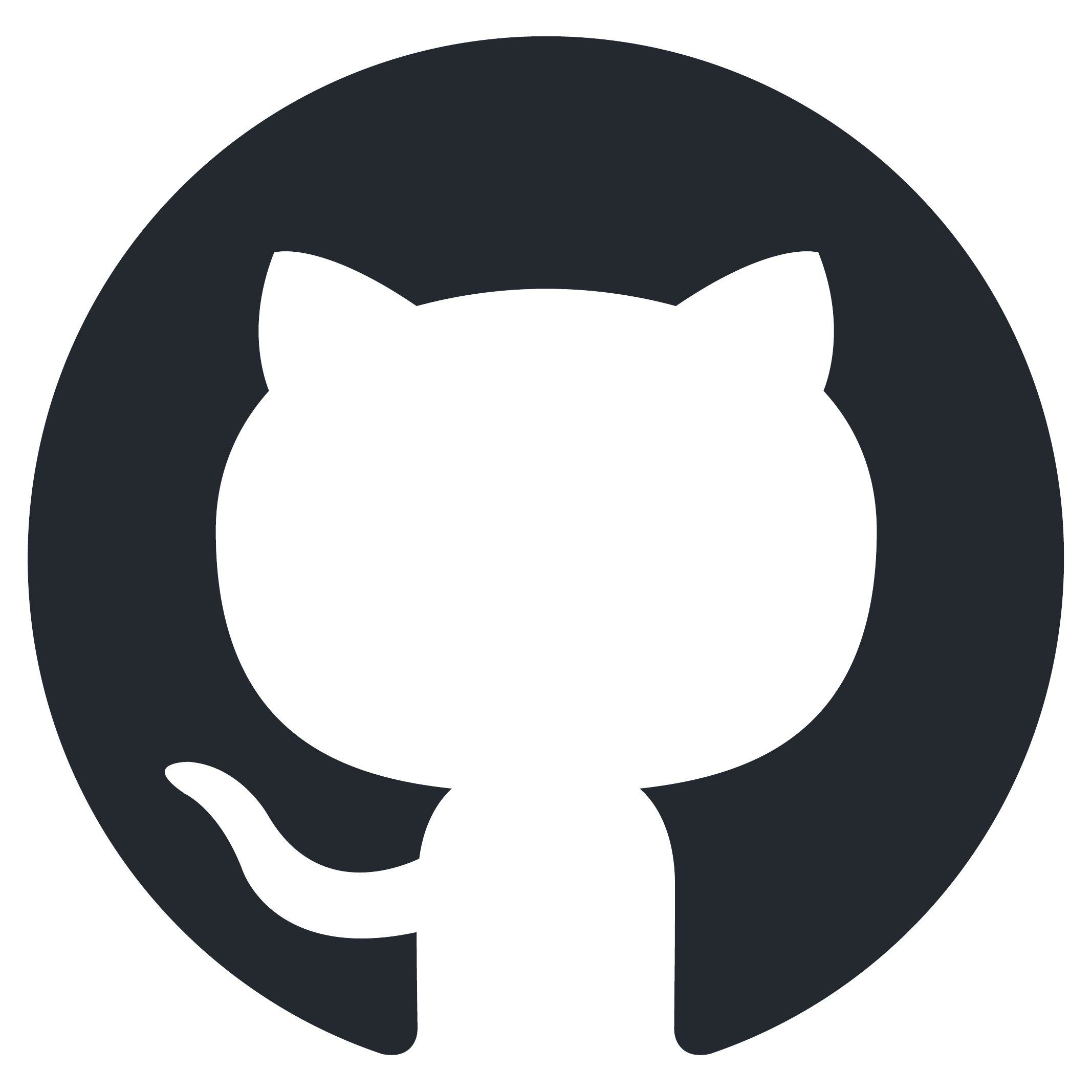}}\xspace}
\newcommand{\aitoo}{\raisebox{-1.5pt}{\includegraphics[height=1.05em]{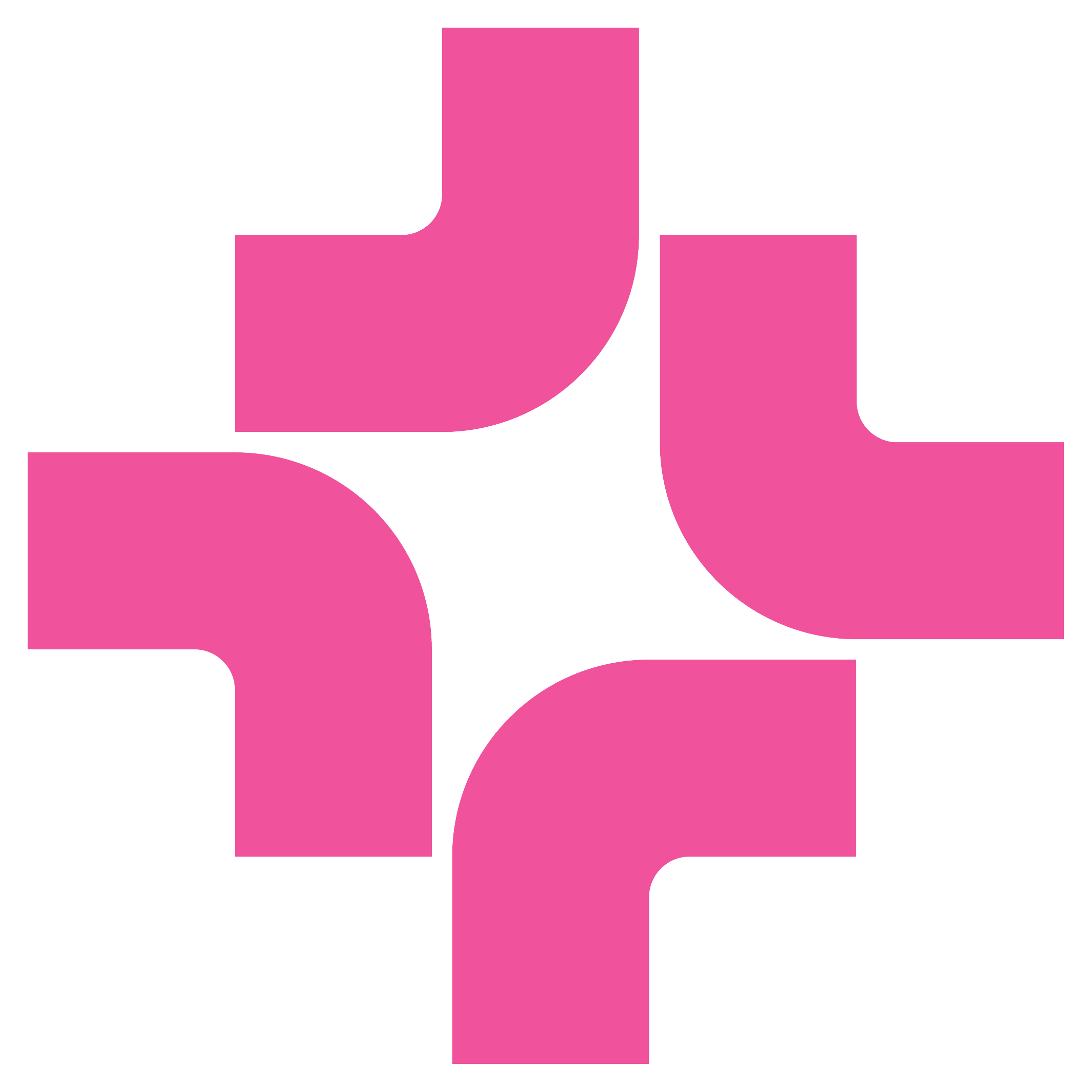}}\xspace}
\newcommand{\wandb}{\raisebox{-1.5pt}{\includegraphics[height=1.05em]{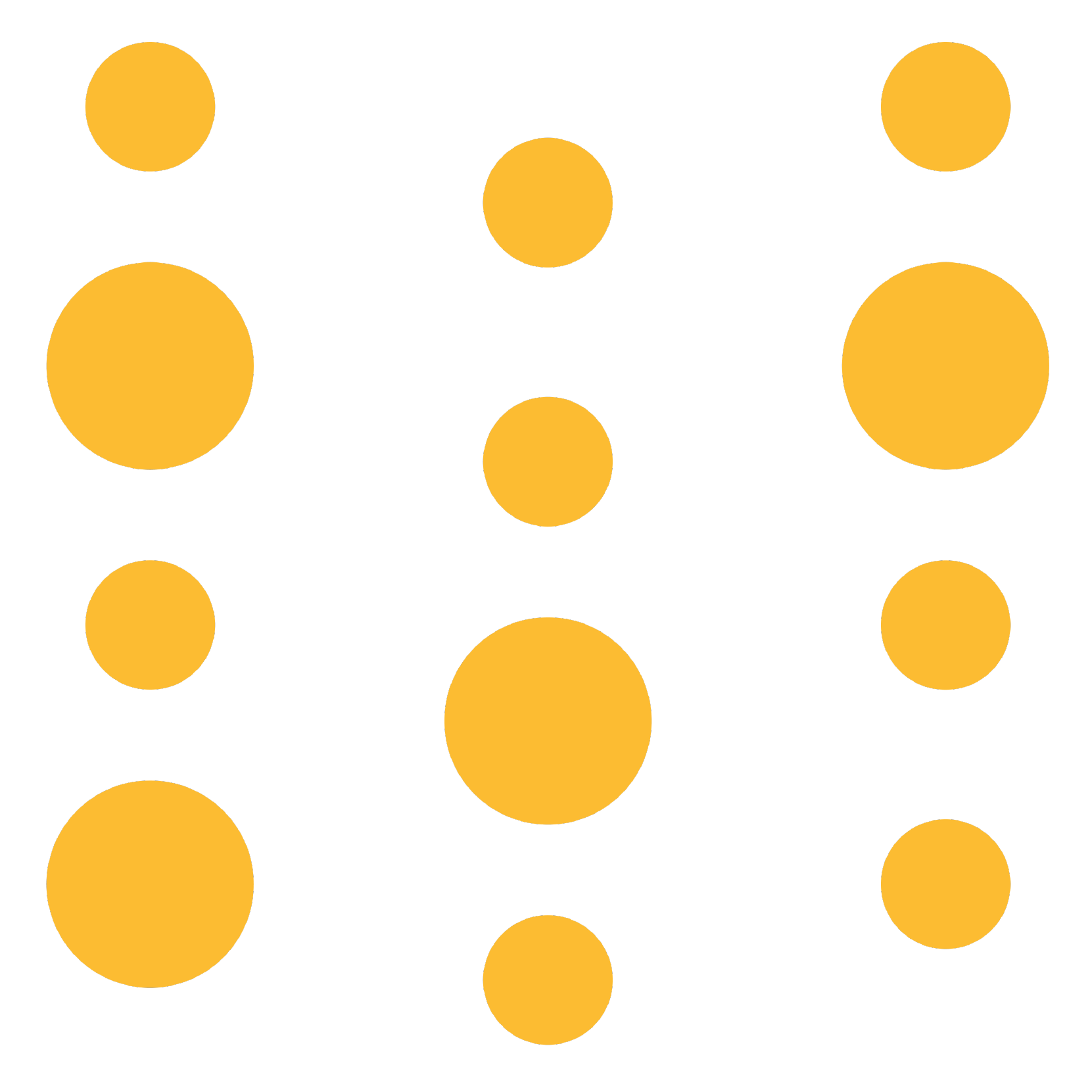}}\xspace}
\newcommand{\comet}{\raisebox{-1.5pt}{\includegraphics[height=1.05em]{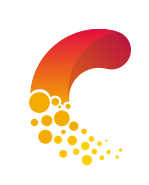}}\xspace}

\usepackage{setspace}

\usepackage{nicematrix}
\newcolumntype{L}[1]{>{\raggedright\let\newline\\\arraybackslash\hspace{0pt}}m{#1}}
\newcolumntype{C}[1]{>{\centering\let\newline\\\arraybackslash\hspace{0pt}}m{#1}}
\newcolumntype{R}[1]{>{\raggedleft\let\newline\\\arraybackslash\hspace{0pt}}m{#1}}
\newcolumntype{P}[1]{>{\centering\let\newline\\\arraybackslash\columncolor{ai2lightpink}}m{#1}}
\newcolumntype{W}[1]{>{\columncolor{white}}c}  %
\newcommand{\coreContrib}{\raisebox{.28em}{\hspace{.05em}\includegraphics[height=.45em]{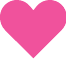}}\hspace{0.1em}}
\newcommand{\starOlmo}{\raisebox{.28em}{\hspace{.05em}\includegraphics[height=.5em]{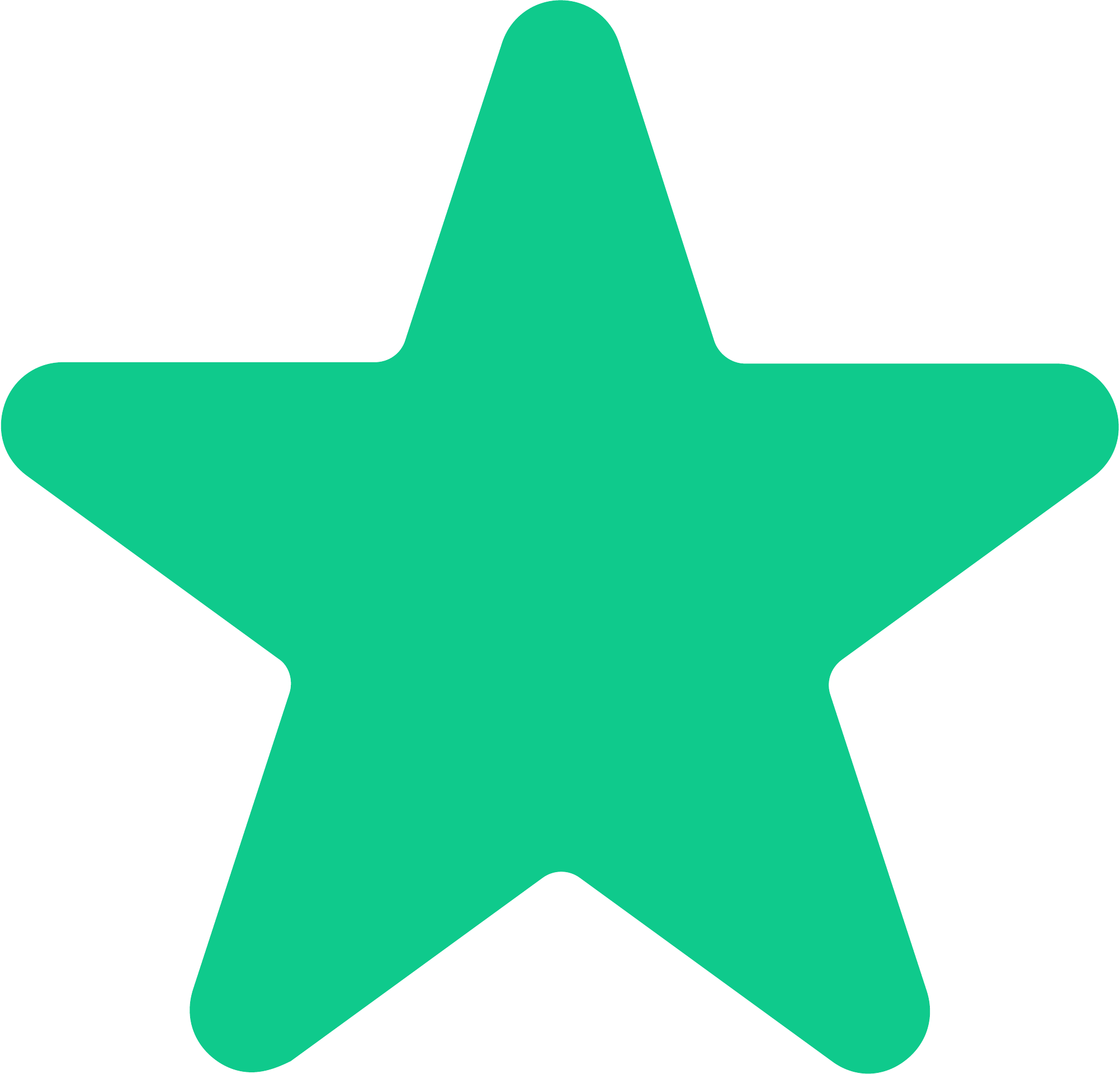}}}
\newcommand{\allenAiAff}{\raisebox{.28em}{\hspace{.02em}\scalebox{0.7}{\textbf{1}}}}

\newcommand{\uwAff}{\raisebox{.28em}{\hspace{.02em}\scalebox{0.7}{\textbf{2}}}}

\newcommand{\nyuAff}{\raisebox{.28em}{\hspace{.02em}\scalebox{0.7}{\textbf{3}}}}
\newcommand{\commaAff}{\raisebox{.28em}{\hspace{.02em}\scalebox{0.7}{\textbf{,}\hspace{0.1em}}}}

\title{2 \olmo 2 Furious}

\authorOne[\starOlmo]{OLMo Team}

\authorTwo[\allenAiAff]{Pete Walsh\coreContrib}
\authorTwo[\allenAiAff]{Luca Soldaini\coreContrib}
\authorTwo[\allenAiAff]{Dirk Groeneveld\coreContrib}
\authorTwo[\allenAiAff]{Kyle Lo\coreContrib}

\authorThree[\allenAiAff]{Shane Arora\coreContrib}
\authorThree[\allenAiAff]{Akshita Bhagia\coreContrib}
\authorThree[\allenAiAff]{Yuling Gu\coreContrib}
\authorThree[\allenAiAff]{Shengyi Huang\coreContrib}
\authorThree[\allenAiAff]{Matt Jordan\coreContrib}
\authorThree[\allenAiAff]{Nathan Lambert\coreContrib}
\authorThree[\allenAiAff]{Dustin Schwenk\coreContrib}
\authorThree[\allenAiAff]{Oyvind Tafjord\coreContrib}

\authorFour[\allenAiAff]{Taira Anderson}{}
\authorFour[\allenAiAff]{David Atkinson}{}
\authorFour[\allenAiAff]{Faeze Brahman}{}
\authorFour[\allenAiAff]{Christopher Clark}
\authorFour[\allenAiAff]{Pradeep Dasigi}
\authorFour[\allenAiAff]{Nouha Dziri}
\authorFour[\allenAiAff]{Allyson Ettinger}
\authorFour[\allenAiAff]{Michal Guerquin}
\authorFour[\allenAiAff]{David Heineman}
\authorFour[\allenAiAff\commaAff\uwAff]{Hamish Ivison}
\authorFour[\allenAiAff\commaAff\uwAff]{Pang Wei Koh}
\authorFour[\allenAiAff\commaAff\uwAff]{Jiacheng Liu}
\authorFour[\allenAiAff]{Saumya Malik}
\authorFour[\allenAiAff\commaAff\nyuAff]{William Merrill}
\authorFour[\allenAiAff]{Lester James V. Miranda}
\authorFour[\allenAiAff]{Jacob Morrison}
\authorFour[\allenAiAff]{Tyler Murray}
\authorFour[\allenAiAff]{Crystal Nam}
\authorFour[\allenAiAff]{Jake Poznanski}
\authorFour[\allenAiAff\commaAff\uwAff]{Valentina Pyatkin}
\authorFour[\allenAiAff]{Aman Rangapur}
\authorFour[\allenAiAff]{Michael Schmitz}
\authorFour[\allenAiAff]{Sam Skjonsberg}
\authorFour[\allenAiAff]{David Wadden}
\authorFour[\allenAiAff]{Christopher Wilhelm}
\authorFour[\allenAiAff]{Michael Wilson}
\authorFour[\uwAff]{Luke Zettlemoyer}

\authorFive[\allenAiAff\commaAff\uwAff]{Ali Farhadi}
\authorFive[\allenAiAff\commaAff\uwAff]{Noah A. Smith\coreContrib}
\authorFive[\allenAiAff\commaAff\uwAff]{Hannaneh Hajishirzi\coreContrib}

\affiliation[\allenAiAff]{Allen Institute for AI}
\affiliation[\uwAff]{University of Washington}
\affiliation[\nyuAff]{New York University}

\contribution[]{\starOlmo\olmotoo was a team effort. \coreContrib{marks core contributors}. See full author contributions \hyperref[sec:contrib]{here}.}

\abstract{

We present \olmotoo, the next generation of our fully open language models. 
\olmotoo includes a family of dense autoregressive language models at 7B, 13B and 32B scales with fully released artifacts---model weights, full training data, training code and recipes, training logs and thousands of intermediate checkpoints.
In this work, we describe our modified model architecture and training recipe, focusing on techniques for achieving better training stability and improved per-token efficiency. 
Our updated pretraining data mixture introduces a new, specialized data mix called \dolminos, which significantly improves model capabilities across many downstream task benchmarks when introduced via late-stage curriculum training (i.e. specialized data during the annealing phase of pretraining). 
Finally, we incorporate best practices from \tulu to develop \olmotooinstruct, focusing on permissive data and extending our final-stage reinforcement learning with verifiable rewards (RLVR).
Our \olmotoo base models sit at the Pareto frontier of performance to training compute, often matching or outperforming open-weight only models like Llama 3.1, Qwen 2.5, and Gemma 2 while using fewer FLOPs and with fully transparent training data, code, and recipe. 
Our fully open \olmotooinstruct models are competitive with open-weight only models of comparable size and even some proprietary models like GPT-3.5 Turbo and GPT 4o Mini.
}

\metadata[\quad\huggingface OLMo 2 Base:]{
\href{https://huggingface.co/allenai/OLMo-2-1124-7B}{\texttt{OLMo-2-1124-7B}} \quad 
\href{https://huggingface.co/allenai/OLMo-2-1124-13B}{\texttt{OLMo-2-1124-13B}} \quad 
\href{https://huggingface.co/allenai/OLMo-2-0325-32B}{\texttt{OLMo-2-0325-32B}}}
\metadata[\quad\huggingface OLMo 2 Instruct:]{
\href{https://huggingface.co/allenai/OLMo-2-1124-7B-Instruct}{\texttt{7B-Instruct}} \quad 
\href{https://huggingface.co/allenai/OLMo-2-1124-13B-Instruct}{\texttt{13B-Instruct}}\quad 
\href{https://huggingface.co/allenai/OLMo-2-0325-32B-Instruct}{\texttt{32B-Instruct}}
}

\metadata[\vspace{.5em}\quad\hfdataset Base Data:]{
    \href{https://huggingface.co/datasets/allenai/olmo-mix-1124}{\texttt{olmo-mix-1124}}~\sans{(pretrain)} \quad 
    \href{https://huggingface.co/datasets/allenai/dolmino-mix-1124}{\texttt{dolmino-mix-1124}}~\sans{(midtrain)}
}

\metadata[\quad\hfdataset Instruct Data:]{
    {\color{ai2accent}\texttt{SFT:}}
    \href{https://huggingface.co/datasets/allenai/tulu-3-sft-olmo-2-mixture}{\texttt{7B}},
    \href{https://huggingface.co/datasets/allenai/tulu-3-sft-olmo-2-mixture}{\texttt{13B}},
    \href{https://huggingface.co/datasets/allenai/tulu-3-sft-olmo-2-mixture-0225}{\texttt{32B}}%
    \quad
    {\color{ai2accent}\texttt{DPO:}}
    \href{https://huggingface.co/datasets/allenai/olmo-2-1124-7b-preference-mix}{\texttt{7B}},
    \href{https://huggingface.co/datasets/allenai/olmo-2-1124-13b-preference-mix}{\texttt{13B}},
    \href{https://huggingface.co/datasets/allenai/olmo-2-0325-32b-preference-mix}{\texttt{32B}}%
    \quad
    \href{https://huggingface.co/datasets/allenai/RLVR-GSM-MATH-IF-Mixed-Constraints}{\color{ai2accent}\texttt{RLVR}}
}

\metadata[\vspace{.5em}\quad\github Training Code:]{
    \href{https://github.com/allenai/OLMo}{\texttt{OLMo}}~\sans{(pretrain v1)} \quad
    \href{https://github.com/allenai/OLMo-core}{\texttt{OLMo-core}}~\sans{(pretrain v2)} \quad
    \href{https://github.com/allenai/open-instruct}{\texttt{open-instruct}}~\sans{(posttrain)}
}
\metadata[\quad\github Eval \& Data Code:]{
    \href{https://github.com/allenai/olmes}{\texttt{olmes}}~\sans{(eval suite)} \quad
    \href{https://github.com/allenai/dolma}{\texttt{dolma}}~\sans{(data curation)}
}

\metadata[\vspace{.5em}\quad\wandb \comet Training Logs:]{
    \href{https://api.wandb.ai/links/ai2-llm/fjn0v0ec}{\texttt{7B}} \quad
    \href{https://api.wandb.ai/links/ai2-llm/ypmumwpc}{\texttt{13B}} \quad
    \href{https://www.comet.com/ai2/olmo-2-0325-32b/reports/olmo-2-0325-32b?shareable=WhT37Wy7jqttDoy6ysDBumQzf}{\texttt{32B}}
}

\metadata[\vspace{.5em}\quad\aitoo Demo:]{\href{https://playground.allenai.org/}{\texttt{playground.allenai.org}}}

\metadata[\vspace{.5em}\quad\emailLogo Contact:]{\color{ai2accent}{\texttt{olmo@allenai.org}}}

\begin{document}

\maketitle

\newpage
\setcounter{tocdepth}{2}
\tableofcontents
\newpage

\begin{figure}[ht]
    \centering
    \includegraphics[width=0.8\textwidth]{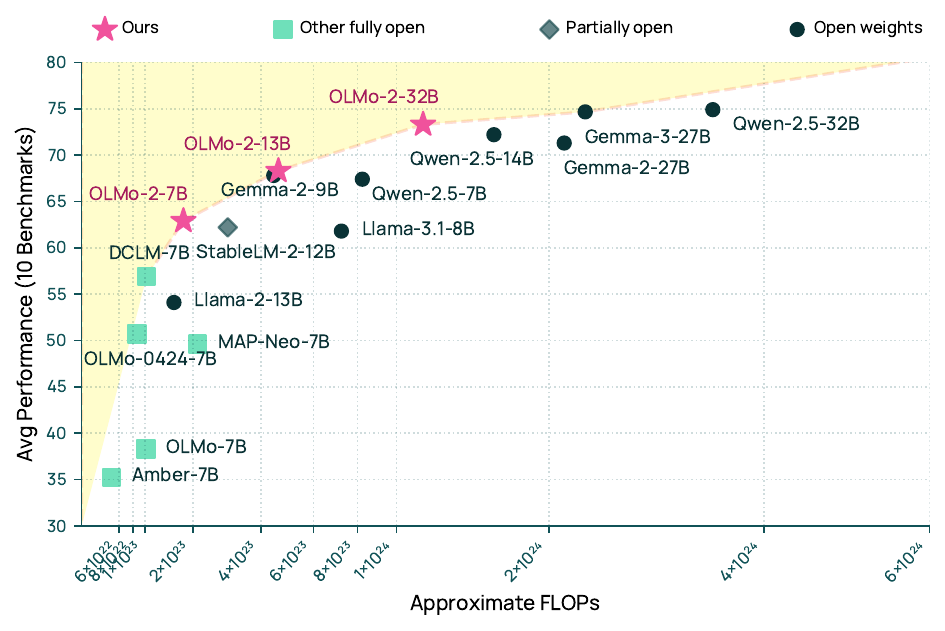}
    \caption{Performance to pretraining FLOPs ($\approx$ 6 $\times$ training tokens $\times$ model size; \citealp{kaplan2020scaling}) for \olmotoo and comparable models. 
    We see that the fully open \olmotoo lies on the Pareto frontier, outperforming many other models of varying levels of openness at multiple sizes.
    For full results, see Table~\ref{tab:evals_overview}.
    }    \label{fig:pareto}
\end{figure}

\section{Introduction}
\label{sec:intro}

The open language model ecosystem has grown rapidly in the past year.
We've seen a surge in open weights models from established developers---Llama 3~\citep{dubey2024llama}, DBRX~\citep{dbrx_blog}, Yi 1.5~\citep{young2024yi}, Qwen 2~\citep{yang2024qwen2}, Falcon~\citep{falcon2,falcon3}, Mistral~\citep{mistral2024large2}, Ministral~\citep{mistral2024ministral}, Phi~\citep{abdin2024phi,Abdin2024Phi4TR}---
and new contributors---
Gemma~\citep{gemma,gemma2,gemmateam2025gemma3technicalreport}, Grok~\citep{grok_blog}, Command R~\citep{cohere2024commandR,cohere2024commandRplus,cohere2024commandR7B}
---substantially closing the gap between publicly available and closed systems~\citep{cottier2024open}.
Yet, these open-weights models are only the \emph{final} artifacts of sophisticated language model recipes and complex development pipelines, and by themselves are not sufficient to support diverse forms of research into language model behaviors and uses.

In response, prior works including our first \olmo~\citep{Groeneveld2024OLMoAT}, Pythia~\citep{biderman2023pythia}, Amber~\citep{liu2023llm360}, DCLM~\citep{dclm}, MAP Neo~\citep{zhang2024mapneo} and SmolLM~\citep{allal2024SmolLM,allal2024SmolLM2} have adopted a {\bf fully open approach}, releasing not just model weights but also training data, training code and well-documented recipes to support reproduction. 
Artifacts from fully open language modeling efforts have played a crucial role in studying training dynamics~\citep{land2024fishing,Jin2024DemystifyingLM}, concept acquisition~\citep{chang2024large}, and memorization~\citep{Antoniades2024GeneralizationVM,Shaib2024DetectionAM} in language models.
Despite these developments, a gap remains between the models with the best reported performance and that of open models.

Modern language model development is an iterative process, whereby limitations of current iterations motivate future development.
Our previous release \citep[\olmoapril;][]{olmo_blog} focused on improving performance on key tasks (\textit{e.g.}, MMLU) through better pretraining data mixing and curricula.
In this technical report, we introduce \textbf{\olmotoo}, a new family of 7B, 13B and 32B models trained on up to 6T tokens. 
On English academic benchmarks, these models are competitive with the open weight Llama 3.1, Qwen 2.5, and Gemma 2 families of models (Figure~\ref{fig:pareto}).
We further validate our pretrained model is an effective base model for downstream post-training by applying our T\"ulu 3 recipe~\citep{lambert2024tulu3}.
The resulting family of models, called \olmotooinstruct, are competitive with powerful open-weights only models and even some popular proprietary models like GPT-3.5 Turbo and GPT 4o Mini.
This technical report focuses on \textbf{four key areas} we targeted during development of \olmotoo:
\begin{itemize}
    \item \textbf{\diveStability.} Language model training runs are often plagued by training instabilities and loss spikes, which are costly and known to be a detriment to final model performance. 
    We discuss techniques we used to improve training stability, which was critical to ensuring performance of the final trained model (Section~\S\ref{sec:stability}).
    \item \textbf{\diveAnnealing.} \olmoapril~\citep{olmo_blog}, DBRX~\citep{dbrx_blog}, and Llama~3~\citep{dubey2024llama} demonstrated the usefulness of data curricula for pretraining, as discussed by \citet{blakeney2024doesdatasparkjoy}.
    We discuss the advantages of splitting pretraining into two stages, with the latter \emph{mid-training} stage being used to infuse new knowledge and patch deficiencies in capabilities. 
    Further, we show how data sources for mid-training can be independently assessed to reduce experimentation cost through a technique we call \emph{micro-annealing} (Section~\S\ref{sec:diveAnnealing}). 
    \item \textbf{\divePost.} A key deliverable for a successful base model is its ability to be finetuned to downstream use-cases.  
    We introduce \olmotooinstruct built on the Tülu~3 recipe~\citep{lambert2024tulu3}, and show how improvements in base models translated to better chat variants. 
    We focus on permissive data and expand the reinforcement learning with verifiable rewards (RLVR) pipeline to multiple stages for maximum performance
    (Section~\S\ref{sec:post-train}).
    \item \textbf{\diveInfra.} 
    High performance and reliable infrastructure is crucial for successful pretraining;
    yet, many pretraining papers do not discuss their training stack, or gloss over crucial details.
    We discuss changes from \olmoapril that enable the improvements of \olmotoo, and how investing in solutions that let us monitor and orchestrate infrastructure helped us reduce failure rates and increase cluster utilization (Section~\S\ref{sec:diveInfra}).

\end{itemize}

Alongside these deep dives, we provide a description of the full model development procedure in Section~\S\ref{sec:olmo-family}: training data, pretraining, post-training, and evaluation. 
We highlight changes from \olmozero and \olmoapril when appropriate, and reference related projects, such as our scaling laws effort to efficiently estimate model downstream performance~\citep{bhagia2024establishingtaskscalinglaws} and benchmark standardization through the OLMES evaluation framework~\citep{olmes}.

\section{\olmotoo Family}
\label{sec:olmo-family}
This section provides an overview of \olmotoo and highlights improvements over \olmoapril and previous OLMo models\footnote{Model architecture changes over \olmozero and \olmoapril are described in Section~\S\ref{sec:architecture}; for an overview of data and training recipes, see \citet{Groeneveld2024OLMoAT} and \citet{olmo_blog} respectively.}. 
The \olmotoo family has more tokens, more parameters, and has better downstream task results compared to \olmoapril.
We explain the crucial details required to achieve competitive results in our mission of making state-of-the-art language models accessible.
Accordingly, we release all training code, data, and recipes openly under the Apache 2.0 license wherever possible, and under the most permissive available license otherwise.

\subsection{Model Architecture}
\label{sec:architecture}

\begin{table}[h]
    \centering
    \small
    \begin{tabular}{llll}
        \toprule
        & \textbf{\olmozero (0224)} & \textbf{\olmoapril} & \textbf{\olmotoo} \\
        \midrule
        \rowcolor{ai2offwhite}
        \textbf{Biases} & None & None & None \\
        \textbf{Activation} & SwiGLU & SwiGLU & SwiGLU \\
        \rowcolor{ai2offwhite}
        \textbf{RoPE $\theta$} & $1 \cdot 10^{4}$ & $1 \cdot 10^{4}$ & $5 \cdot 10^{5}$ \\
        \textbf{QKV Normalization} & None & Clip to $8$ & QK-Norm \\
        \rowcolor{ai2offwhite}
        \textbf{Layer Norm} & non-parametric & non-parametric & RMSNorm \\
        \textbf{Layer Norm Applied to} & Inputs & Inputs & Outputs \\
        \rowcolor{ai2offwhite}
        \textbf{Z-Loss Weight} & $0$ & $0$ & $10^{-5}$ \\
        \textbf{Weight Decay on Embeddings} & Yes & Yes & No \\
        \bottomrule
    \end{tabular}
    \vspace{1em}
    \caption{Summary of how \olmo family model architectures have evolved over time. Latest \olmotoo changes were motivated by experiments showing improved training stability. 
    Full descriptions in \S\ref{sec:architecture}.}
    \label{tab:model_architecture}
\end{table}

Table~\ref{tab:model_architecture} provides an overview of how the model architecture has evolved through iterations in the \olmo family. We provide details below:

We adopt a decoder-only transformer architecture based on \citet{vaswani2017attention},
and deliver 7B, 13B and 32B parameter variants as described in Table~\ref{tab:model_sizes}.
Our architecture is very similar to the first iteration of \olmo \citep{Groeneveld2024OLMoAT}, with several changes to improve training stability (see Section~\S\ref{sec:stability}) and performance.
The original \olmo modified the decoder-only transformer architecture~\citep{vaswani2017attention} with:
\begin{itemize}
    \item \textbf{No biases:} We exclude all bias terms from our architecture~\citep[\textit{inter alia}]{Groeneveld2024OLMoAT, palm}.
    \item \textbf{SwiGLU activation function:} We use the SwiGLU activation function \citep{Shazeer2020GLUVI} and set the corresponding hidden size to approximately $\frac{8}{3}d$, but increased to the closest multiple of 128 ($11,008$ for our 7B model) to improve throughput.
    \item \textbf{Rotary positional embeddings (RoPE):} We replace absolute positional embeddings with rotary positional embeddings (RoPE; \citealp{Su2021RoFormerET}).
\end{itemize}

When building \olmoapril, we made modifications for training stability and downstream performance: 
\begin{itemize}
    \item \textbf{QKV Clipping:} For training stability, also as seen in DBRX~\citep{dbrx_blog}.
    \item \textbf{Increased context:} From 2048 to 4096.
\end{itemize}

Finally, this work introduces \olmotoo which made further modifications:

\begin{itemize}
    \item \textbf{RMSNorm:} We use the RMSNorm \citep{RMSNorm} variant of LayerNorm \citep{Ba2016LayerNorm} without a bias term to normalize activations, instead of nonparametric LayerNorm.
    \item \textbf{Reordered norm:} We normalize the outputs to the attention and feedforward (MLP) layers within each transformer block, instead of the inputs. So the formula for each block becomes:
    \begin{align}
        & \pmb{h}  := \pmb{x} + \text{RMSNorm}(\text{Attention}(\pmb{x})) \\
        & \pmb{h}_{\text{out}}  := \pmb{h} + \text{RMSNorm}(\text{MLP}(\pmb{x}))
    \end{align}
    where $\pmb{x}$ is the input to the layer, $\pmb{h}$ is an intermediate hidden state, and $\pmb{h}_{\text{out}}$ is the output.
    This strategy was first proposed by~\citet{swin2} to stabilize training.
    \item \textbf{QK-norm:} Following \cite{Dehghani2023ScalingVT} we normalize the key and query projections with RMSNorm before calculating attention. This avoids attention logits being too large, which can lead to training loss divergence.
    \item \textbf{Z-Loss:} Following~\citet{palm}, \citet{chameleon}, and~\citet{mitch}, we adopt z-loss regularization, as it has been empirically shown to improve run stability. 
    \item \textbf{RoPE $\pmb{\theta = 5e5}$:} We increase the RoPE $\theta$ to 500,000 from 10,000. This approach increases the resolution of positional encoding, matching~\citet{dubey2024llama}.
\end{itemize}

\subsection{Tokenizer}

\olmozero and \olmoapril were trained using a modified version of the GPT-NeoX-20B tokenizer~\citep{black2022gpt} that includes special tokens \texttt{|||PHONE\_NUMBER|||}, \texttt{|||EMAIL\_ADDRESS|||}, and \texttt{|||IP\_ADDRESS|||}, which were used to mask personal identifiable information.

As suggested by \citet{tao2024scaling}, we employ a larger tokenizer vocabulary for \olmotoo.
We borrow pre-tokenizer and vocabulary from \texttt{cl100k}, the tokenizer developed for GPT-3.5~\citep{gpt35} and GPT-4~\citep{gpt4}, which is licensed under Apache 2.0\footnote{\href{https://github.com/openai/tiktoken/issues/92}{\path{github.com/openai/tiktoken/issues/92}}}.
To maintain backwards compatibility with early Dolma data sources, we add the same masking tokens used in previous \olmo models.\footnote{Specifically, these tokens such as \texttt{|||IP\_ADDRESS|||} appear in early subsets of \dolma dataset. 
We opt to keep them in vocabulary so that, if tokenizing any of these older sources, they will not get split into multiple tokens.}

\begin{table}[h]
    \centering
    \small
    \begin{tabular}{l c c c}
    \toprule
        \textbf{Tokenizer} & \textbf{OLMES} (\sans{CF}) & \textbf{OLMES Gen} & \textbf{MMLU} (\sans{CF}) \\
    \midrule
        \olmozero tokenizer & 59.8 & 42.4 & 34.8 \\
         \olmotoo tokenizer & 60.6 & 42.7 & 35.2 \\
    \bottomrule     
    \end{tabular}
    \caption{
        Comparison of \olmozero and \olmotoo tokenizers on a 1B model pretrained for 100B tokens from DCLM baseline.
        Following~\citet{olmes}, OLMES and MMLU use CF format, which is more informative for small models.
    }
    \label{tab:tokenizers}
\end{table}

We compare the two tokenizers at a smaller scale in Table~\ref{tab:tokenizers}. 
We see measurable gains when switching to the new tokenizer, particularly in OLMES tasks. 
Per \cite{tao2024scaling}, at this model size and compute budget, the larger \olmotoo tokenizer is at a slight disadvantage; 
we expect improvement coming from larger vocabulary to be more decisive at larger scales and for models trained on more tokens.

\subsection{Base Model Training Recipe}
\label{section:training-recipe}
Following previous \olmo models, as well as recent advances in curriculum learning~\citep{blakeney2024doesdatasparkjoy,ibrahim2024simplescalablestrategiescontinually}, base \olmotoo models are trained in \textbf{two stages} each with its corresponding data mix. 
The first \textit{pretraining} stage is the longest ($\geqslant90\%$ training FLOPs), and uses mostly web-sourced data.
In this stage, we use an iteration on our pretraining mix of high-quality web data drawing on other recent open data releases.
During the second stage, which we refer to as \textit{mid-training}  ($5\textup{--}10\%$ of training FLOPs), we up-sample the highest-quality web documents and curated non-web sources;
we also employ synthetic data crafted to patch math capabilities of the model. 

\begin{table}[!ht] %
    \centering %
    \begin{small}
    \begin{tabular}{lccc} %
        \toprule
        & \textbf{\olmotoo 7B} & \textbf{\olmotoo 13B} & \textbf{\olmotoo 32B} \\ 
        \midrule
        \rowcolor{ai2offwhite}\textbf{Layers} & 32 & 40 & 64 \\ 
        \textbf{Hidden Size}~$(d_{model})$ & 4096 & 5120 & 5120 \\ 
        \rowcolor{ai2offwhite}\textbf{Attention Heads (Q/KV)} & 32/32 (MHA) & 40/40 (MHA) & 40/8 (GQA) \\ 
        \textbf{Batch Size} & 1024 & 2048 & 2048 \\ 
        \rowcolor{ai2offwhite}\textbf{Sequence Length} & 4096 & 4096 & 4096 \\ 
        \textbf{Gradient Clipping} & 1.0 & 1.0 & 1.0 \\ 
        \rowcolor{ai2offwhite}\textbf{Peak LR} & $3.0 \cdot 10^{-4}$ & $9.0 \cdot 10^{-4}$ & $6.0 \cdot 10^{-4}$ \\ 
        \textbf{LR Warmup} & 2000 steps & 2000 steps & 2000 steps \\ 
        \rowcolor{ai2offwhite}
        \textbf{LR Schedule (Cosine)}
        &
        \begin{tabular}{@{}c@{}}
            5T tokens 
        \end{tabular}
        &
        5T tokens
        &
        \begin{tabular}{@{}c@{}}
        6.5T tokens
        \end{tabular}
        \\
        \textbf{LR Schedule Truncation}
        &
        \begin{tabular}{@{}c@{}}
            \textit{\footnotesize(after 4T)}
        \end{tabular}
        &
        n/a
        &
        \begin{tabular}{@{}c@{}}
        \textit{\footnotesize{after 6T}}
        \end{tabular}
        \\
        \bottomrule
    \end{tabular}
    \end{small}
    \caption{\centering \olmotoo hyperparameters.} 
    \label{tab:model_sizes}
\end{table}

\paragraph{Stage 1: Pretraining}

The first stage---\textit{pretraining}---is the longest ($90\textup{--}95\%$ of training FLOPs).
We report key architecture and training details in Table~\ref{tab:model_sizes}.
Key details include our switch from multi-head attention (MHA) to grouped query attention (GQA)~\citep{ainslie-etal-2023-gqa} to scale the 32B model, inspired by its use in concurrent work Qwen 3~\citep{yang2025qwen3technicalreport}.
\olmotoo training used random initialization from a truncated normal distribution with a mean of 0 and a standard deviation of 0.02 and a learning rate schedule that warms up the learning rate from 0 to the peak learning rate over 2000 steps, followed by a cosine decay calibrated to reach 10\% of the peak learning rate after a specified max tokens.

\paragraph{Stage 2: Mid-training}
We refer to the shorter second stage as \textit{mid-training}  ($5\textup{--}10\%$ of training FLOPs), where we linearly decay the learning rate to zero over the remaining length of the run.\footnote{While the concept of multiple stages of self-supervised training is not new (\textit{e.g.}, \citealt{gururangan2020dontStopPretraining}), we adopt the term \textit{mid-training} from \citet{abdin2024phi} and \citet{openai2024midtraining}.}

We curated a smaller, focused mixture---\textbf{\dolminos}---to imbue the model with domain knowledge from increased exposure to STEM references and high quality text as well as skills that remained lacking after the initial pretraining stage (e.g. math-solving capabilities). 
We up-sample high-quality web documents and curated non-web sources;
we also employ synthetic data crafted to patch math capabilities of the model.

\paragraph{Model Merging or ``Souping''}
To get the most out of this high-quality data, and to find a better local minimum, we perform this step multiple times with different random data orders, and then average the resulting models~\citep{matena2022mergingmodelsfisherweightedaveraging,modelsoups}. 
For \olmotoo 7B, we anneal three separate times for 50B tokens each, with different randomized data orders; 
we average the resulting models to produce the final model.
For both \olmotoo 13B and \olmotoo 32B, we train three separate times for 100B tokens each (same number of update steps as the 7B), and then a fourth time for 300B tokens.
The final model is the average of all four models.
For further details, refer to Section~\S\ref{sec:diveAnnealing}.

\paragraph{Overall} In total, \olmotoo 7B is trained on $4.05$ trillion tokens ($3.90$ trillion for pretraining stage), \olmotoo 13B is trained on $5.6$ trillion tokens ($5$ trillion for pretraining stage), and \olmotoo 32B is trained on $6.6$ trillion tokens  ($6.06$ trillion for pretraining stage).

\subsection{Base Model Data}
\label{sec:pretrain-data}

We provide a brief overview of the data mix for pretraining and mid-training in this section.

\subsubsection{Pretraining data: \olmomix}

\begin{table}[h]
\centering
\small
\renewcommand{\arraystretch}{1}
\begin{tabular}{l l r r r r}
\toprule
\textbf{Source} & 
\textbf{Type} & 
\textbf{Tokens} & 
\textbf{Words} & 
\textbf{Bytes} & 
\textbf{Docs} \\
\midrule
\rowcolor{midgrey}\multicolumn{6}{c}{\textbf{\textit{Pretraining \ding{70} OLMo 2 1124 Mix}}} \\
\rowcolor{lightgrey}DCLM-Baseline & Web pages & 3.71T & 3.32T & 21.32T & 2.95B \\
    \rowcolor{lightgrey}    
    \begin{tabular}{@{}l@{}}
        \rowcolor{lightgrey}
        {StarCoder} 
        \\[-.6em]
        {
            \hspace{0.8em}
            \scriptsize
            \color{neutralFive}
            filtered version
        }
        \\[-.6em]
        {
            \hspace{0.8em}
            \scriptsize
            \color{neutralFive}
            from OLMoE Mix
        }
    \end{tabular} &
    Code & 
    83.0B & 
    70.0B & 
    459B & 
    78.7M 
\\    
    \rowcolor{lightgrey}    
    \begin{tabular}{@{}l@{}}
        \rowcolor{lightgrey}
        {peS2o} 
        \\[-.6em]
        {
            \hspace{0.8em}
            \scriptsize
            \color{neutralFive}
            from Dolma 1.7
        }
    \end{tabular} & 
    Academic papers & 
    58.6B & 
    51.1B & 
    413B & 
    38.8M 
\\
\rowcolor{lightgrey}arXiv & STEM papers & 20.8B & 19.3B & 77.2B & 3.95M \\
\rowcolor{lightgrey} OpenWebMath & Math web pages & 12.2B & 11.1B & 47.2B & 2.89M \\
\rowcolor{lightgrey}Algebraic Stack & Math proofs code & 11.8B & 10.8B & 44.0B & 2.83M \\
    \rowcolor{lightgrey}
    \begin{tabular}{@{}l@{}}
        \rowcolor{lightgrey}
        {Wikipedia \& Wikibooks} 
        \\[-.6em]
        {
            \hspace{0.8em}
            \scriptsize
            \color{neutralFive}
            from Dolma 1.7
        }
    \end{tabular} &
    Encyclopedic & 
    3.7B & 
    3.16B & 
    16.2B & 
    6.17M 
\\
\rowcolor{lightgrey}\textbf{Total} & & \textbf{3.90T} & \textbf{3.48T} & \textbf{22.38T} & \textbf{3.08B} \\
\bottomrule
\end{tabular}
\caption{
\textbf{Composition of the pretraining data for \olmotoo{}}. The \textsc{OLMo 2 1124 Mix} is composed of
StarCoder~\citep{li2023starcoder,kocetkov2022stack3tbpermissively}, peS2o~\citep{peS2o}, web text from DCLM~\citep{dclm} and Wiki come from Dolma 1.7~\citep{soldaini2024dolma}. arXiv comes from Red-Pajama~\citep{together2023redpajama},  while OpenWebMath~\citep{paster2023openwebmath} and Algebraic Stack come from ProofPile II~\citep{azerbayev2023llemma}.
}
\label{table:data-stage-1}
\end{table}

The mix used for this stage is shown in Table~\ref{table:data-stage-1}. It consists of approximately 3.9 trillion tokens, with over 95\% derived from web data. 
We refer to this set as \olmomix. 
This is the same pretraining data used in  \textsc{OLMoE}~\citep{muennighoff2024olmoeopenmixtureofexpertslanguage}:
We combine data from DCLM~\citep{dclm} and Dolma 1.7~\citep{soldaini2024dolma}.
From DCLM, we use the ``\textit{baseline 1.0}'' mix.\footnote{Available at \href{https://huggingface.co/datasets/mlfoundations/dclm-baseline-1.0}{\hfdataset\path{mlfoundations/dclm-baseline-1.0}}}
From Dolma, we use the arXiv~\citep{together2023redpajama}, OpenWebMath~\citep{paster2023openwebmath}, Algebraic Stack, peS2o~\citep{peS2o}, and Wikipedia subsets. 
arXiv, OpenWebMath, and Algebraic Stack were originally part of ProofPile II~\citep{azerbayev2023llemma}.
Finally, we include code from StarCoder~\citep{li2023starcoder}, which is derived from permissively-licensed repositories from GitHub~\citep{kocetkov2022stack3tbpermissively}. 
In an attempt to include higher quality code, we remove any document from a repository with fewer than 2 stars on GitHub.
Further, through manual inspection of this source, we found it to contain documents encoded in binary format or containing mostly numerical content; 
to remove them, we discarded documents whose most frequent word constitutes over 30\% of the document, or whose top-2 most frequent words constitute over 50\% of the document.
To mitigate possible training loss spikes, we remove documents with repeated sequences of 32 or more n-grams.
We report details and show effectiveness of this intervention in Section~\S\ref{sec:datafilter}.

\subsubsection{Mid-training data: \dolminos}

\begin{table}[h]
\centering
\small
\renewcommand{\arraystretch}{1}
\begin{tabular}{l l r r r r}
\toprule
\textbf{Source} & 
\textbf{Type} & 
\textbf{Tokens} & 
\textbf{Words} & 
\textbf{Bytes} & 
\textbf{Docs} \\
\midrule
    \rowcolor{ai2midwhite}
    \multicolumn{6}{c}{\textbf{\textit{Mid-Training \ding{70} Dolmino High Quality Subset}}} 
\\
    \rowcolor{ai2offwhite}
    \begin{tabular}{@{}l@{}}
        \rowcolor{ai2offwhite}
        {DCLM-Baseline} 
        \\[-.6em]
        {
            \hspace{0.8em}
            \scriptsize
            \color{neutralFive}
            FastText top 7\%
        }
        \\[-.6em]
        {
            \hspace{0.8em}
            \scriptsize
            \color{neutralFive}
            $\text{FineWeb}\geqslant2$
        }
    \end{tabular} & 
    High quality web & 
    752B &  
    670B & 
    4.56T & 
    606M 
\\
    \rowcolor{ai2offwhite}
    \begin{tabular}{@{}l@{}}
        \rowcolor{ai2offwhite}
        {FLAN} 
        \\[-.6em]
        {
            \hspace{0.8em}
            \scriptsize
            \color{neutralFive}
            from Dolma 1.7
        }
        \\[-.6em]
        {
            \hspace{0.8em}
            \scriptsize
            \color{neutralFive}
            decontaminated
        }
    \end{tabular} & 
    Instruction data & 
    17.0B & 
    14.4B & 
    98.2B & 
    57.3M 
\\
    \rowcolor{ai2offwhite}
    \begin{tabular}{@{}l@{}}
        \rowcolor{ai2offwhite}
        {peS2o} 
        \\[-.6em]
        {
            \hspace{0.8em}
            \scriptsize
            \color{neutralFive}
            from Dolma 1.7
        }
    \end{tabular} & 
    Academic papers & 
    58.6B & 
    51.1B & 
    413B & 
    38.8M 
\\
    \rowcolor{ai2offwhite}
    \begin{tabular}{@{}l@{}}
        \rowcolor{ai2offwhite}
        {Wikipedia \& Wikibooks} 
        \\[-.6em]
        {
            \hspace{0.8em}
            \scriptsize
            \color{neutralFive}
            from Dolma 1.7
        }
    \end{tabular} &
    Encyclopedic & 
    3.7B & 
    3.16B & 
    16.2B & 
    6.17M 
\\
    \rowcolor{ai2offwhite}
    \begin{tabular}{@{}l@{}}
        \rowcolor{ai2offwhite}
        {Stack Exchange} 
        \\[-.6em]
        {
            \hspace{0.8em}
            \scriptsize
            \color{neutralFive}
            09/30/2024 dump
        }
        \\[-.6em]
        {
            \hspace{0.8em}
            \scriptsize
            \color{neutralFive}
            curated Q\&A data
        }
    \end{tabular} &
    Q\&A & 
    1.26B & 
    1.14B & 
    7.72B & 
    2.48M 
\\
    \rowcolor{ai2offwhite}
    \textbf{High quality total} & 
    & 
    \textbf{832.6B} & 
    \textbf{739.8B} & 
    \textbf{5.09T} & 
    \textbf{710.8M} 
\\
    \rowcolor{ai2midpink}
    \multicolumn{6}{c}{
        \textbf{
            \textit{
                Mid-training \ding{70} Dolmino Math Mix
            }
        }
    } 
\\
    \rowcolor{ai2lightpink}
    TuluMath & 
    Synthetic math & 
    230M & 
    222M & 
    1.03B & 
    220K 
\\
    \rowcolor{ai2lightpink}
    Dolmino SynthMath & 
    Synthetic math & 
    28.7M & 
    35.1M & 
    163M & 
    725K 
\\
    \rowcolor{ai2lightpink}
    TinyGSM-MIND & 
    Synthetic math & 
    6.48B & 
    5.68B & 
    25.52B & 
    17M 
\\
    \rowcolor{ai2lightpink}
    \begin{tabular}{@{}l@{}}
        \rowcolor{ai2lightpink}
        {MathCoder2} 
        \\[-.4em]
        {Synth Books} 
        \\[-.6em]
        {
            \hspace{0.8em}
            \scriptsize
            \color{neutralFive}
            Ajibawa-2023
        }
        \\[-.6em]
        {
            \hspace{0.8em}
            \scriptsize
            \color{neutralFive}
            M-A-P Matrix
        }
    \end{tabular} &
    Synthetic Math & 
    3.87B & 
    3.71B & 
    18.4B & 
    2.83M 
\\
    \rowcolor{ai2lightpink}
    \begin{tabular}{@{}l@{}}
        \rowcolor{ai2lightpink}
        {Metamath} 
        \\[-.6em]
        {
            \hspace{0.8em}
            \scriptsize
            \color{neutralFive}
            OWM-filtered
        }
    \end{tabular} &
    Math & 
    84.2M & 
    76.6M & 
    741M & 
    383K 
\\
    \rowcolor{ai2lightpink}
    \begin{tabular}{@{}l@{}}
        \rowcolor{ai2lightpink}
        {CodeSearchNet} 
        \\[-.6em]
        {
            \hspace{0.8em}
            \scriptsize
            \color{neutralFive}
            OWM-filtered
        }
    \end{tabular} &
    Code & 
    1.78M & 
    1.41M & 
    29.8M & 
    7.27K 
\\
    \rowcolor{ai2lightpink}
    \begin{tabular}{@{}l@{}}
        \rowcolor{ai2lightpink}
        {GSM8K} 
        \\[-.6em]
        {
            \hspace{0.8em}
            \scriptsize
            \color{neutralFive}
            Train split
        }
    \end{tabular} &
    Math & 
    2.74M & 
    3.00M & 
    25.3M & 
    17.6K 
\\
    \rowcolor{ai2lightpink}
    \textbf{Math total} & 
    & 
    \textbf{10.7B} & 
    \textbf{9.73B} & 
    \textbf{45.9B} & 
    \textbf{21.37M} 
\\
\bottomrule
\end{tabular}
\caption{
\textbf{Composition of the mid-training data (Dolmino)}.  
From this set, we create samples of 50B, 100B and 300B tokens to mid-train \olmotoo on. 
See Section~\S\ref{sec:diveAnnealing} for details regarding individual source details, and Table \ref{tab:dolmino-composition-pretty} for the specific composition of each annealing mixture.
}
\label{table:stage-2-data}
\end{table}

After the initial pretraining stage on mostly web data, we further train with a mixture of web data that has been more restrictively filtered for quality and a collection of domain-specific high quality data, much of which is synthetic. 
The purpose of this mixture is to imbue the model with math-centric skills and provide focused exposure to STEM references and high quality text. 
We generate several variants of this mixture, with varying sizes, but generally refer to this mixture as \dolminos. 
The base sources from which \dolminos is subsampled are described in Table~\ref{table:stage-2-data}. 
We refer the reader to Section~\S\ref{sec:diveAnnealing} for a {\bf deep dive} detailing our processes for experimenting and curating data for this mix.

\subsection{Evaluation and Results}
\label{sec:basemodeleval}

\olmotoo is evaluated via standard language model benchmarks. 
Further, we apply post-training to \olmotoo and evaluate the result---\olmotooinstruct---on a diverse set of tasks to assess the adaptation potential of our base model.

\begin{table}[ht!]
\setlength\tabcolsep{2pt} 
\renewcommand{\arraystretch}{1.1}
\begin{center}
\begin{small}
\begin{tabular}{lcc|cccccc|cccc}
\toprule
    &&\multicolumn{7}{c}{\quad \quad \quad \quad \textbf{\texttt{Dev Benchmarks}}} & \multicolumn{4}{c}{\textbf{\texttt{Held-out Evals}}} \\
    {\textbf{\fontsize{8}{8}\selectfont~Model}} &
    {\textbf{\fontsize{8}{8}\selectfont~Avg}} &
    {\textbf{\fontsize{8}{8}\selectfont~FLOP$\times 10^{23}$}} & 
    {\textbf{\fontsize{8}{8}\selectfont~MMLU}} & 
    {$\textbf{\fontsize{8}{8}\selectfont~ARC}_\textbf{\fontsize{6}{6}\selectfont~C}$} & 
    {\textbf{\fontsize{8}{8}\selectfont~HSwag}} & 
    {\textbf{\fontsize{8}{8}\selectfont~WinoG}} & 
    {\textbf{\fontsize{8}{8}\selectfont~NQ}} & 
    {\textbf{\fontsize{8}{8}\selectfont~DROP}} & 
    {\textbf{\fontsize{8}{8}\selectfont~AGIEval}} & 
    {\textbf{\fontsize{8}{8}\selectfont~GSM8K}} & 
    {$\textbf{\fontsize{8}{8}\selectfont~MMLU}_\textbf{\fontsize{6}{6}\selectfont~PRO}$} &
    {\textbf{\fontsize{8}{8}\selectfont~TriviaQA}} 
    \\
\midrule
\rowcolor{midgrey} 
\multicolumn{13}{c}{\textbf{Open-weights models 7-14B Parameters}} \\
\rowcolor{lightgrey} Mistral 7B & 58.9 & \textit{n/a} & 63.5 & 78.3 & 83.1 & 77.7 & 37.2 & 51.8 & 47.3 & 40.1 & 30.0 & 80.3 \\
\rowcolor{lightgrey} Llama 3.1 8B & 61.8 & 7.2 & 66.9 & 79.5 & 81.6 & 76.6 & 33.9 & 56.4 & 51.3 & 56.5 & 34.7 & 80.3 \\
\rowcolor{lightgrey} Qwen 2.5 7B & 67.4 & 8.2 & 74.4 & 89.5 & 89.7 & 74.2 & 29.9 & 55.8 & 63.7 & 81.5 & 45.8 & 69.4 \\
\rowcolor{lightgrey} Qwen 3 8B & 66.6 & \emph{n/c} & 76.8 & 91.2 & 89.5 & 69.9 & 21.8 & 61.8 & 64.3 & 74.8 & 50.6 & 66.5 \\
\rowcolor{lightgrey} Gemma 2 9B & 67.8 & 4.4 & 70.6 & 89.5 & 87.3 & 78.8 & 38.0 & 63.0 & 57.3 & 70.1 & 42.0 & 81.8 \\
\rowcolor{lightgrey} Llama 2 13B & 54.1 & 1.6 & 55.7 & 67.3 & 83.9 & 74.9 & 38.4 & 45.6 & 41.5 & 28.1 & 23.9 & 81.3 \\
\rowcolor{lightgrey} Mistral Nemo 12B & 66.9 & \textit{n/a} & 69.5 & 85.2 & 85.6 & 81.5 & 39.7 & 69.2 & 54.7 & 62.1 & 36.7 & 84.6 \\
\rowcolor{lightgrey} Qwen 2.5 14B & 72.3 & 16.0 & 79.3 & 94.0 & 94.0 & 80.0 & 37.3 & 51.5 & 71.0 & 83.4 & 52.8 & 79.2 \\
\rowcolor{lightgrey} Qwen 3 14B & 73.6 & \emph{n/c} & 80.7 & 93.4 & 92.3 & 76.4 & 31.8 & 75.0 & 70.3 & 87.3 & 55.7 & 73.2\\
\rowcolor{midgrey} 
\multicolumn{13}{c}{\textbf{Open-weights models 24-70B Parameters}} \\
\rowcolor{lightgrey} Gemma 2 27B & 71.3 & 21.0 & 75.7 & 90.7 & 88.4 & 74.5 & 44.7 & 70.1 & 61.5 & 75.7 & 44.7 & 87.4 \\
\rowcolor{lightgrey} Qwen 2.5 32B & 74.9 & 16.0 & 83.1 & 95.6 & 96.0 & 84.0 & 37.0 & 53.1 & 78.0 & 83.3 & 59.0 & 79.9 \\
\rowcolor{lightgrey} Qwen 3 32B & 68.9 & \emph{n/c} & 83.3 & 94.9 & 93.5 & 79.0 & 31.9 & 67.4 & 72.4 & 34.0 & 60.7 & 72.2 \\
\rowcolor{lightgrey} Mistral Small 24B & 75.2 & n/a & 80.7 & 93.3 & 91.3 & 77.8 & 42.3 & 74.4 & 69.1 & 79.7 & 54.2 & 88.8 \\
\rowcolor{lightgrey} Gemma 3 27B & 74.7 & 23.0 & 79.5 & 93.4 & 88.2 & 75.0 & 45.4 & 73.2 & 69.5 & 80.4 & 52.9 & 89.1 \\
\rowcolor{lightgrey} Llama 3.1 70B & 75.5 & 64.0 & 79.2 & 93.1 & 87.6 & 78.9 & 51.3 & 78.9 & 66.3 & 80.6 & 47.1 & 92.2 \\
\rowcolor{ai2midwhite}\multicolumn{13}{c}{\textbf{Models with \textit{partially available} data}} \\
\rowcolor{ai2offwhite} StableLM 2 12B & 62.2 & 2.9 & 62.4 & 81.9 & 84.5 & 77.7 & 37.6 & 55.5 & 50.9 & 62.0 & 29.3 & 79.9 \\
\rowcolor{ai2offwhite} Zamba 2 7B & 65.2 & \textit{n/c} & 68.5 & 92.2 & 89.4 & 79.6 & 36.5 & 51.7 & 55.5 & 67.2 & 32.8 & 78.8 \\
\rowcolor{ai2midpink}\multicolumn{13}{c}{\textbf{\textit{Fully-open} models}} \\
\rowcolor{ai2lightpink} Amber 7B & 35.2 & 0.5 & 24.7 & 44.9 & 74.5 & 65.5 & 18.7 & 26.1 & 21.8 & 4.8 & 11.7 & 59.3 \\
\rowcolor{ai2lightpink} OLMo 7B & 38.3 & 1.0 & 28.3 & 46.4 & 78.1 & 68.5 & 24.8 & 27.3 & 23.7 & 9.2 & 12.1 & 64.1 \\
\rowcolor{ai2lightpink} MAP Neo 7B & 49.6 & 2.1 & 58.0 & 78.4 & 72.8 & 69.2 & 28.9 & 39.4 & 45.8 & 12.5 & 25.9 & 65.1 \\
\rowcolor{ai2lightpink} OLMo 7B 0424 & 50.7 & 1.0 & 54.3 & 66.9 & 80.1 & 73.6 & 29.6 & 50.0 & 43.9 & 27.7 & 22.1 & 58.8 \\
\rowcolor{ai2lightpink} DCLM 7B & 56.9 & 1.0 & 64.4 & 79.8 & 82.3 & 77.3 & 28.8 & 39.3 & 47.5 & 46.1 & 31.3 & 72.1 \\ %
\rowcolor{ai2lightpink} \textbf{OLMo 2 7B} & \textbf{62.9} & \textbf{1.8} & \textbf{63.7} & \textbf{79.8} & \textbf{83.8} & \textbf{77.2} & \textbf{36.9} & \textbf{60.9} & \textbf{50.4} & \textbf{67.5} & \textbf{31.0} & \textbf{78.0} \\
\rowcolor{ai2lightpink} \textbf{OLMo 2 13B} & \textbf{68.3} & \textbf{4.6} & \textbf{67.5} & \textbf{83.5} & \textbf{86.4} & \textbf{81.5} & \textbf{46.7} & \textbf{70.7} & \textbf{54.2} & \textbf{75.1} & \textbf{35.1} & \textbf{81.9} \\
\rowcolor{ai2lightpink} \textbf{OLMo 2 32B} & \textbf{73.3} & \textbf{13.0} & \textbf{74.9} & \textbf{90.4} & \textbf{89.7} & \textbf{83.0} & \textbf{50.2} & \textbf{74.3} & \textbf{61.0} & \textbf{78.8} & \textbf{46.9} & \textbf{88.0} \\
\bottomrule
\end{tabular}
\end{small}
\vspace{2mm}
  \caption{
  Evaluations comparing \olmotoo to other base models on a {\bf subset of the OLMES suite} (full suite details and results in Appendix~\ref{app:eval-base}).
  Training FLOPs are computed using the approximation from \citet{kaplan2020scaling} and expressed as powers of $10^{23}$.
  We could not estimate compute for any Mistral model~\citep{jiang2023mistral,mistralnemo} because their total training token count is unknown. 
  Training FLOPs for Qwen 3~\citep{yang2025qwen3technicalreport} (concurrent work) and Zamba 2~\citep{zamba2_model} are not reported due to difference in architecture. 
  Qwen 2.5 models~\citep{qwen2.5} are trained on a ``maximum of 18 trillion tokens''; developers have \href{https://web.archive.org/web/20241125024509/https://github.com/QwenLM/Qwen2.5/issues/562}{declined to disclose} exact token counts for each model size. 
  \olmotoo models were {\bf not evaluated} on held-out datasets prior to release; 
  we note that, for other models, we cannot guarantee the same. 
  }
\label{tab:evals_overview}
\end{center}
\end{table}

\paragraph{Base Model Evaluation:} 
\label{sec:base-eval}
We evaluated \olmotoo and other baseline models using the OLMES evaluation suite~\citep{olmes}, which includes a range of benchmark datasets for both multiple-choice and generative tasks, using standardized prompts and in-context examples for few shot predictions. Full descriptions of benchmark tasks in Appendix~\ref{app:eval-base}.
For multiple-choice tasks, we evaluate accuracy; for generative tasks, we evaluate F1 to account for partial matches.
Additionally, to avoid overfitting our recipe to these benchmarks, we maintained a \textbf{held-out suite of tasks} which were not used for model development decisions; we advocate for a standard practice of declaring development vs held-out evaluation tasks for model developers.\footnote{GSM8k~\citep{cobbe2021trainingverifierssolvemath} was only partially held-out, as we subsampled 200 of 1319 GSM8k examples for mid-training data development when we noticed poor math capabilities after pretraining; we call this dev set GSM$^*$. The remaining 1119 GSM8k examples we reserve as held-out and report final performance on them only.}

Table~\ref{tab:evals_overview} contains overall results. 
We find our \textbf{\olmotoo models are competitive with the best open-weights models} of comparable size, despite \olmotoo requiring \textbf{far fewer training FLOPs} (see Figure~\ref{fig:pareto}) and maintaining \textbf{full openness (e.g. training data)}.
We find that gains observed on development metrics largely translate to our unseen evaluation suite, indicative of a generalizable training recipe.

Overall, we find that gains observed on development metrics largely translate to our unseen evaluation suite. 
Of course, we have no guarantee that tasks we consider unseen during development of \olmotoo are not part of the development set of other models we compare. 
Nevertheless, we think it should be \textbf{standard practice} for model developers to keep a subset of evaluation tasks unseen and to declare which these are, in technical reports.
Further, we encourage other open-weight \textbf{model developers to clearly state which tasks are being monitored} during model development.

\paragraph{Post-Training Recipe and Evaluation}
\label{sec:adapt-eval}
For post-training we apply our \tulu~\citep{lambert2024tulu3} recipe with supervised finetuning, on-policy preference tuning, and reinforcement learning with verifiable rewards (RLVR).\footnote{
We made minor modifications to the preference data to use generations from permissively-licensed models and added a multi-stage RLVR training protocol to optimize final performance, but otherwise followed the recipe as-is. 
}
The resulting models---\olmotooinstruct---are evaluated in Table~\ref{tab:instruct_results} on general and precise instruction following, math, knowledge reasoning, and safety tasks from the same evaluation suite used by \citet{lambert2024tulu3}.
Full descriptions of benchmark tasks in Appendix~\ref{app:eval-post}.

Table~\ref{tab:instruct_results} contains downstream results.
We find \textbf{\olmotooinstruct models are competitive with the best instruction-tuned open-weights models and even some popular proprietary models}.
This shows the usefulness of \olmotoo as a powerful base model that serves as an excellent starting point for fully open post-training research.
Full post training details are in Section~\S\ref{sec:post-train}.

\begin{table}[!htbp]
\centering
\setlength\tabcolsep{3pt}
{\small
\begin{tabular}{lcccccccccccc}
\toprule
    \textbf{Instruct Model} & 
    {\textbf{\fontsize{8}{8}\selectfont~Avg}} & 
    {\textbf{\fontsize{8}{8}\selectfont~FLOP$\times 10^{23}$}} & 
    {\textbf{\fontsize{8}{8}\selectfont~AE2}} & 
    {\textbf{\fontsize{8}{8}\selectfont~BBH}} & 
    {\textbf{\fontsize{8}{8}\selectfont~DROP}} & 
    {\textbf{\fontsize{8}{8}\selectfont~GSM8K}} & 
    {\textbf{\fontsize{8}{8}\selectfont~IFE}} & 
    {\textbf{\fontsize{8}{8}\selectfont~MATH}} & 
    {\textbf{\fontsize{8}{8}\selectfont~MMLU}} & 
    {\textbf{\fontsize{8}{8}\selectfont~Safety}} & 
    {\textbf{\fontsize{8}{8}\selectfont~PQA}} & 
    {\textbf{\fontsize{8}{8}\selectfont~TQA}}
\\
\midrule
\rowcolor{midgrey}\multicolumn{13}{c}{\textbf{Closed API models}} \\
\rowcolor{lightgrey} GPT-3.5 Turbo 0125 & 60.5 & n/a & 38.7 & 66.6 & 70.2 & 74.3 & 66.9 & 41.2 & 70.2 & 69.1 & 45.0 & 62.9 \\
\rowcolor{lightgrey} GPT 4o Mini 0724 & 65.7 & n/a & 49.7 & 65.9 & 36.3 & 83.0 & 83.5 & 67.9 & 82.2 & 84.9 & 39.0 & 64.8 \\
\rowcolor{midgrey}\multicolumn{13}{c}{\textbf{Open weights models 1–1.7B Parameters}} \\

\rowcolor{lightgrey}
Gemma 3 1B & 38.3 & 0.12 & 20.4 & 39.4 & 25.1 & 35.0 & 60.6 & 40.3 & 38.9 & 70.2 & 9.6 & 43.8 \\

\rowcolor{lightgrey}
Llama 3.2 1B & 39.3 & 0.67 & 10.1 & 40.2 & 32.2 & 45.4 & 54.0 & 21.6 & 46.7 & 87.2 & 13.8 & 41.5 \\

\rowcolor{lightgrey}
Qwen 2.5 1.5B & 41.7 & 1.7 & 7.4 & 45.8 & 13.4 & 66.2 & 44.2 & 40.6 & 59.7 & 77.6 & 15.5 & 46.5 \\

\rowcolor{midgrey}\multicolumn{13}{c}{\textbf{Open weights models 7-14B Parameters}} \\
\rowcolor{lightgrey}Ministral 8B 2410 & 53.5 & n/a & 31.4 & 70.8 & 56.2 & 80.0   & 56.4 & 40.0   & 68.5 & 56.2 & 20.2 & 55.5 \\
\rowcolor{lightgrey}Llama 3.1 8B & 59.1 & 7.2 & 25.8 & {71.9} & 61.7 & 83.4 & 80.6 & 42.5 & 71.3 & 70.2 & 28.4 & 55.1 \\
\rowcolor{lightgrey}Tulu 3 8B & 60.7 & 7.2 & 34.0 & 69.0 & 62.6 & {87.6} & {82.4} & 43.7 & 68.2 & 75.4 & {29.1} & 55.0 \\
\rowcolor{lightgrey}Qwen 2.5 7B & 61.6 & 8.2 & 29.7 & 70.2 & 54.4 & 83.8 & 74.7 & 69.9 & 76.6 & 75.0   & 18.1 & 63.1 \\
\rowcolor{lightgrey}Gemma 2 9B & 58.1 & 4.4 & {43.7} & 64.9   & 58.8 & 79.7 & 69.9 & 29.8 & 69.1 & 75.5 & 28.3 & 61.4 \\
\rowcolor{lightgrey}Qwen 2.5 14B & 65.3 & 16.0 & 34.6 & 78.4 & 50.5 & 83.9 & 82.4 & {70.6} & {81.1} & {79.3} & 21.1 & {70.8} \\
\rowcolor{midgrey}\multicolumn{13}{c}{\textbf{Open weights models 24-32B Parameters}} \\
\rowcolor{lightgrey} Gemma 2 27B & 61.3 & 21.0 & 49.0 & 72.7 & 67.5 & 80.7 & 63.2 & 35.1 & 70.7 & 75.9 & 33.9 & 64.6 \\
\rowcolor{lightgrey} Qwen 2.5 32B & 68.1 & 35.0 & 39.1 & 82.3 & 48.3 & 87.5 & 82.4 & 77.9 & 84.7 & 82.4 & 26.1 & 70.6 \\
\rowcolor{lightgrey} Mistral Small 24B & 67.5 & n/a & 43.2 & 80.1 & 78.5 & 87.2 & 77.3 & 65.9 & 83.7 & 66.5 & 24.4 & 68.1 \\
\rowcolor{lightgrey} Qwen QwQ 32B & - & 35.0 & 82.4 & 89.6 & 54.7 & 95.5 & 85.8 & 98.1 & 88.4 & 69.9 & - & - \\
\rowcolor{lightgrey} Gemma 3 27B & 71.3 & 23.0 & 63.4 & 83.7 & 69.2 & 91.1 & 83.4 & 76.2 & 81.8 & 69.1 & 30.9 & 63.9 \\
\rowcolor{midgrey}\multicolumn{13}{c}{\textbf{Open weights models \char126 70B Parameters}} \\
\rowcolor{lightgrey} Qwen 2.5 72B & 68.8 & 79.0 & 47.7 & 80.4 & 34.2 & 89.5 & 87.6 & 75.9 & 85.5 & 87.0 & 30.6 & 69.9 \\
\rowcolor{lightgrey} Llama 3.1 70B & 70.7 & 64.0 & 32.9 & 83.0 & 77.0 & 94.5 & 88.0 & 56.2 & 85.2 & 76.4 & 46.5 & 66.8 \\
\rowcolor{lightgrey} Llama 3.3 70B & 72.7 & 64.0 & 36.5 & 85.8 & 78.0 & 93.6 & 90.8 & 71.8 & 85.9 & 70.4 & 48.2 & 66.1 \\
\rowcolor{ai2midpink}\multicolumn{13}{c}{\textbf{Fully-open Language Models}} \\
\rowcolor{ai2lightpink} OLMo 1B 0724 & 24.4 & 0.22 & 2.4 & 29.9 & 27.9 & 10.8 & 25.3 & 2.2 & 36.6 & 52.0 & 12.1 & 44.3 \\
\rowcolor{ai2lightpink}
SmolLM2 1.7B & 34.2 & 1.1 & 5.8 & 39.8 & 30.9 & 45.3 & 51.6 & 20.3 & 34.3 & 52.4 & 16.4 & 45.3 \\
\rowcolor{ai2lightpink} OLMo 7B 0424 & 33.1 & 1.0 & 8.5 & 34.4 & 47.9 & 23.2 & 39.2 & 5.2 & 48.9 & 49.3 & 18.9 & 55.2 \\
\rowcolor{ai2lightpink} \textbf{OLMo 2 1B} & \textbf{42.7} & \textbf{0.35} & \textbf{9.1} & \textbf{35.0} & \textbf{34.6} & \textbf{68.3} & \textbf{70.1} & \textbf{20.7} & \textbf{40.0} & \textbf{87.6} & \textbf{12.9} & \textbf{48.7} \\
 \rowcolor{ai2lightpink} \textbf{OLMo 2 7B} & \textbf{56.5} & \textbf{1.8} & \textbf{29.1} & \textbf{51.4} & \textbf{60.5} & \textbf{85.1} & \textbf{72.3} & \textbf{32.5} & \textbf{61.3} & \textbf{93.3} & \textbf{23.2} & \textbf{56.5} \\
\rowcolor{ai2lightpink} \textbf{OLMo 2 13B} & \textbf{63.5} & \textbf{4.6} & \textbf{39.5} & \textbf{63.0} & \textbf{71.5} & \textbf{87.4} & \textbf{82.6} & \textbf{39.2} & \textbf{68.5} & \textbf{89.7} & \textbf{28.8} & \textbf{64.3} \\
\rowcolor{ai2lightpink} \textbf{OLMo 2 32B} & \textbf{68.8} & \textbf{13.0} & \textbf{42.8} & \textbf{70.6} & \textbf{78.0} & \textbf{87.6} & \textbf{85.6} & \textbf{49.7} & \textbf{77.3} & \textbf{85.9} & \textbf{37.5} & \textbf{73.2} \\
\bottomrule
\end{tabular}
}
\caption{The results for OLMo 2 Instruct at 1B, 7B, 13B, and 32B relative to peer open weight models.
The following evaluation names are abbreviated: Avg -- Average, AE2 -- AlpacaEval 2, BBH -- BigBenchHard, IFE -- IFEval, PQA -- PopQA, TQA -- TruthfulQA. All models in this table are the instruction tuned variants.
For Qwen QwQ 32B, PopQA and TruthfulQA had challenges with answer extraction, where the model would return the answer within the \texttt{<think>} tokens, so we did not report a score.
For Qwen QwQ 32B we conducted evaluations by removing the thinking tokens and grading the following answer. 
It was evaluated with their recommended sampling parameters (32K context length, 0.6 temperature, sampling, top$_p$ 0.95, min$_p$ 0, top$_k$ 30) in the model card for all evaluations except safety, which just had a shorter context length of 8K tokens. Multiple choice evaluations, PopQA and TruthfulQA had challenges with answer extraction, where the model would return the answer within the <think> tokens, so we did not report a score. Even outside of extraction issues, the very long context generation of reasoning models has caused challenges to many pieces of open evaluation tooling, which we need to improve.
}
\label{tab:instruct_results}
\end{table}

\section{Deep Dive: \diveStability}
\label{sec:stability}

While \olmoapril achieved performance within expected ranges for its compute budget, the training dynamics were characterized by a couple  of concerns:
\begin{itemize}
    \item \textbf{Sudden spikes} in the loss, and more frequently, in the gradient norm during training. In experiments, we found that increasing model size increased the frequency of spikes.
    Furthermore, our experiments revealed that more dramatic spikes in gradient norm often preceded training loss spikes.
    \item \textbf{Slow growth} in the magnitude of the gradient norm over the training run. This was correlated with increasing frequency of spikes in the gradient norm (and training loss).
\end{itemize}

\begin{figure}[h]
    \centering
    \includegraphics[width=0.8\textwidth]{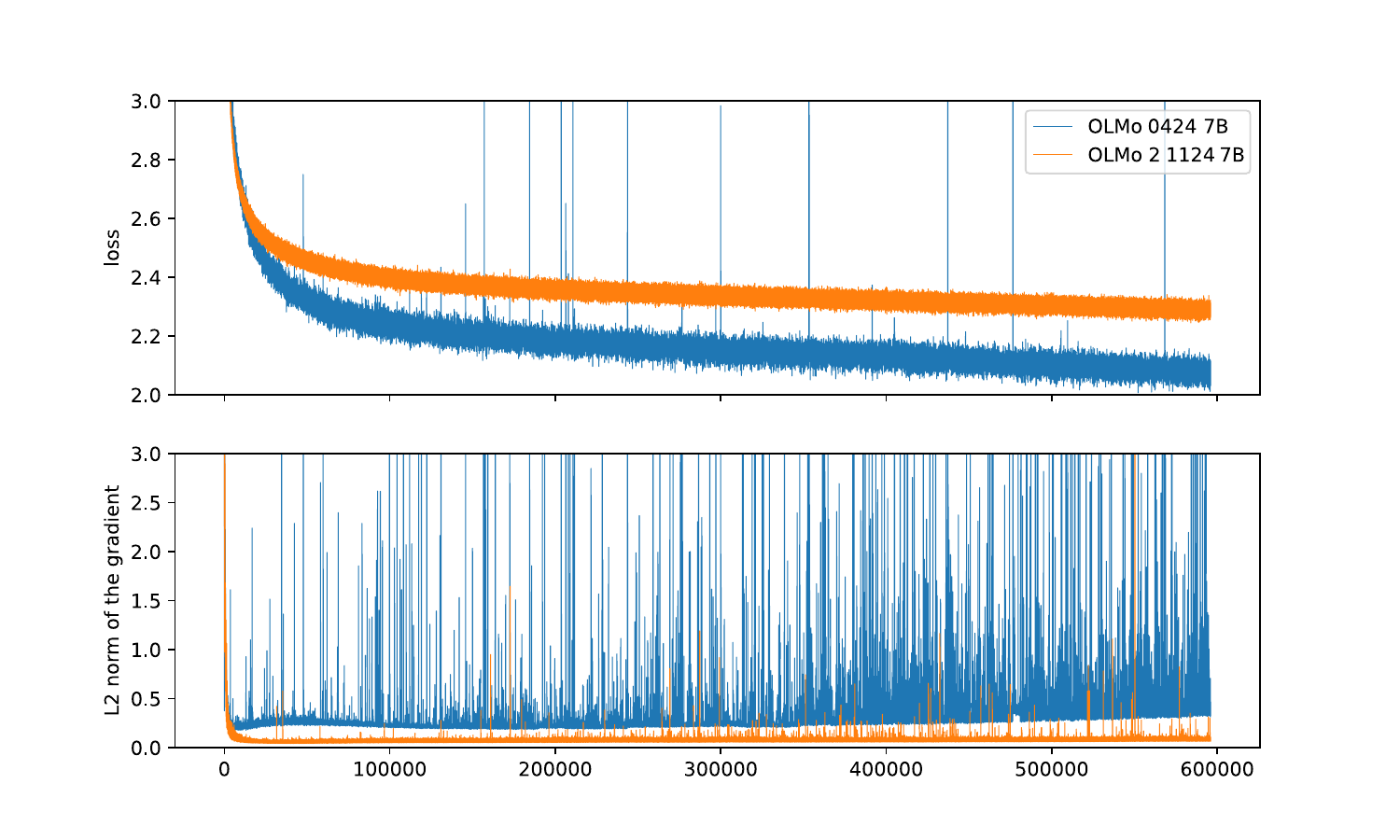}
    \caption{Training loss and gradient norm curves (over training steps) for \olmoapril and \olmotoo. 
    The \olmoapril training run was characterized by frequent loss spikes (top), often preceded by more frequent spikes in the gradient norm, which grew over time (bottom).
    We note that overall training loss for \olmotoo is higher because the underlying training data changed between the runs.} \label{fig:mitchishvpeteish}
\end{figure}

Ultimately, a combination of these issues would lead to training divergence, making training at larger scales impossible.
This situation motivated our training stability investigation into the causes of these issues and their mitigations.
Figure~\ref{fig:mitchishvpeteish} shows our training curves before and after implementing our mitigations, which we summarize below:\nopagebreak  %
\begin{itemize}
    \item  \textbf{Repeated n-grams:} We filter pretraining data to remove repeated n-grams in pretraining data, as they can lead to loss spikes~(\S\ref{sec:datafilter}).
    \item \textbf{Initialization:} We switch from scaled initialization~\citep{Zhang2019ImprovingDT} to initializing all parameters with a mean of 0 and a standard deviation of 0.02 (\S\ref{sec:initialization}).
    \item \textbf{RMSNorm:} We use the RMSNorm variant of LayerNorm to normalize activations instead of non-parametric LayerNorm~(\S\ref{sec:qknorm}).
    \item \textbf{Reordered norm:} We normalize the outputs to the attention and feed-forward (MLP) layers within each transformer block instead of the inputs~(\S\ref{sec:qknorm}).
    \item \textbf{QK-norm:} We normalize the key and query projections with RMSNorm before calculating attention~(\S\ref{sec:qknorm}).
    \item \textbf{Z-Loss:} We adopt z-loss regularization, a regularization term that keeps final output logits from growing too large~(\S\ref{sec:zloss}).
    \item \textbf{Weight decay:} We exclude embeddings from weight decay~(\S\ref{sec:embeddingwd}).
    \item \textbf{$\epsilon$ in AdamW:} We lower the $\epsilon$ of AdamW from $10^{-5}$ to $10^{-8}$~(\S\ref{sec:epsilon}).
\end{itemize}

In the following, we will discuss the experiments and results that led us to these interventions.  
We compare our revised strategies with \olmoapril, the most recent version of \olmo with fully-open model weights, data, and documentation.

\subsection{Repeated n-Grams}
\label{sec:datafilter}

Data can be a cause of both gradient norm and loss spikes.
When investigating training batches at which spikes occurred, we found a high prevalence of instances containing long, repeated n-gram sequences. Here are three examples of such sequences: \\

\begin{verbatim}

      g4ODg4ODg4ODg4ODg4ODg4ODg4ODg4ODg4ODg4ODg4ODg4ODg4ODg4OD...
      [\n  365, 0, 667, 1000, 1000, 667, 667, 667, 667, 667, ...
      ' 255, 255, 255, 255, 255, 255, 255, 255, \n255, 255, ...
      
\end{verbatim}

In a series of experiments, we found these sequences are often associated with spikes, though we note that this relationship is not deterministic:
\begin{itemize}
    \item The same n-gram sequence may spike for a larger model but not for a smaller model trained on the same data.
    \item The same n-gram sequence may spike for one data training ordering, but not after the data is reshuffled.
    \item The same n-gram sequence associated with a spike can also be found elsewhere in training batches that did not spike.
\end{itemize}

\begin{figure}[h]
    \centering
    \includegraphics[width=0.49\textwidth]{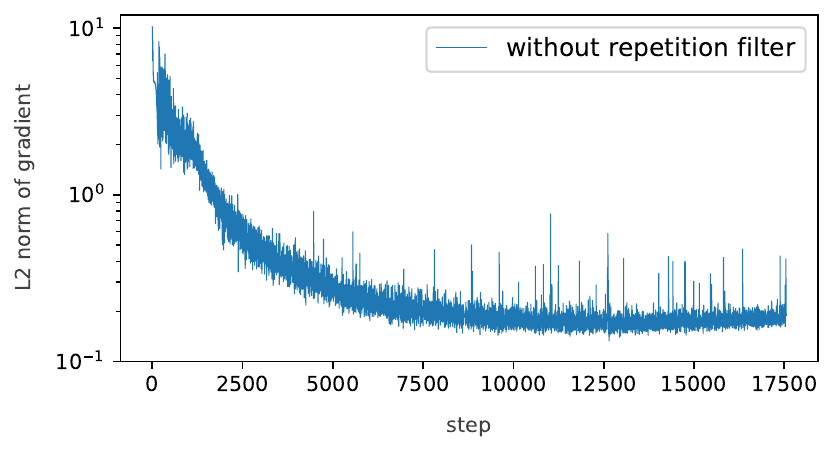}
    \includegraphics[width=0.49\textwidth]{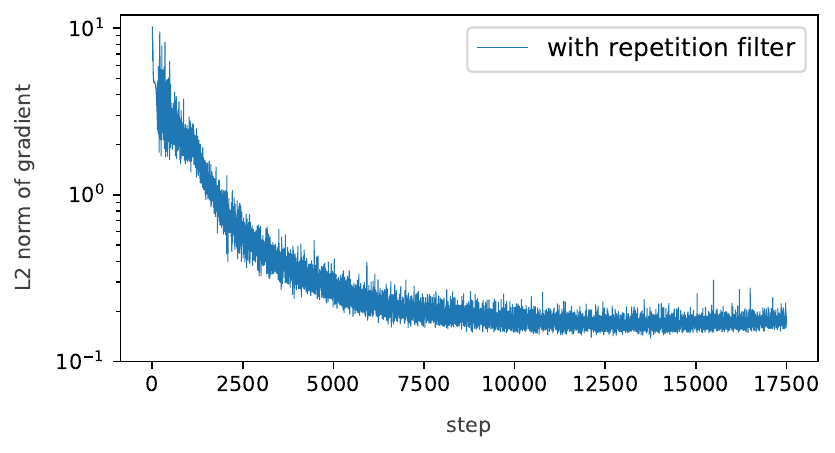}
    \caption{Comparison of the gradient norm for two runs, one without n-gram filter, and one with.
    Ignoring long repetitive sequences of n-grams eliminates many spikes.}
    \label{fig:datafixspikes}
\end{figure}

Nevertheless, we have found evidence that broad removal of such sequences across training decreases the frequency of spikes, on average.
At data curation time (Section~\S\ref{sec:pretrain-data}), we apply a filter that removes all documents with a sequence of 32 or more repeated n-grams, where an n-gram is any span of 1 to 13 tokens. 
We also implement an additional safeguard in the trainer that detects these sequences during data loading and masks them when computing the loss.
Figure~\ref{fig:datafixspikes} shows the effect of masking the loss of input sequences containing repeated n-grams. 
This intervention results in a clear mitigation---though not complete elimination---of gradient spikes. 
It had no effect on the slow growth in gradient norm.

\subsection{Model Initialization}
\label{sec:initialization}

\begin{figure}[t]
    \centering
    \includegraphics[width=0.9\textwidth]{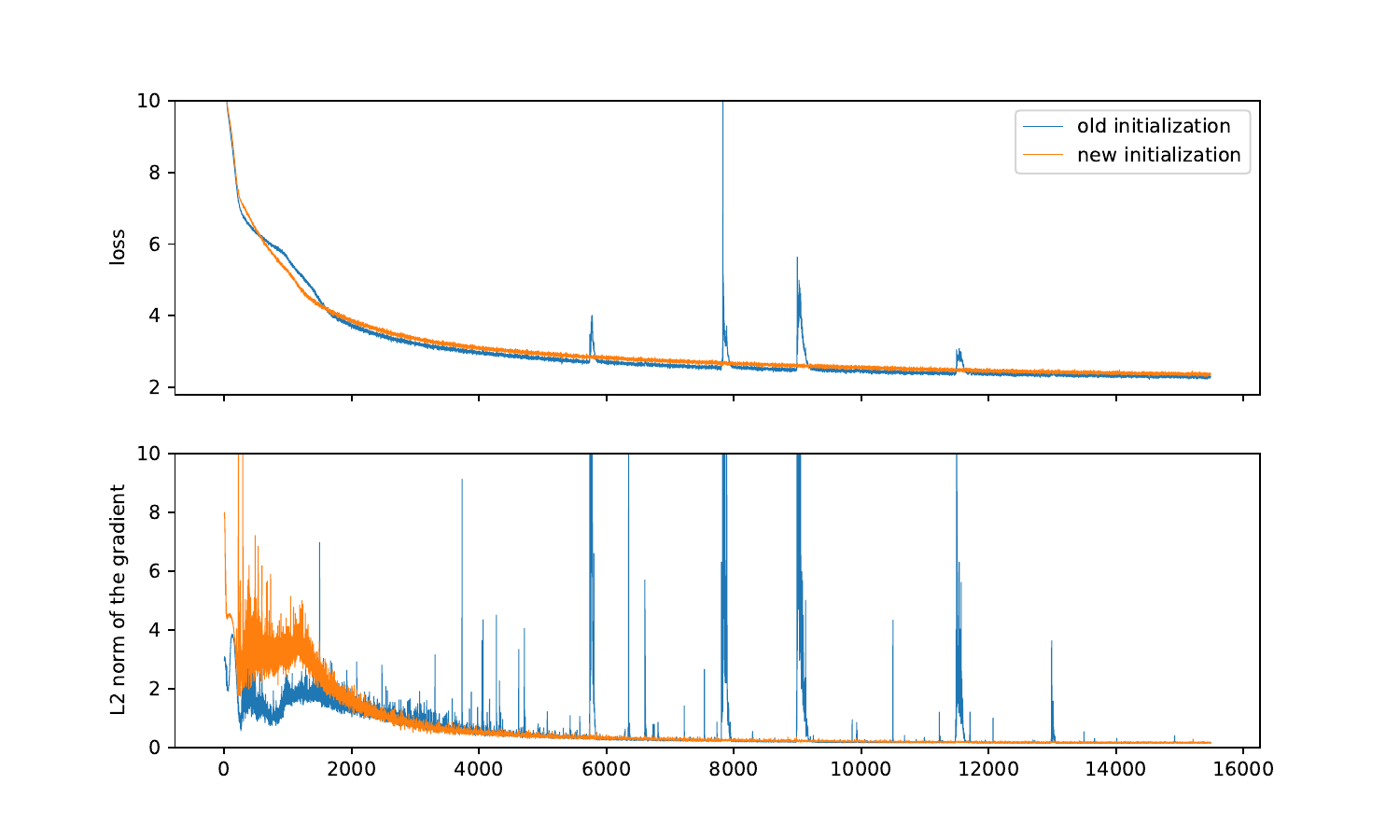}
    \caption{In our test setting, the \olmoapril initialization scheme shows instabilities quickly, while \olmotoo stays stable.}
    \label{fig:inits}
\end{figure} 

\newcommand\dmodel{d_{\mathrm{model}}}

Figure~\ref{fig:inits} shows the improvement to training stability from \olmotoo's initialization scheme.
In \olmotoo, we  initialize every parameter from a normal distribution with a mean of 0 and a standard deviation of 0.02.
In contrast, \olmoapril's initialization, first suggested in~\citet{Zhang2019ImprovingDT} and implemented by~\citet{openlm}, scaled input projections by $1 / \sqrt{\dmodel}$, and output projections by $1 / \sqrt{2 \cdot \dmodel \cdot \mathrm{layer\_idx}}$ at every layer.
In other words, later layers were initialized to smaller values.

We perform several analyses to study the impact of initialization, showing that \olmotoo's initialization is superior to \olmoapril initialization.  Our empirical analysis suggests it better preserves the scale of activations and gradients across layers, allowing deep models to be trained more stably, and it exhibits properties associated with hyperparameter transfer across models of different widths.
These two properties together give us confidence that deep models will train stably and that the initialization hyperparameters of our smaller models could transfer to larger scales.

\begin{figure}[t]
    \centering
    \includegraphics[width=0.4\textwidth]{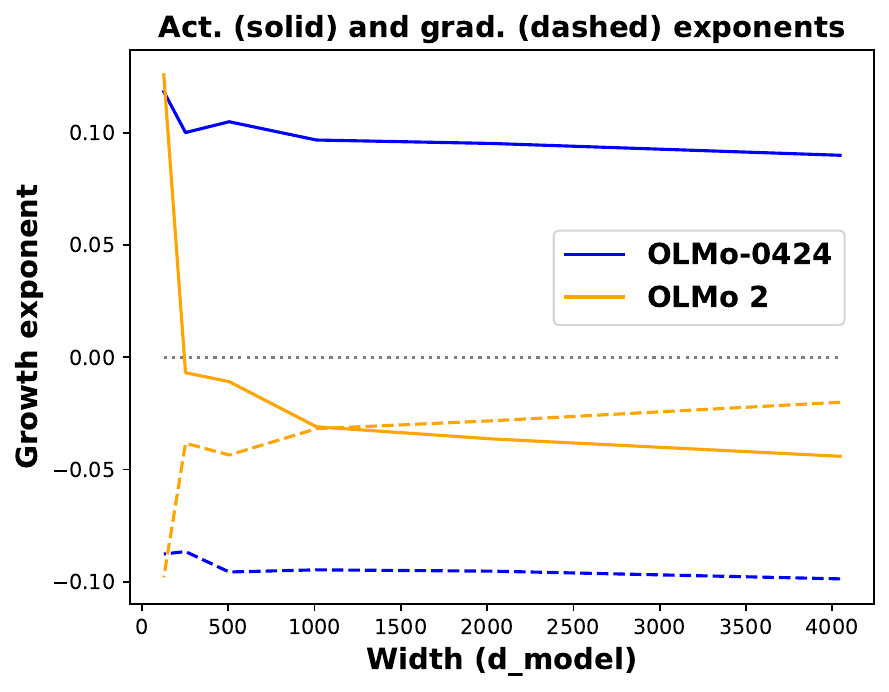}
    \caption{Across widths, growth exponents for the \olmotoo initialization are closer to 0 compared to the \olmoapril initialization, which suggests deeper models will train more stably.}
    \label{fig:lyupanov}
\end{figure}

\paragraph{Gradient and activation growth} A fundamental concern for training deep networks is ensuring that the activations and gradients do not blow up or vanish across layers, causing learning to become unstable or stagnate.
Rather, we want the scale of the activations and gradients to remain roughly the same from layer to layer. 
Inspired by recent related work~\citep{cowsik2024geometric}, we evaluate different candidate initializations in terms of how they affect the 2-norm of the activations and gradients across layers.
Concretely, we randomly initialize a model, pass 50 random documents from The Pile~\citep{pile} through it, and collect the activations and gradients (of loss with respect to the activations) at the initial and final layers (ignoring embeddings). We then average these tensors across documents and time steps to get vectors $\pmb v$ at the initial layer and $\pmb{v'}$ at the final layer, both of length $\dmodel$.
Finally, we compute the following measure of expansion or contraction across layers, which we call the \emph{growth exponent}:
\begin{equation*}
    \lambda = \frac{1}{n_\mathrm{layers}} \log \left( \frac{\norm{\pmb{v'}}}{\norm{\pmb v}} \right)
\end{equation*}
We compute $\lambda$ for both the activations and gradients. Ideally, both $\lambda$'s remain near 0, indicating that the activations and gradients do not explode or vanish across layers.
Figure~\ref{fig:lyupanov} plots the growth exponents for different randomly initialized models as a function of their widths (4096 corresponds to a full 7B model).
Crucially, the growth exponent for \olmotoo is closer to 0 than for \olmoapril across model widths. 
This suggests the \olmotoo initialization will be more stable when training deep models in low precision, as both the activations and the gradients are more resistant to exploding or vanishing across layers compared to the original \olmoapril initialization.

\begin{figure}[t]
    \centering
    \includegraphics[width=0.4\linewidth]{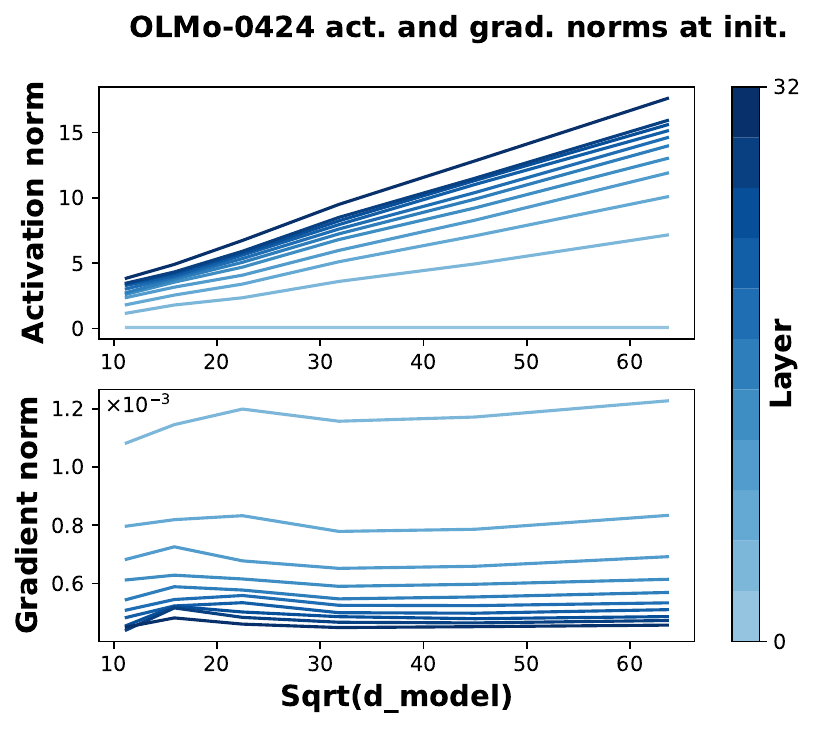}
    \includegraphics[width=0.4\linewidth]{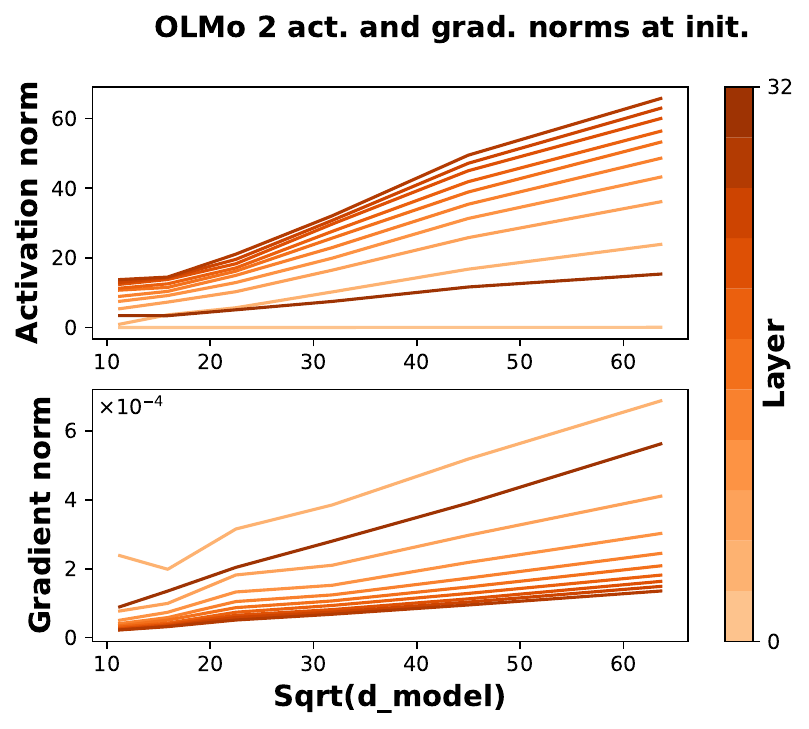}
    \caption{Activation and gradient norms vs. $\sqrt{d_\mathrm{model}}$ for the \olmoapril and \olmotoo initializations. Crucially, the gradient norms for \olmotoo positively correlate with $\sqrt{d_\mathrm{model}}$, which they did not for the \olmoapril initialization. This suggests the \olmotoo initialization will show better hyperparameter transfer across widths \citep{yang2024spectral}.}
    \label{fig:width-scaling}
\end{figure}

\paragraph{Hyperparameter transfer across width} Another appealing property of the new initialization is that it scales the activation and gradient norms with width ($\dmodel$) in a way that has been argued theoretically to be important for hyperparameter transfer across different widths. Specifically, \citet{yang2024spectral} suggest that a sufficient condition for hyperparameter transfer across width is that the magnitude of each activation scalar value and its update (learning rate times gradient) remain fixed as width increases.
Equivalently, the norms of the activations and their update vectors should positively correlate with $\sqrt{\dmodel}$. We plot the activation and gradient norms at initialization against $\sqrt{\dmodel}$ in Figure~\ref{fig:width-scaling}.
Crucially, the gradient norm is more positively correlated with $\sqrt{\dmodel}$ for \olmotoo compared to \olmoapril.
Combined with \citet{yang2024spectral}, this suggests that, with an initial learning rate independent of model width, the new \olmotoo initialization will transfer better across different model widths compared to the \olmoapril initialization.

\paragraph{Spike score}
Since fast spikes are difficult to understand with contemporary graphing tools, we compute a \textit{spike score} as an objective measure.
Concretely, we define the spike score as the percentage of values in a time series that are at least seven standard deviations away from a rolling average of the last $1,000$ values\footnote{
    Spike score is conceptually similar to spike mitigation proposed by \citet{karpathy2024spikes}.
}.
We use spike score primarily on training loss and L2 norm of the gradient, but the measure can be computed on any time series.

\paragraph{Empirical results}
To experiment with model initialization, we first create a baseline run that reproduces spikes quickly.
We do so by mainly reducing the warmup period.
The effect was immediate and dramatic (Figure~\ref{fig:inits}), and persists across model scales and token counts.
In our ablation, the new initialization had no loss spikes, and the spike score for the L2 norm of the gradient went from $0.40$ to $0.03$.
The new initialization converges slightly slower; 
we make up for this difference by improving other hyperparameter settings (Section~\S\ref{sec:hp_improve}).

\subsection{Architecture Improvements}

\subsubsection{Nonparametric layer norm and RMSNorm}

\olmotoo uses RMSNorm, which is standard in most transformer implementations.
\olmoapril used a nonparametric layer norm for performance and to work around bugs in the libraries we were using, but by the time we developed \olmotoo, the bugs were no longer an issue, the hardware was faster, and we wanted to settle on a safe approach.
Our ablations show no difference between the two, so we switch back to RMSNorm.

\subsubsection{Reordered norm and QK-norm}
\label{sec:qknorm}

\begin{figure}[h]
    \centering
    \includegraphics[width=0.9\textwidth]{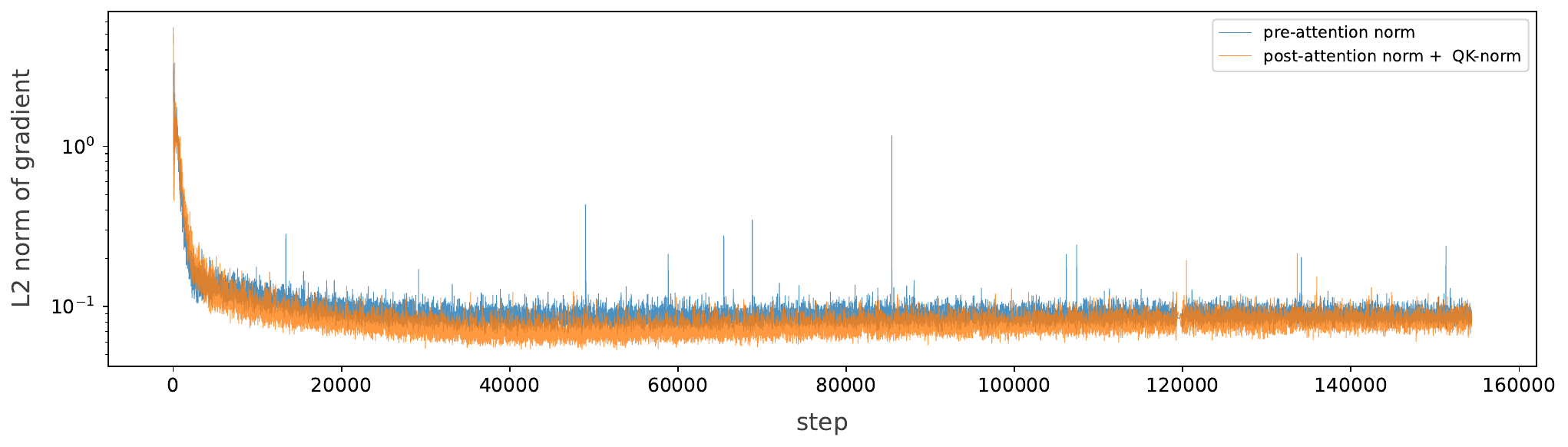}
    \caption{Applying layer norm after the attention and feedforward layers along with a QK-norm improves stability compared to a more standard pre-attention layer norm. These changes reduce the spike score of the gradients from 0.108 to 0.069 when applied together.}
    \label{fig:qk_norm_reorder}
\end{figure}

\begin{samepage}

Figure~\ref{fig:qk_norm_reorder} shows the effect of applying the layer normalization to the \emph{outputs} of the MLP and attention blocks instead of the inputs.
We further apply another normalization, also RMSNorm, to the queries and keys in the attention block.
In isolation, neither of these changes yield good results, but together they improve both the growth and the spikiness of the L2 norm of the gradient.
The following table summarizes the difference in the location of the layer normalization:

\begin{center}
\begin{tabular}{ll}
\toprule
\textbf{\olmoapril} & \textbf{\olmotoo} \\
\midrule
  $ \pmb{h} := \pmb{x} + \text{Attention}(\text{LN}(\pmb{x})) $ & $ \pmb{h} := \pmb{x} + \text{RMSNorm}(\text{Attention}(\pmb{x})) $ \\
  $ \pmb{h}_{\text{out}} := \pmb{h} + \text{MLP}(\text{LN}(\pmb{h})) $ & $ \pmb{h}_{\text{out}} := \pmb{h} + \text{RMSNorm}(\text{MLP}(\pmb{h})) $ \\
\bottomrule
\end{tabular}

$\pmb{x}$ is the input to the layer, $\pmb{h}$ is an intermediate hidden state, and $\pmb{h}_{\text{out}}$ is the output.
\end{center}

\end{samepage}

\citet{swin2} first introduced layer norm the idea of reordering layer norm. 
It was subsequently picked up by~\citet{chameleon}. 
QK-norm was first developed in~\citet{scalingvisiontransformers22}. 

\subsubsection{Z-Loss}
\label{sec:zloss}

Following~\citet{palm},~\citet{chameleon}, and~\citet{mitch}, we apply z-loss regularization by adding $10^{-4} \cdot \log^2 Z$ to our loss function, where $Z$ is the denominator in the softmax over the logits.
This discourages the activations in the final softmax from growing too large, improving the stability of the model.

\begin{figure}[h]
    \centering
    \includegraphics[width=0.9\textwidth]{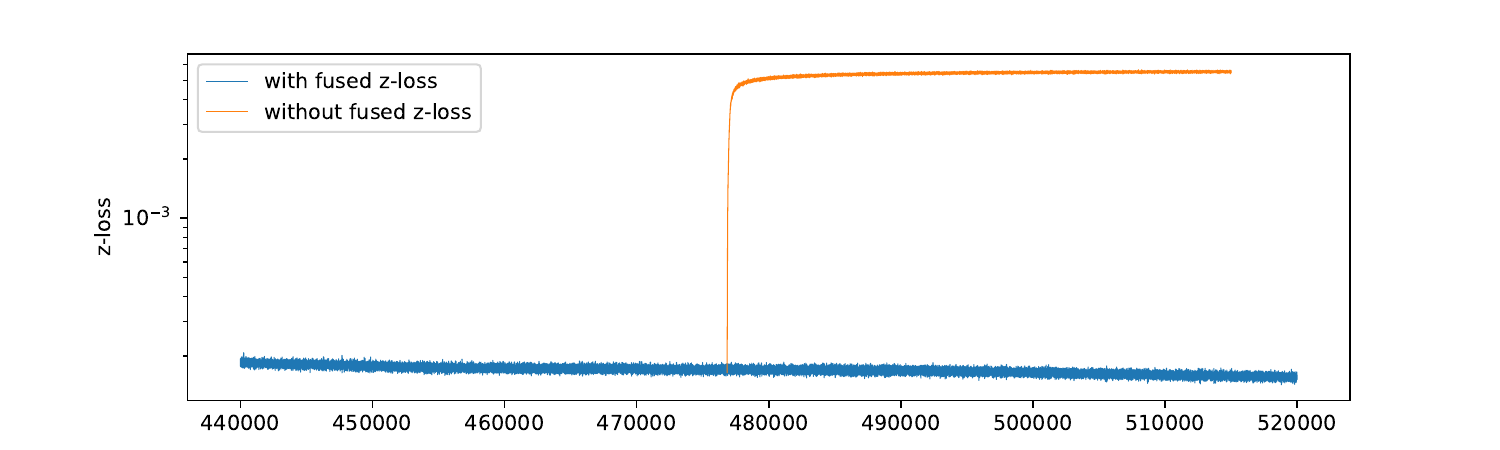}
    \caption{Flash Attention's implementation of z-loss does not match a manual implementation in PyTorch. While the forward pass produces the same number, differences in the backwards pass cause the curves to diverge.}
    \label{fig:zloss}
\end{figure}

Figure~\ref{fig:zloss} shows a stark difference between the z-loss implementation of the popular Flash Attention library~\citep{dao2023flashattention2}, and an implementation using only Python primitives.
Apart from the attention mechanism it is known for, Flash Attention also provides an optimized implementation of cross-entropy loss, which includes a version of z-loss.
To retain flexibility in settings that are not compatible with Flash Attention, we have a separate implementation written in PyTorch.
Both implementations produce the same result in the forward pass, but exhibit different behavior in the backward pass.
We suspect the root cause lies in differences in precision.
In our experiments, this does not affect cross entropy loss during training, or the model's performance on downstream tasks.
However, out of an abundance of caution we abandon the fork with custom z-loss implementation and re-train from the original point of divergence.
During a training run we cannot switch implementations safely, so we avoid doing so as much as possible.  

\subsection{Hyperparameter Improvements}
\label{sec:hp_improve}

\subsubsection{$\mathbf{\epsilon}$ in AdamW}
\label{sec:epsilon}

\begin{figure}[h]
    \centering
    \includegraphics[width=0.99\textwidth]{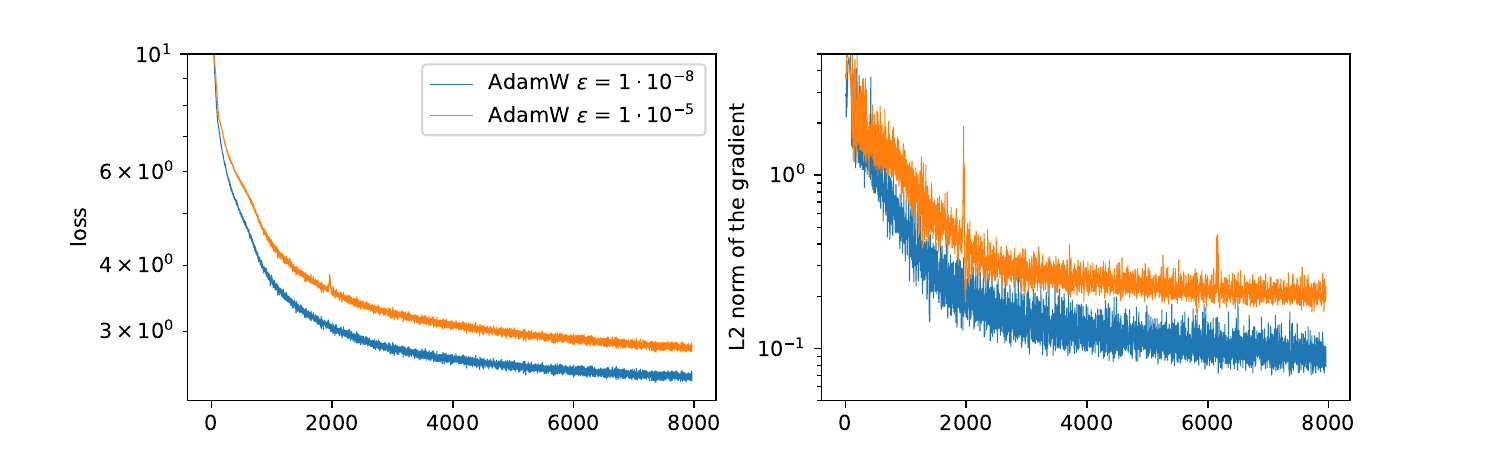}
    \caption{Setting AdamW's $\epsilon$ to $10^{-8}$ lowers and stabilizes the norm of the gradient early in training. The training loss also improves faster. This trend continues even with runs that are longer than what is shown here.}
    \label{fig:adameps}
\end{figure}

Figure~\ref{fig:adameps} shows the result of decreasing the AdamW $\epsilon$ from $10^{-5}$ to $10^{-8}$.
$10^{-8}$ is the default in PyTorch, but some popular LM training code bases come with a default of $10^{-5}$.
The lower value allows for larger updates early in training, and helps the model learn faster during a period where we've typically seen a lot of instability.
As a result, the gradient norm settles much more quickly and remains permanently lower.

\subsubsection{Weight decay on embeddings}
\label{sec:embeddingwd}

\begin{figure}[h]
    \centering
    \includegraphics[width=0.49\textwidth]{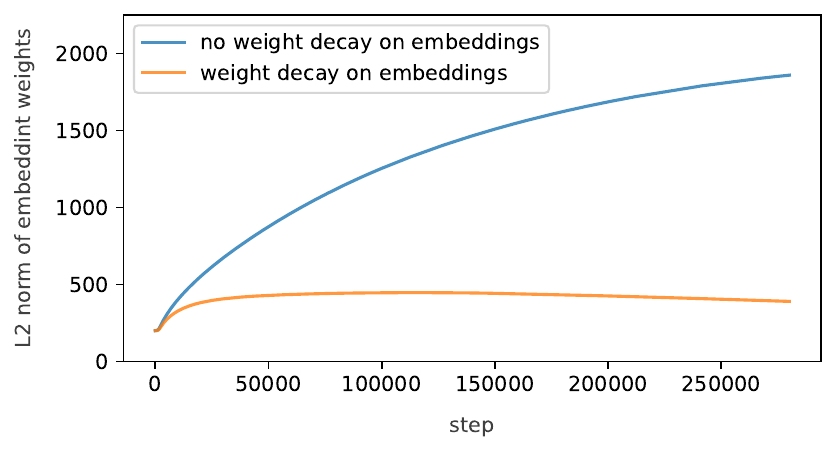}
    \includegraphics[width=0.49\textwidth]{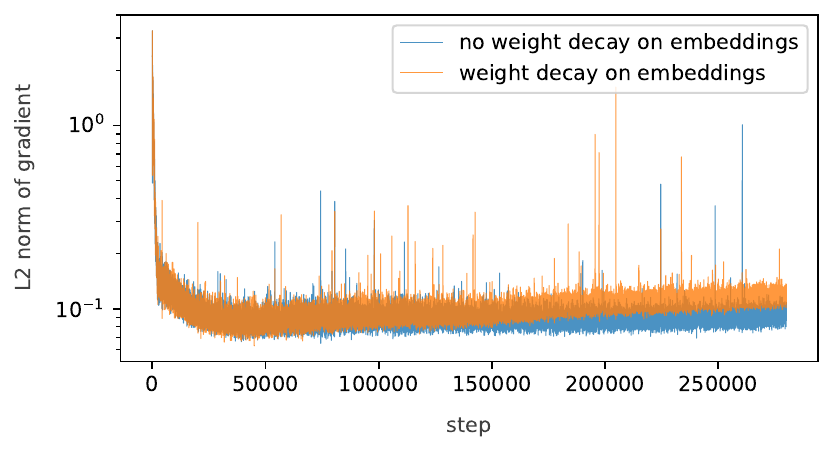}
    \caption{Weight decay applied to token embeddings leads to a gradual decrease in the embedding norm and a corresponding increase in the gradient norm. Decaying embeddings also has a modest negative impact on stability, producing more spikes than a comparable run without (spike scores of 0.16 and 0.092 respectively).}
    \label{fig:embed_wd}
\end{figure}

Figure~\ref{fig:embed_wd} shows the change in training dynamics following a decision to exclude weight decay for embeddings.
\olmo uses a standard formulation of weight decay, where every parameter is multiplied by $1 - (0.1 \cdot lr)$ at every step.
This regularization term discourages parameters from growing too large, but in the case of token embeddings it overshoots the mark and results in very small embeddings.
As discussed by \citet{spikenomore}, small embeddings can produce large gradients in early layers because the Jacobian of $\mathsf{layer\_norm}(x)$ w.r.t. $x$ is inversely proportional to $\norm{x}$, and, in early layers, the norm of the residual stream is essentially the norm of the embeddings.
We experiment with the full range of remedies discussed in~\citet{spikenomore}, but found that they impacted the speed of convergence.
Instead, we simply turn off weight decay for embeddings and observe that embedding norms settle in a healthy region as training progresses.

\section{Deep Dive: \diveAnnealing}
\label{sec:diveAnnealing}

Recent works have suggested that a multi-stage approach to base model training can lead to measurable improvements in capabilities~\citep{blakeney2024doesdatasparkjoy,ibrahim2024simplescalablestrategiescontinually,feng2024maximize}.
In previous \olmo iterations, we also found that both   learning rate schedule (\olmozero; \citealt{Groeneveld2024OLMoAT}) and data mixture (\olmoapril; \citealt{olmo_blog}) play an important role.
We refer to interventions at this stage of model development as \textit{mid-training}\footnote{while the concept of chaining of multiple stages of self-supervised training is not new (\textit{e.g.}, \citealt{gururangan2020dontStopPretraining}), we trace the use of \textit{mid-training} to \citet{abdin2024phi} and \citet{openai2024midtraining}.}.

From afar, our approach is simple: after the pretraining stage, we generate domain-specific data mixtures and restart training, linearly driving the learning rate down to zero. 
Our goal is to imbue specialized knowledge and improve capabilities; feedback on these improvements comes from key benchmarks, such as math-specific tasks such as GSM8K. 

\subsection{Learning rate annealing}
\label{sec:learning_rate_deep_dive}

Our starting point for learning rate experiments was the setting from~\citet{dubey2024llama}.
To initialize the optimizer state for the 7B variant, we linearly warm up the learning rate to its peak of $3 \cdot 10^{-4}$ over the first 2000 steps.
Then, we use a standard cosine decay over 5T tokens.
Previous experience with \olmoapril suggests that the last part of a cosine decay schedule can be cut off and replaced by a linear decay to zero with little loss of performance.
Accordingly, for the 7B variant, we stop the schedule at 4T tokens and then switch to mid-training as described in Section~\S\ref{sec:diveAnnealing}.
The 13B ran with a higher peak learning rate from the start, so we decided to run it to 5T tokens before moving to the mid-training stage.

\begin{figure}[h]
    \centering
    \includegraphics[width=0.49\textwidth]{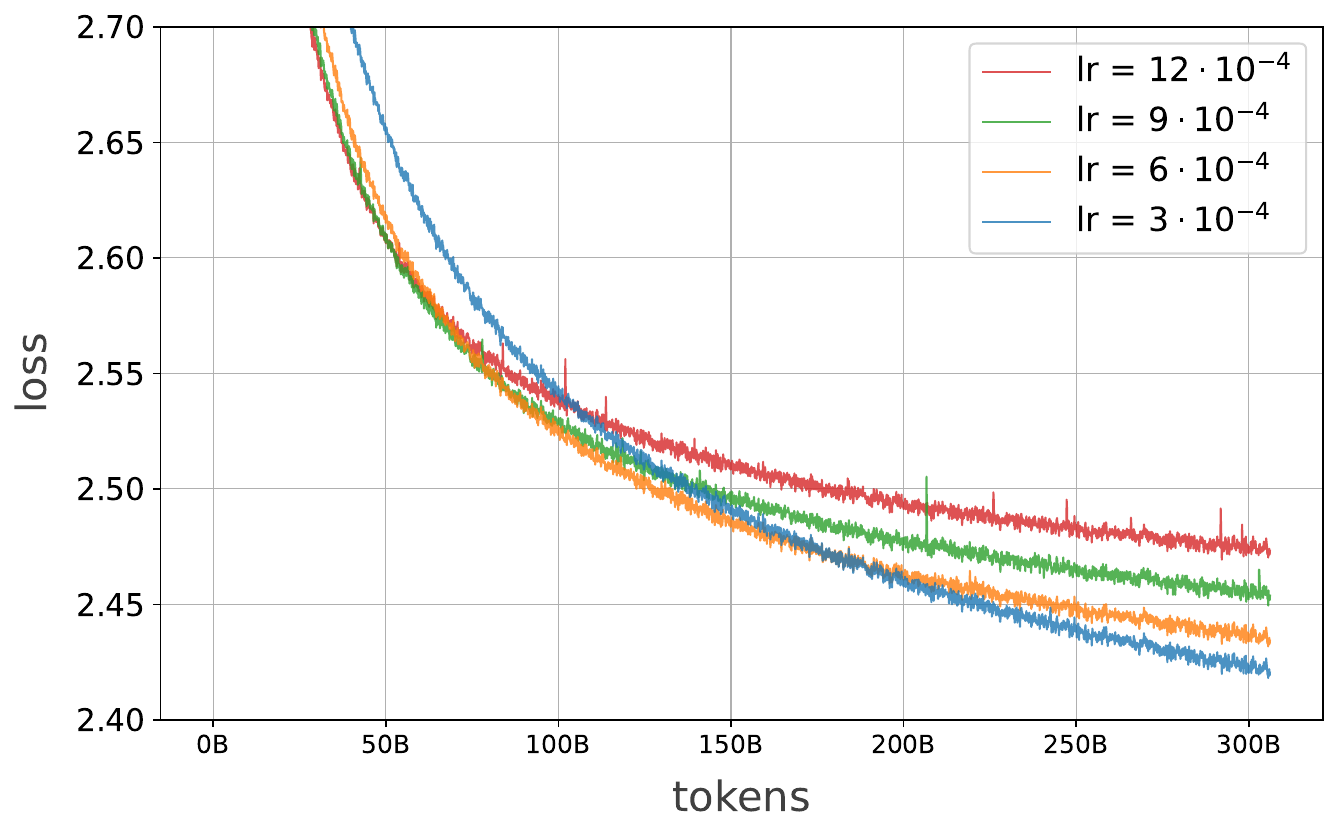}
    \includegraphics[width=0.49\textwidth]{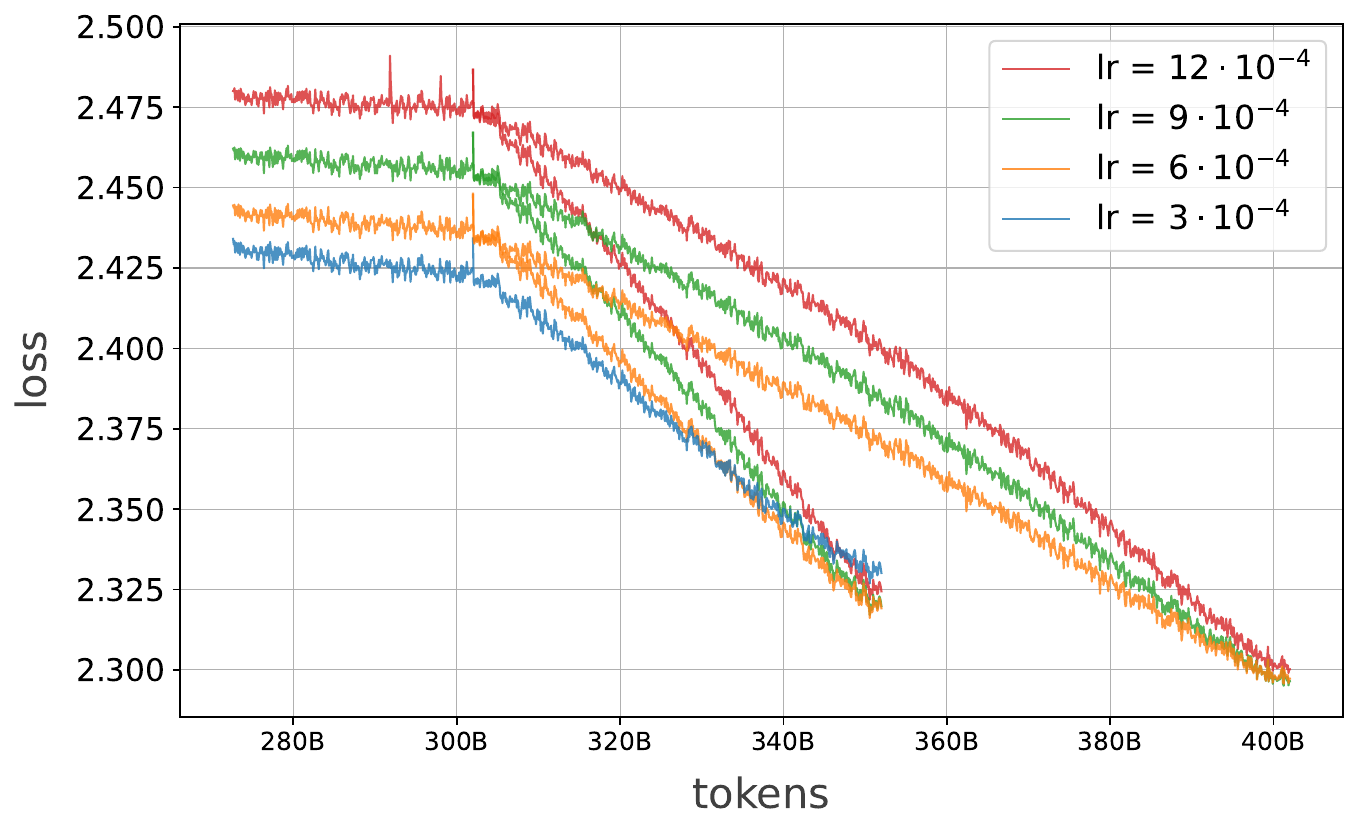}
    \caption{Higher learning rates perform better at first but are eventually overtaken by lower rates. However, linearly decaying the learning rate to zero over 50B or 100B tokens results in equivalent training loss.}
    \label{fig:lrsweep}
\end{figure}

Figure~\ref{fig:lrsweep} shows different runs with four additional learning rate values: $6 \cdot 10^{-4}$, $9 \cdot 10^{-4}$, $12 \cdot 10^{-4}$, and $30 \cdot 10^{-4}$.
In particular, we tried double, triple, quadruple, 10$\times$, and 30$\times$ the original learning rate.
The last, $30 \cdot 10^{-4}$, showed training instabilities already during learning rate warm-up, with several loss spikes that did not recover fully, so we abandoned this variant quickly.
The other values trained normally and showed an interesting pattern.
Looking purely at training loss, higher learning rates universally perform better early on (as long as they avoid loss spikes), but eventually the lower learning rate setting overtakes the others (Figure~\ref{fig:lrsweep}).
Notably, when comparing $3 \cdot 10^{-4}$ and $6 \cdot 10^{-4}$, the cross-over point is well past 200B tokens.
A shorter hyperparameter experiment might come to the wrong conclusion.

One of the motivations for this line of experimentation was to find out whether a higher learning rate would make the annealing step more effective.
The conjecture is that the worse training loss during pretraining is compensated for when the learning rate decays to zero.
To test this hypothesis, we took a checkpoint from each of our four variants after 300B tokens, and decayed the learning rate to zero over 50B tokens.
To account for the possibility that the effect of higher learning rates needs more steps to unfold, we tried the three higher settings and decayed the learning rate over 100B tokens, for a total of seven experiments.
The results show that a higher learning rate does make mid-training more effective, but it does so by exactly the amount that the pretraining is worse.
All four variants show the same training loss at the end of the procedure, though the lowest setting lags behind the others by a small amount.

Table~\ref{tab:lr_exp_downstream} shows that the result is consistent for longer training runs as well.
We took two variants, $3 \cdot 10^{-4}$ and $6 \cdot 10^{-4}$, and repeated the experiment after training for 1T and for 2T tokens.
We chose these variants because $3 \cdot 10^{-4}$ is the baseline from~\citet{dubey2024llama}, and $6 \cdot 10^{-4}$ showed, by a slim margin, the best training loss.
Our results show virtually no difference between the two settings, both on training loss and a mix of nine downstream tasks from the OLMES suite~\citep{olmes} shown in Table~\ref{tab:lr_exp_downstream}.
Evaluating the models on downstream tasks is noisier, but mirrors the findings based on training loss only.

\begin{table}[ht]
\setlength\tabcolsep{10pt} 
\begin{center}
\begin{small}
\begin{tabular}{l l l c}
\toprule
\textbf{Learning Rate} & \textbf{Pretraining Stage} & \textbf{Mid-training Stage} & \textbf{OLMES} \sans{(CF, valid)} \\
\midrule
\rowcolor{ai2offwhite}
$3 \cdot 10^{-4}$ & 300B tokens & 50B tokens & 62.5 \\
\rowcolor{ai2offwhite}
$6 \cdot 10^{-4}$ & 300B tokens & 50B tokens & 63.9 \\
\rowcolor{ai2offwhite}
$9 \cdot 10^{-4}$ & 300B tokens & 50B tokens & 64.1 \\
\rowcolor{ai2offwhite}
$12 \cdot 10^{-4}$ & 300B tokens & 50B tokens  & 63.6 \\
$6 \cdot 10^{-4}$ & 300B tokens & 100B tokens  & 64.6 \\
$9 \cdot 10^{-4}$ & 300B tokens & 100B tokens & 64.5 \\
$12 \cdot 10^{-4}$ & 300B tokens & 100B tokens & 64.2 \\
\rowcolor{ai2offwhite}
$3 \cdot 10^{-4}$ & 2T tokens & 100B high quality tokens  & 73.8 \\
\rowcolor{ai2offwhite}
$6 \cdot 10^{-4}$ & 2T tokens & 100B high quality tokens & 73.9 \\
\bottomrule
\end{tabular}
\end{small}
\vspace{2mm}
\caption{Results on 9 multiple-choice tasks from the \textit{validation} subset of OLMES (\textit{cloze formulation} format) for various peak learning rates and schedule lengths. Average scores vary by less than two points across all variants, with most scores within half a point of each other. }
\label{tab:lr_exp_downstream}
\end{center}
\end{table}

Finally, we wanted to see if a higher learning rate during the pretraining stage would result in a more effective mid-training stage when switching to higher quality data.
To match our training setup as much as possible within the available compute budget, we took the same two settings ($3 \cdot 10^{-4}$ and $6 \cdot 10^{-4}$), and linearly decayed the learning rate to 0 over 100B high quality tokens.
Once again, the results show little difference.
The final scores on the OLMES evaluation suite are within 0.1 points of each other.
However, looking at other metrics may still reveal a meaningful difference between the two settings.
The mix of high quality tokens targets math specifically, and on GSM8K (which is not part of the OLMES suite), the high learning rate setting is 2.8 points better than the lower learning rate.
More study is needed to turn this interesting data point into a dependable result.

This finding contradicts machine learning folk wisdoms such as ``higher learning rates are always better'' or ``area under the learning curve matters''~\citep{mccandlish2018empirical}.
It expands on~\citet{mitch}, who observed that smaller models' performance is largely invariant to learning rate over several orders of magnitude when trained to the end of a cosine schedule, and further found that QK-norm (section \ref{sec:qknorm}) and z-loss (section \ref{sec:zloss}), which we use as well, enhance this effect.
We find that these results still hold even at much larger scales of tokens and parameters, and, crucially for our training efforts, with our modified learning rate schedule.

Due to cost concerns we did not explore the full range of learning rates.
This is the main limitation of this line of experimentation.
It would be interesting to run a wider sweep of learning rates to accurately define the boundaries of the plateau we appear to be training in.

\subsection{Data Curriculum: \dolminos}
\label{sec:data-curriculum}

In this section, we describe our experimental process for curating our mid-training data.
We collectively refer to the resulting dataset and mixtures created for this mid-training stage as \textbf{\dolminos}. 
An overview of the contents of this dataset is provided in Section~\S\ref{sec:pretrain-data} (Table~\ref{table:stage-2-data}). 
In detail, we use the following procedure in our mid-training recipe:
\begin{itemize}
    \item Identify a mix of high-quality sources to improve performance across the entire development benchmark suite (Section~\S\ref{sec:annealing_general_mix}).
    \item For patching specific capabilities (specifically, in the case of \olmotoo, math), collect and evaluate domain-specific datasets to mix during mid-training (Section~\S\ref{sec:annealing_math_mix}). We found that these sources can be independently assessed through a technique we dub \textit{microannealing} (Section~\S\ref{sec:microanneals}); their effectiveness persists when mixed with rest of sources.  
    \item Following experiments described in Section~\S\ref{sec:learning_rate_deep_dive}, we mix high-quality sources and math-specific data in three different token budgets (50B, 100B, 300B). 
    The smaller mix is used to mid-train \olmotoo 7B, while \olmotoo 13B and 32B are annealed on the larger ones. 
    For both \olmotoo 7B, 13B and 32B, we find that averaging weights of different checkpoints trained on same mixture but different data order seeds consistently improves over individual checkpoints (Section~\S\ref{sec:mid_soup}). 
    To demonstrate this on the small scale, we also include results for a 1B model that receives similar interventions as the 7B model.
\end{itemize}

\begin{table}[ht]
\setlength\tabcolsep{3pt}
\begin{center}
\begin{small}
\begin{tabular}{lc|cccccc|cccc}

\toprule
    \multicolumn{2}{l}{} &
    \multicolumn{6}{c}{\textbf{\texttt{Dev Benchmarks}}} & 
    \multicolumn{4}{c}{\textbf{\texttt{Held-out Evals}}} 
\\
    {\textbf{\fontsize{8}{8}\selectfont{Checkpoint}}} &
    {\textbf{\fontsize{8}{8}\selectfont~Avg}} &
    {\textbf{\fontsize{8}{8}\selectfont~MMLU}} & 
    {$\textbf{\fontsize{8}{8}\selectfont~ARC}_\textbf{\fontsize{6}{6}\selectfont~C}$} & 
    {\textbf{\fontsize{8}{8}\selectfont~HSwag}} & 
    {\textbf{\fontsize{8}{8}\selectfont~WinoG}} & 
    {\textbf{\fontsize{8}{8}\selectfont~NQ}} & 
    {\textbf{\fontsize{8}{8}\selectfont~DROP}} & 
    {\textbf{\fontsize{8}{8}\selectfont~AGIEval}} & 
    {\textbf{\fontsize{8}{8}\selectfont~GSM8K}} & 
    {$\textbf{\fontsize{8}{8}\selectfont~MMLU}_\textbf{\fontsize{6}{6}\selectfont~PRO}$} &
    {\textbf{\fontsize{8}{8}\selectfont~TQA}}
\\
\midrule
\rowcolor{midgrey}\multicolumn{12}{c}{\textbf{\olmotoo 1B}} \\
\rowcolor{lightgrey}{{Pretraining}} & 31.9 & 26.9 & 26.1 & 67.5 & 67.8 & 16.1 & 25.1 & 24.5 & 3.3 & 11.1 & 50.1 \\
\rowcolor{lightgrey}{{Pretraining \& mid-training}\hspace{.5em}} & 43.7 & 44.3 & 51.3 & 69.5 & 66.5 & 20.8 & 34.0 & 36.3 & 43.8 & 16.1 & 54.7 \\
\rowcolor{midgrey}\multicolumn{12}{c}{\textbf{\olmotoo 7B}} \\
\rowcolor{lightgrey}{{Pretraining}} & 53.0 & 59.8 & 72.6 & 81.3 & 75.8 & 29.0 & 40.7 & 44.6 & 24.1 & 27.4 & 74.6 \\
\rowcolor{lightgrey}{{Pretraining \& mid-training}\hspace{.5em}} & 62.9 & 63.7 & 79.8 & 83.8 & 77.2 & 36.9 & 60.8 & 50.4 & 67.5 & 31.0 & 78.0 \\
\rowcolor{ai2midwhite}\multicolumn{12}{c}{\textbf{\olmotoo 13B}} \\
\rowcolor{ai2offwhite}{{Pretraining}} & 58.9 & 63.4 & 80.2 & 84.8 & 79.4 & 34.6 & 49.6 & 48.2 & 37.3 & 31.2 & 80.3 \\
\rowcolor{ai2offwhite}{{Pretraining \& mid-training}\hspace{.5em}} & 68.3 & 67.5 & 83.5 & 86.4 & 81.5 & 46.7 & 70.7 & 54.2 & 75.1 & 35.1 & 81.9 \\
\rowcolor{ai2midpink}\multicolumn{12}{c}{\textbf{\olmotoo 32B}} \\
\rowcolor{ai2lightpink}{{Pretraining}} & 66.3 & 72.9 & 88.7 & 84.2 & 82.4 & 40.6 & 57.2 & 56.8 & 56.2 & 38.5 & 85.4 \\
\rowcolor{ai2lightpink}{{Pretraining \& mid-training}\hspace{.5em}} & 73.3 & 74.9 & 90.4 & 89.7 & 83.0 & 50.2 & 74.3 & 61.0 & 78.8 & 43.3 & 88.0 \\
\bottomrule
\end{tabular}
\end{small}
\vspace{2mm}
\caption{
Evaluations comparing \olmotoo 1B, 7B, 13B and 32B at the end of pretraining and mid-training stages (setup mirrors Table~\ref{tab:evals_overview}). 
Pretrain checkpoints have been trained on 4 trillion (1B, 7B), 5 trillion (13B) and 7 trillion (32B) tokens respectively. 
For 7B, we obtain the final mid-train checkpoints by averaging three training runs on 50B \textsc{Dolmino} tokens;
for 13B and 32B, we use three runs on 100B tokens and one run on 300B tokens.
For 1B, the final checkpoint is the result of training on 50B \textsc{Dolmino} tokens \textit{without} averaging.}
\label{tab:evals-pre-post-anneal}
\end{center}
\end{table}

Table~\ref{tab:evals-pre-post-anneal} summarizes the dramatic impact of this mid-training phase on both development and held-out evals. 
\olmotoo 7B model improves, on average by 10.6 points, surpassing the larger 13B model after the pretraining stage.
For its part, \olmotoo 13B benefits equally from mid-training, improving its average performance by 10.3 points.
Both models see improvements in knowledge-intensive, multiple-choice  (Arc challenge: $72.6\rightarrow79.8$ for 7B, $80.2\rightarrow83.5$ for 13B; MMLU: $59.8\rightarrow63.7$ for 7B, $63.4\rightarrow67.5$ for 13B; AGIEval: $44.6\rightarrow50.4$ for 7B, $48.2\rightarrow54.2$ for 13B), reading comprehension (Natural Questions: $29.0\rightarrow36.9$ for 7B, $34.6\rightarrow46.7$ for 13B; DROP: $40.7\rightarrow60.8$ for 7B, $49.6\rightarrow70.7$ for 13B), and math skills (GSM8K: $24.1\rightarrow67.5$ for 7B, $37.3\rightarrow75.1$ for 13B) benchmarks.

\subsection{\dolminos: High Quality Sources}
\label{sec:annealing_general_mix}

\newcommand{\MeBaseline}{{\sans{PT Mix}}\xspace}
\newcommand{\MeDclmSeven}{{\sans{Web }}$^{\text{{\sans{FT}}}_\text{{\sans{7}}}}$\xspace}
\newcommand{\MeFinewebThree}{{\sans{Web }}$^{\text{{\sans{FT}}}_\text{{\sans{7}}}}_{\text{{\sans{FW}}}_\text{{\sans{3}}}}$\xspace}
\newcommand{\MeFinewebTwo}{{\sans{Web }}$^{\text{{\sans{FT}}}_\text{{\sans{7}}}}_{\text{{\sans{FW}}}_\text{{\sans{2}}}}$\xspace}
\newcommand{\MeFinewebTwoMath}{{\sans{Web }}$^{\text{{\sans{FT}}}_\text{{\sans{7}}}}_{\text{{\sans{FW}}}_\text{{\sans{2}}}}${\sans + Math}\xspace}
\newcommand{\MeFinewebTwoIns}{{\sans{Web }}$^{\text{{\sans{FT}}}_\text{{\sans{7}}}}_{\text{{\sans{FW}}}_\text{{\sans{2}}}}${\sans + Ins}\xspace}
\newcommand{\MeFinewebTwoMathIns}{{\sans{Web }}$^{\text{{\sans{FT}}}_\text{{\sans{7}}}}_{\text{{\sans{FW}}}_\text{{\sans{2}}}}${\sans + Math + Ins}\xspace}

\begin{table}[ht]
\renewcommand{\arraystretch}{.9}
\setlength\tabcolsep{3pt}
\begin{center}
\begin{footnotesize}
\newcommand{\rot}[1]{\rotatebox{90}{#1}}
\begin{tabular}{c @{\hskip 6pt} l @{\hskip 6pt} l c c c c c c c}
\toprule
    & \multirow{2}{*}{\vspace{-1.5em}\textbf{\footnotesize{Source}}} 
    &
    & \multicolumn{7}{c}{\textbf{\footnotesize~Mix \%}} 
    \vspace{.5em}
    \\[-.3em]
    &
    &
    &
    \MeBaseline &
    \MeDclmSeven &
    \MeFinewebThree &
    \MeFinewebTwo &
    \makecell[lc]{
        \MeFinewebTwo
        \\[-.2em]
        {\sans{+ Math}}
    } &
    \makecell[lc]{
        \MeFinewebTwo
        \\[-.2em]
        {\sans{+ Ins}}
    } &
    \makecell[lc]{
        \MeFinewebTwo
        \\[-.2em]
        {\sans{+ Math}}
        \\[-.2em]
        {\sans{+ Ins}}
    }
    \\
\midrule
    \rowcolor{ai2offwhite}
    &
    \sans{DCLM} & 
    \begin{tabular}{@{}l@{}}
        {\scriptsize\color{darkergrey}from}
        \\[-.4em]
        {\scriptsize\color{darkergrey}pretrain}
    \end{tabular} & 
    95.2 & 
    - & 
    - & 
    - & 
    - & 
    - & 
    - 
\\
    \rowcolor{ai2offwhite}
    & 
    \sans{DCLM} &
    \begin{tabular}{@{}l@{}}
        {\scriptsize\color{darkergrey}FT top 7\%}\\
    \end{tabular} & 
    - &
    57.1 &
    - &
    - &
    - &
    - &
    -
\\
    \rowcolor{ai2offwhite}
    & 
    \sans{DCLM} &
    \begin{tabular}{@{}l@{}}
        {\scriptsize\color{darkergrey}FT top 7\%}
        \\[-.4em]
        {\scriptsize\color{darkergrey}$\text{FineWeb}\geqslant3$}
    \end{tabular} & 
    - &
    - &
    54.2 &
    - &
    - &
    - &
    -
\\
    \rowcolor{ai2offwhite}
    \multirow{-4}{*}{\rot{\textbf{\hspace{2em}\scriptsize{WEB}}}} & 
    \sans{DCLM} &
    \begin{tabular}{@{}l@{}}
        {\scriptsize\color{darkergrey}FT top 7\%}
        \\[-.4em]
        {\scriptsize\color{darkergrey}$\text{FineWeb}\geqslant2$}
    \end{tabular} & 
    - &
    - &
    - &
    57.9 &
    61.8 &
    75.5 &
    57.5
\\
    &
    \sans{Flan} &
    \begin{tabular}{@{}l@{}}
        {\scriptsize\color{darkergrey}Dolma 1.7}
        \\[-.4em]
        {\scriptsize\color{darkergrey}decontaminated}
    \end{tabular} & 
    - &
    - &
    - &
    - &
    - &
    8.8 &
    6.7
\\
    \multirow{-2}{*}{\rot{\textbf{\hspace{2em}\scriptsize{INST}}}} &
    \sans{Stack Exchange} &
    \begin{tabular}{@{}l@{}}
        {\scriptsize\color{darkergrey}2024/09/30 dump}
        \\[-.4em]
        {\scriptsize\color{darkergrey}Q\&A format}
    \end{tabular} & 
    - &
    - &
    - &
    - &
    - &
    0.7 &
    0.5
\\
    \rowcolor{ai2offwhite}
    &
    \sans{Starcoder} &
    \begin{tabular}{@{}l@{}}
        {\scriptsize\color{darkergrey}from}
        \\[-.4em]
        {\scriptsize\color{darkergrey}pretrain}
    \end{tabular} & 
    2.1 &
    19.5 &
    20.9 &
    19.2 &
    - &
    - &
    -
\\
    \rowcolor{ai2offwhite} 
    \multirow{-2}{*}{\rot{\hspace{2em}\textbf{\scriptsize{CODE}}}} &
    \sans{CodeSearchNet} &
    \begin{tabular}{@{}l@{}}
        {\scriptsize\color{darkergrey}unfiltered}\\
    \end{tabular} & 
    - &
    - &
    - &
    - &
    0.1 &
    0.2 &
    0.1
\\
    &
    \sans{Gutenberg Books} &
    \begin{tabular}{@{}l@{}}
        {\scriptsize\color{darkergrey}from}
        \\[-.4em]
        {\scriptsize\color{darkergrey}Dolma 1.7}
    \end{tabular} & 
    - &
    1.2 &
    1.3 &
    1.2 &
    - &
    - &
    -
\\
    &
    \sans{peS2o} &
    \begin{tabular}{@{}l@{}}
        {\scriptsize\color{darkergrey}from}
        \\[-.4em]
        {\scriptsize\color{darkergrey}pretrain}
    \end{tabular} & 
    1.5 &
    6.6 &
    7.1 &
    6.5 &
    10.7 &
    13.0 &
    9.9
\\
    &
    \sans{Wikipedia} &
    \begin{tabular}{@{}l@{}}
        {\scriptsize\color{darkergrey}from}
        \\[-.4em]
        {\scriptsize\color{darkergrey}pretrain}
    \end{tabular} & 
    0.1 &
    0.9 &
    0.9 &
    0.9 &
    1.6 &
    1.9 &
    1.4
\\
    &
    \sans{StackExchange} &
    \begin{tabular}{@{}l@{}}
        {\scriptsize\color{darkergrey}from}
        \\[-.4em]
        {\scriptsize\color{darkergrey}RedPajama v1}
    \end{tabular} & 
    - &
    4.0 &
    4.3 &
    4.0 &
    - &
    - &
    -
\\
    \multirow{-5}{*}{\rot{\textbf{\hspace{4em}\scriptsize{REFERENCE}}}} &
    \sans{ArXiv} &
    \begin{tabular}{@{}l@{}}
        {\scriptsize\color{darkergrey}from}
        \\[-.4em]
        {\scriptsize\color{darkergrey}pretrain}
    \end{tabular} & 
    0.5 &
    4.9 &
    5.2 &
    4.8 &
    - &
    - &
    -
\\
    \rowcolor{ai2offwhite}&
    \sans{Algebraic Stack} &
    \begin{tabular}{@{}l@{}}
        {\scriptsize\color{darkergrey}from}
        \\[-.4em]
        {\scriptsize\color{darkergrey}pretrain}
    \end{tabular} & 
    0.3 &
    2.8 &
    3.0 &
    2.7 &
    - &
    - &
    -
\\
    \rowcolor{ai2offwhite}&
    \sans{OpenWebMath} &
    \begin{tabular}{@{}l@{}}
        {\scriptsize\color{darkergrey}from}
        \\[-.4em]
        {\scriptsize\color{darkergrey}pretrain}
    \end{tabular} & 
    0.3 &
    2.9 &
    3.1 &
    2.8 &
    5.2 &
    - &
    4.8
\\
    \rowcolor{ai2offwhite}&
    \sans{GSM8k} &
    \begin{tabular}{@{}l@{}}
        {\scriptsize\color{darkergrey}train}
        \\[-.4em]
        {\scriptsize\color{darkergrey}split}
    \end{tabular} & 
    - &
    - &
    0.003 &
    0.003 &
    0.003 &
    - &
    0.003
\\
    \rowcolor{ai2offwhite}&
    \sans{Mathpile} &
    \begin{tabular}{@{}l@{}}
        {\scriptsize\color{darkergrey}commercial subset}
        \\[-.4em]
        {\scriptsize\color{darkergrey}train split}
    \end{tabular} & 
    - &
    - &
    - &
    - &
    2.1 &
    - &
    1.9
\\
    \rowcolor{ai2offwhite}
    \multirow{-5}{*}{\rot{\textbf{\hspace{2.3em}\scriptsize{MATH}}}} &
    \sans{AutoMathText} &
    \begin{tabular}{@{}l@{}}
        {\scriptsize\color{darkergrey}unfiltered}
    \end{tabular} & 
    - &
    - &
    - &
    - &
    18.5 &
    - &
    17.2
\\[.2em]
\bottomrule
\end{tabular}
\end{footnotesize}
\end{center}
\caption{
    A summary of high-quality sources we evaluate for mid-training. 
    We experiment with mixing these sources in 6 mixes, each consisting of 50 billion tokens. 
    Percentages on the table indicate the fraction of each 50B mix that is comprised by data from the respective source. 
    \MeBaseline is sampled (with repetition) from the pretraining stage. 
}
\label{table:hq_mixes}
\end{table}

Following the recipe from the previous \olmo iteration~\citep{olmo_blog}, we start by curating a higher quality subset of pretraining mix, and expand it with more academic and encyclopedic material. 
In particular, we consider the following sources (summarized in Table~\ref{table:hq_mixes}):

\paragraph{High quality web} 
To filter the web subset used in pretraining, we experiment with two existing quality classifiers: 
\begin{itemize}
    \item \textbf{FastText classifier from \citet{dclm}.} 
    To train this model\footnote{\href{https://huggingface.co/mlfoundations/fasttext-oh-eli5}{\huggingface\texttt{mlfoundations/fasttext-oh-eli5}}}, \citeauthor{dclm} sampled positive documents from the Reddit subset in ELI5~\citep{fan2019eli5}, and demonstrations from Open Hermes 2.5\footnote{\href{https://huggingface.co/datasets/teknium/OpenHermes-2.5}{\hfdataset\texttt{datasets/teknium/OpenHermes-2.5}}}. Negatives are sampled at random from the DCLM pipeline.
    \item \textbf{FineWeb Edu classifier from \citet{penedo2024fineweb}.} 
    This model\footnote{\href{https://huggingface.co/HuggingFaceFW/fineweb-edu-classifier}{\huggingface\texttt{HuggingFaceFW/fineweb-edu-classifier}}} is fine-tuned from the Arctic Embed M\footnote{\href{https://huggingface.co/Snowflake/snowflake-arctic-embed-m}{\huggingface\texttt{Snowflake/snowflake-arctic-embed-m}}} encoder~\citep{merrick2024arctic} on over 400,000 web pages\footnote{\href{https://huggingface.co/datasets/HuggingFaceFW/fineweb-edu-llama3-annotations5}{\hfdataset\texttt{datasets/HuggingFaceFW/fineweb-edu-llama3-annotations}}} labeled by Llama 3 70B Instruct. 
    This classifier scores documents from 0 to 5 according to adherence to academic topics and polished content. 
\end{itemize}

Following \citet{dclm}, we use the DCLM FastText classifier with a threshold of $0.03311014$, which retains approximately 65.6\% of the web subset. 
We combine this filter with the scores from FineWeb Edu classifier; 
we experiment by retaining documents with score over 3 (5.8\% retained), as well as a more relaxed threshold of 2 (20.3\% retained).

\paragraph{Instruction data and Q\&A pairs}
We leverage the same subset of FLAN~\citet{wei2021flan,longpre2023flan} from \dolma 1.7~\citep{soldaini2024dolma}. We decontaminated this source by extracting training, validation, and test instances from all tasks in our evaluation suite (Section~\S\ref{sec:basemodeleval}) and removed FLAN documents with 10\% or more overlapping ngrams with any task instance.

We source question and answer pairs from the Stack Exchange network, a collection of 186 forums dedicated to a wide variety of topics. 
Content on Stack Exchange network is licensed under various commercial-friendly Creative Common licenses. 
We use the latest database dump (September $\text{30}^\text{th}$, 2024) at the time of writing, which is distributed by the Internet Archive\footnote{\href{https://archive.org/details/stackexchange_20240930}{\texttt{archive.org/details/stackexchange\_20240930}}}.
We filter questions to those that have an accepted answer; further, we remove Q\&A pairs whose questions have fewer than 3 votes or answers have fewer than 5 votes. 
Once filtered, we concatenate questions and answers together using a sequence of new lines that contains one more \texttt{\textbackslash{n}} than the longest sequence of newlines in either the question or answer.

\paragraph{Code} We evaluate retaining the same subset of code used during pretraining;
furthermore, we consider smaller, curated sources of code interleaved with natural supervision, such as docstrings in CodeSearchNet~\citep{husain2019CodeSearchNet}; 
Q\&A pairs from StackExchange described in the paragraph above also contain code.

\paragraph{Academic, encyclopedic and other reference content} 
We source high-quality non-web datasets from Dolma 1.7~\citep{soldaini2024dolma}.
This includes peS2o~\citep{peS2o}, Wikipedia, and Wikibooks, Gutenberg books, arXiv and StackExchange (from Red-Pajama v1; \citealp{together2023redpajama}), Algebraic Stack (ProofPile II; \citealp{azerbayev2023llemma}).

\paragraph{Math}
In parallel to developing the math subset of \dolminos (Section~\S\ref{sec:annealing_math_mix}), we consider preliminary math subset to gauge how math documents combine with the non-math portion of the mix. 
In particular, we used OpenWebMath~\citep{paster2023openwebmath}, the train split of GSM8K~\citep{cobbe2021trainingverifierssolvemath}, the train split of the permissively licensed (``commercial'') subset of MathPile~\citep{wang2023mathpile}, and AutoMathText~\citep{zhang2024automathtext}.

\begin{table}[h]
\centering
\begin{small}
\renewcommand{\arraystretch}{1.1}
\begin{tabular}{lcccc}
\toprule
\textbf{Mid-training mix} & \textbf{OLMES} \sans{(MCF)} & \textbf{OLMES-Gen} & \textbf{MMLU} \sans{(MCF)} & \textbf{GSM*} \\
\midrule
\rowcolor{ai2offwhite}\textit{n/a (pretrain checkpoint)} & 69.6 & 63.2 & 59.8 & 28.5 \\
\MeBaseline & 74.0 & 64.5 & 61.8 & 27.0 \\
\rowcolor{ai2offwhite}\MeDclmSeven & 73.5 & 64.1 & 61.9 & 24.5 \\
\MeFinewebThree & 73.5 & 63.0 & 62.4 & 30.5 \\
\rowcolor{ai2offwhite}\MeFinewebTwo & 75.2 & 63.8 & 63.1 & 28.5 \\
\MeFinewebTwoIns & 74.2 & 64.1 & 63.0 & 46.0 \\
\rowcolor{ai2offwhite}\MeFinewebTwoMath & {\bf75.7} & 69.7 & 62.3 & {\bf52.0} \\
\MeFinewebTwoMathIns & {\bf75.7} & {\bf70.2} & {\bf63.1} & 46.5 \\[.2em]
\bottomrule
\end{tabular}
\end{small}
\caption{
Comparison of mid-training mixes introduced in Table~\ref{table:hq_mixes}. 
Each row corresponds to a 50 billion token training run following learning rate schedule described in Section~\S\ref{sec:learning_rate_deep_dive} (except first row).
Weights are initialized from a \olmotoo checkpoint pretrained for 4T tokens. 
We compare each run on a mix of OLMES core tasks (multiple choice format; see Table~\ref{tab:evals_overview}), OLMES generative tasks (Table~\ref{tab:evals_overview}), MMLU (multiple choice format; \citealp{hendryckstest2021}), and a random sample of 200 GSM8K~\citep{cobbe2021trainingverifierssolvemath} questions we use as development set (GSM*; Section~\S\ref{app:eval-base}).
Results on the {\bf{final mid-training mix}}  are in Table~\ref{tab:evals-pre-post-anneal}.
}
\label{table:hq_mixes_results}
\end{table}

Results of mixes shown in Table~\ref{table:hq_mixes} are summarized in Table~\ref{table:hq_mixes_results}. 
All results correspond to mid-training runs on 50 billion tokens, initialized from a 7B model checkpoint pretrained on 4 trillion tokens.

We find that, as noted in Section~\S\ref{sec:learning_rate_deep_dive}, learning rate anneal (\MeBaseline) alone yields notable improvements across all averages (OLMES $+4.4$; OLMES-Gen $+1.3$; MMLU $+20$), but not on our math development set (GSM* $-1.5$). 
Switching to mixes that contain higher quality web data and reference content further improves performance: \MeFinewebTwo further improves $+1.2$ points over \MeBaseline in OLMES and $+1.3$ in MMLU; it is slightly worse on OLMES-Gen ($-0.4$) and within margin of error on GSM* ($+1.5$). 
Finally including instruction data and math sources in the mix yields the best performance. 
\MeFinewebTwoMathIns mix achieves best overall results, with $+1.7$ on OLMES, $+5.7$ on generative tasks, $+1.3$  on MMLU, and $+19.5$ on GSM*. 
We note that \MeFinewebTwoMath mix performs slightly better on math tasks, motivating our investigation in better math subsets that combine well with other high-quality sources in Section~\S\ref{sec:annealing_math_mix}.

\subsection{\dolminos: Math Mix}
\label{sec:annealing_math_mix}

Early mid-training mixes ($\text{\fontsize{8}{8}\sans{Web}}_{\ast}^{\ast}$ only rows in Table~\ref{table:hq_mixes_results}) show models struggle in math-related benchmarks.
Thus, improving performance on these sets is a central focus of our mid-training investigations. 
We investigate both human-authored and synthetically generated or augmented data;
we derived the latter through an iterative procedure aimed at fixing common errors in our math validation sets.

We describe both the data sources and their generation/filtration procedure in Section~\S\ref{sec:midtrain:math_sources}; 
then, in Section~\S\ref{sec:microanneals}, we detail \textit{microanneals}, the experimentation technique we use to finalize math sources. 
The resulting mix is summarized in Table~\ref{table:stage-2-data}.

\subsubsection{Math Sources}
\label{sec:midtrain:math_sources}

\paragraph{TuluMath} We follow the recent \textit{persona-driven} methodology in \citet{chan2024scaling} to generate math synthetic data. The key idea is to use different personas (e.g., ``A machine learning researcher focused on neural networks'') with a data synthesis prompt (e.g., ``create a math problem'') to steer an LM to synthesize data with corresponding perspectives. 
Specifically, we condition on available personas from Persona Hub \citep{chan2024scaling} to generate prompts targeting Math problems both those that require advanced mathematical skills as well as grade school problems. We zero-shot-prompt  \texttt{GPT-4o}\footnote{\texttt{2024-08-06}} to generate problems that are unique and specific to a given persona input. 
Having generated the problems, we then generate multi-step math solutions using \texttt{GPT-4o}. Exact prompts used to generate problems and solutions are provided in Appendix Figures \ref{fig:persona-prompt-sft-math} and \ref{fig:persona-prompt-sft-math-res}. In total, we collected $\sim$ 230M synthetic math tokens.

\paragraph{DolminoSynthMath} This is a collection of 28M synthetic math tokens designed specifically to improve performance on GSM8K as well as raw mathematical calculations. It is composed of three parts: first we generate 11M tokens of basic mathematical question and answer pairs such as ``77 * 14 = 1078'' and pair each of these with a variety of prompts. We find that including such data dramatically mitigates the mistakes our model makes within individual CoT reasoning steps at inference time. Next we include a custom collection of 7,924 synthetic GSM8K examples, which are produced by consuming a GSM8K training example and replacing all of its numbers in both the provided question and answer, with the hope that this would provide signal to the model to extract the computation graph from a word problem and ignore irrelevant semantic features. Finally we include a MIND-rewriting \citep{akter2024mindmathinformedsynthetic} of each of the GSM8K training examples, where the synthetic data was generated using Qwen2.5-7B-Instruct~\citep{qwen2.5}.

\paragraph{TinyGSM-MIND} We generated approximately 6.5B tokens of synthetic math data from rewritten versions of Tiny-GSM \citep{liu2023tinygsmachieving80gsm8k}. Tiny-GSM is a collection of 11M synthetic GSM8K-like questions, where the answers are provided in the form of python code. We filter this set to only include answers that have code that is executable and only contains statements that are variable assignments. We then annotate each line of the code that is an assignment operator with the numerical value of the resulting variable. Then we pass all of these annotated examples to Qwen2.5-7B-Instruct to be rewritten in the style of MIND \citep{akter2024mindmathinformedsynthetic} using the `Two Students' and `Problem Solving' prompts.

\paragraph{MathCoder2-Synthetic} We emulate the synthetic data generation procedure of MathCoder2 \citep{wang2024mathcoder} to filter existing synthetic data from open-source repositories. In particular, we collect the synthetic textbooks from HuggingFace user \texttt{Ajibawa-2023},\footnote{\href{https://huggingface.co/datasets/ajibawa-2023/Maths-College}{\hfdataset\texttt{datasets/ajibawa-2023/Maths-College}}}$^\text{,}$\footnote{\href{https://huggingface.co/datasets/ajibawa-2023/Education-College-Students}{\hfdataset\texttt{datasets/ajibawa-2023/Education-College-Students}}} and from the M-A-P Matrix dataset and perform additional filtering on them. In particular we train a FastText classifier as follows: we ask GPT-4o to annotate 10,000 OpenWebMath examples \citep{paster2023openwebmath} as either math-related or non-math-related; we then use these as positive and negative examples for a FastText classifiers. We apply this classifier to the synthetic textbooks and only keep the math-related ones.

\paragraph{ProofPile OWM-Filtered} We use the same OpenWebMath filter generated in the previous step and apply it to Metamath~\citep{yu2023metamath} and CodeSearchNet~\citep{husain2019CodeSearchNet}.

\paragraph{GSM8K-Train} Finally, we include the training split of GSM8K \citep{cobbe2021trainingverifierssolvemath}.

\subsubsection{Evaluating Math Data with Microanneals}
\label{sec:microanneals}

To select the highest quality subset of all available and synthetic math data, we perform a series of several \textit{microanneals}, which were annealing runs focused on small math subsets. 
The general recipe for these microanneals is as follows:
\begin{enumerate}
\item  identify a source or small collection of math sources that we want to assess the data quality of; 
\item collect roughly the same quantity of data from the general data mix (e.g., DCLM) as from the math sources to ensure a mixture of high-quality web text alongside domain-specific math;
\item  train this 50/50 mixture as if it were an annealing run, making sure to linearly drive the learning rate down at the proper rate for this smaller collection of data. 
\end{enumerate}
This procedure facilitates evaluating  the quality of individual data sources at a fraction of the cost of a full annealing run. In total, we run 19 separate microanneals with a total token count of 130B tokens, equivalent to less than 3 full 50B annealing runs. Putting this cost into perspective, the totality of the 19 microanneals requires less compute than the 3 50B token souping ingredients used for our 7B model.
More explicitly, it shows improvements at a much finer-grained data-source resolution, with results visible after training for less than 10B tokens.

\begin{table}[h]
\setlength\tabcolsep{10pt}
\begin{center}
\begin{small}
\begin{tabular}{lcccc}
\toprule
\rowcolor{midgrey}\multicolumn{5}{c}{\textbf{\textit{Microanneal Experiment 1}}} \\
\rowcolor{lightgrey}{\bf{Mix}} & {\bf{Web ratio}} & {\bf{Tokens}} & {\bf{MMLU}} (avg) & {\bf{GSM*}} \\
\rowcolor{lightgrey}Baseline & \textit{n/a} & \textit{n/a} & 59.8 & 28.5 \\
\rowcolor{lightgrey}Math 35/65 & 65.0\% & 576M & 60.1 & 63.5 \\
\rowcolor{lightgrey}Math 10/90 & 88.3\% & 1.72B & 60.9 & 61.0 \\
\midrule
\rowcolor{ai2midwhite}\multicolumn{5}{c}{\textbf{\textit{Microanneal Experiment 2}}} \\
\rowcolor{ai2offwhite}{\bf{Mix}} & {\bf{Web ratio}} & {\bf{Tokens}} & {\bf{MMLU}} (avg) & {\bf{GSM*}} \\
\rowcolor{ai2offwhite} Baseline & \textit{n/a} & \textit{n/a} & 59.8 & 28.5 \\
\rowcolor{ai2offwhite} 1x Math & 65.0\% & 576M & 60.1 & 63.5 \\
\rowcolor{ai2offwhite} 2x Math & 49.3\% & 798M & 60.3 & 66.0 \\
\rowcolor{ai2offwhite} 4x Math & 48.6\% & 1.57B & 60.5 & 65.0 \\
\midrule
\rowcolor{ai2midpink}\multicolumn{5}{c}{\textbf{\textit{Microanneal Experiment 3}}} \\
\rowcolor{ai2lightpink}{\bf{Mix}} & {\bf{Web ratio}} & {\bf{Tokens}} & {\bf{MMLU}} (avg) & {\bf{GSM*}} \\
\rowcolor{ai2lightpink}
\rowcolor{ai2lightpink} Baseline & \textit{n/a} & \textit{n/a} & 59.8 & 28.5 \\
\rowcolor{ai2lightpink} TinyGSM-Inline & 47.9\% & 3.17B & 60.4 & 25.0 \\
\rowcolor{ai2lightpink} TinyGSM-MIND & 52.1\% & 6.40B & 61.4 & 65.5 \\
\rowcolor{ai2lightpink} 2x TinyGSM-MIND & 51.3\% & 12.6B & 62.1 & 70.0 \\
\bottomrule
\end{tabular}
\end{small}
\vspace{2mm}
\caption{Results from microanneal experiments to \olmotoo math capabilities. 
We evaluate math/not-math mixture ratio, impact of repeating math tokens, and different math datasets.
We use a random sample of 200 GSM8K~\citep{cobbe2021trainingverifierssolvemath} questions we use as development set (GSM*; Section~\S\ref{app:eval-base}) as a proxy for math capabilities.
We monitor average MMLU scores to ensure \olmotoo remains performant on knowledge intensive tasks.
}
\label{tab:microanneal_results-pretty}

\end{center}
\end{table}

We illustrate how microanneals lead to our final math mix through three sets of experiments reported in Table~\ref{tab:microanneal_results-pretty}.
The primary evaluation metrics we use to evaluate the quality here is MMLU, and GSM*,  which is our 200-example subset of the GSM8K evaluation set. Note that one goal of mid-training is to improve GSM8K performance, but we only allow ourselves to inspect performance on 200 of the 1319 GSM8K examples to inform decisions about data mixtures. 

\paragraph{\textit{Microanneal experiment 1:} domain specific data is helpful even in small proportions} We run the following experiment: starting from a 7B model that has completed pretraining, and a mixture of TuluMath, DolminoSynthMath, Metamath, CodeSearchNet, and GSM8K-Train, accounting for approximately 200M tokens, we train on both a 35/65 math/DCLM mixture and a 10/90 mixture and evaluate both the MMLU and GSM*. We see that the pre-anneal had a GSM* score of 28.5, the 35/65 mixture yields a GSM* of 63.5, and the 10/90 mixture yields a GSM* of 61. This suggests that it is not strictly necessary to have a large proportion of domain-specific data in the annealing mixture, just that domain-specific data is present.

\paragraph{\textit{Microanneal experiment 2:} some duplication is beneficial} Starting from the same setup as the previous experiment, we duplicate the math data for a total of two copies, and four copies. We see that one copy of the math yields a GSM* score of 61, two copies yields a score of 66, and four copies yields a score of 65. This suggests that even if there is a scarcity of high-quality domain-specific data, duplicating it a small number of times can still provide some gains.

\paragraph{\textit{Microanneal experiment 3:} rewriting can help dramatically} Here we once again start with a 7B  model that has completed pretraining and evaluate the effect that rewriting Tiny-GSM into a natural language format has on GSM* evaluation scores. Recall that Tiny-GSM has answers written in the form of code, and that our pretraining mix is only 2\% code. We run a microannealing run on a mixture using an inline-annotated form of TinyGSM and compare it to just the `Problem Solving' MIND rewritten variant of TinyGSM. Relative to the baseline, the code version of TinyGSM degrades GSM* performance, while the rewritten version dramatically improves the performance. This suggests the  power of rewriting as a tool to cheaply convert data to a more amenable form for training.

\subsection{Final Midtraining mix and Checkpoint Soups}
\label{sec:mid_soup}

\begin{table}[h]
\centering
\newcolumntype{E}{>{\columncolor{lightgrey}}c}
\newcolumntype{Q}{>{\columncolor{ai2offwhite}}c}
\newcolumntype{D}{>{\columncolor{ai2lightpink}}c}
\adjustbox{max width=\linewidth}{
\begin{tabular}{l@{\hspace{.2em}}c E E Q Q D D}
\toprule
    \multirow{2}{*}{\textbf{Source}} & 
    \multirow{2}{*}{\textbf{Tokens}} & 
    \multicolumn{2}{>{\columncolor{midgrey}}c}{\textbf{50B}} &
    \multicolumn{2}{>{\columncolor{ai2midwhite}}c}{\textbf{100B}} &
    \multicolumn{2}{>{\columncolor{ai2midpink}}c}{\textbf{300B}} 
\\
    \cmidrule(lr){3-4} \cmidrule(lr){5-6} \cmidrule(lr){7-8} 
    & 
    & 
    \multicolumn{1}{>{\columncolor{midgrey}}c}{\textbf{Source \%}} & 
    \multicolumn{1}{>{\columncolor{midgrey}}c}{\textbf{Mix \%}} &
    \multicolumn{1}{>{\columncolor{ai2midwhite}}c}{\textbf{Source \%}} & 
    \multicolumn{1}{>{\columncolor{ai2midwhite}}c}{\textbf{Mix \%}} & 
    \multicolumn{1}{>{\columncolor{ai2midpink}}c}{\textbf{Source \%}} & 
    \multicolumn{1}{>{\columncolor{ai2midpink}}c}{\textbf{Mix \%}} 
\\
\midrule
Filtered DCLM & 752B & 3.23 & 47.2 & 6.85 & 50.2 & 20.78 & 51.9 \\
Decontam. FLAN & 17.0B & 50.0 & 16.6 & 100 & 16.7 & 200 & 11.3 \\
StackExchange Q\&A & 1.26B & 100 & 2.45 & 200 & 2.47 & 400 & 1.68 \\
peS2o & 58.6B & 5.15 & 5.85 & 16.7 & 9.52 & 100 & 19.4 \\
Wikipedia/Wikibooks & 3.7B & 100 & 7.11 & 100 & 3.57 & 400 & 4.86 \\
Dolmino Math & 10.7B & 100 & 20.8 & 200 & 17.5 & 400 & 10.8 \\
\bottomrule
\end{tabular}
}
\vspace{3pt}
\caption{\dolminos compositions. The Source \% column indicates the fraction of the source that was used in the \textsc{Dolmino} mix. Numbers in this column greater than 100 indicate we used the data, e.g. 400 indicates a 4x repeat. The Mix \% column describes the proportion of the \textsc{Dolmino} mix that is composed of this source, i.e., this column should sum to 100\%.}
\label{tab:dolmino-composition-pretty}
\end{table}

The final composition of \dolminos is shown in Table~\ref{table:stage-2-data}. 
As previously mentioned, we sample 3 mixes of 50B, 100B, and 300B tokens; 
composition of each is summarized in Table~\ref{tab:dolmino-composition-pretty}.
Since experiments in Section~\S\ref{sec:annealing_general_mix}~and~\S\ref{sec:microanneals} show that keeping mixing proportion roughly constant across sources is beneficial, we repeat Stack Exchange Q\&A data and mid-training math data twice for the 100B tokens mix, and four times for the 300B mix;
additionally, we repeat FLAN twice and Wiki data four times for the 300B mix.
Across all mixes, filtered web data from the DCLM baseline represents roughly 50\% of the total tokens budget.
 
We train \olmotoo 7B on the 50B mix.
To account for the larger batch size (Section~\S\ref{section:training-recipe}), we use the 100B mix for \olmotoo 13B, ensuring the same number of steps during learning rate anneal. 
Further, we experiment with a longer anneal phase with \olmotoo 13B using the 300B mix.
We follow the same procedure for the 32B model.

\begin{table}[h!]
\renewcommand{\arraystretch}{.8}
\begin{center}
\begin{small}
\begin{tabular}{r@{\hskip 6pt}lcccc}
\toprule
    \multicolumn{2}{c}{\textbf{Mid-training mix}} & 
    \textbf{OLMES} \sans{(MCF)} & 
    \textbf{OLMES-Gen} & 
    \textbf{MMLU} \sans{(MCF)} & 
    \textbf{GSM*} 
\\
\midrule
    \rowcolor{ai2offwhite}
    &
    best single &
    75.6 &
    68.5 &
    61.2 &
    71.0 
\\
    \rowcolor{ai2offwhite}
    \multirow{-2}{*}{A} &
    3 x soup &
    77.0 &
    69.4 &
    62.0 &
    74.0 
\\
    &
    best single &
    75.3 &
    69.9 &
    61.5 &
    73.0 
\\
    \multirow{-2}{*}{B}&
    3 x soup &
    77.3 &
    70.1 &
    62.7 &
    77.0 
\\
    \rowcolor{ai2offwhite}
    &
    best single &
    76.3 &
    70.9 &
    62.8 &
    66.0 
\\
    \rowcolor{ai2offwhite}
    \multirow{-2}{*}{C} &
    3 x soup &
    76.8 &
    71.3 &
    63.5 &
    66.0 
\\[.3em]
    &
    best single &
    77.5 &
    71.2 &
    63.4 &
    59.5
\\
    \multirow{-2}{*}{D} &
    3 x soup &
    77.8 &
    71.7 &
    63.5 &
    60.0
\\
    \rowcolor{ai2offwhite}
    &
    best single &
    73.4 &
    63.1 &
    62.2 &
    60.5
\\
    \rowcolor{ai2offwhite}
    \multirow{-2}{*}{E} &
    3 x soup &
    75.3 &
    64.2 &
    63.1 &
    43.0
\\
    &
    best single &
    77.1 &
    69.9 &
    63.7 &
    73.5
\\
    \multirow{-2}{*}{F} &
    3 x soup &
    77.9 &
    70.4 &
    63.7 &
    74.5 
\\
\bottomrule
\end{tabular}
\end{small}
\end{center}
\caption{
Comparison of six mid-training mixes between best single checkpoint and the average of three checkpoints (\textit{soup}) trained on different data permutations. 
We run all experiments starting from 7B pretrained checkpoint; we run mid-training stage for 50B tokens. 
Souping consistently equals or outperform the single best checkpoint trained on the same mix.
}
\label{tab:soupes}
\end{table}

\paragraph{Mid-training model merging or ``soups''} 
Performing a na\"ive average of multiple model checkpoints trained with a different data order has been proven effective in both computer vision~\citep{modelsoups} and language modeling~\citep{dclm} applications. 
We confirm the effectiveness of this approach, also known as \textit{model merging or ``souping''}, on six different mid-training mixes, as shown in Table~\ref{tab:soupes}.
For all experiments, we find that merging 3 checkpoints annealed on three permutations of the same data mix consistently produces equal or better performance than any individual training run.

Based on this evidence, we extensively use model merging to obtain our final \olmotoo 7B and 13B models. 
For \olmotoo 7B, we average three checkpoints trained on the 50B sample of \dolminos.
For \olmotoo 13B and 32B, we average four checkpoints: three trained on the 100B sample, and one trained on a 300B sample;
we find this approach to be empirically better than averaging just the three 100B runs alone.

\section{Deep Dive: \divePost}
\label{sec:post-train}

To adapt OLMo 2 to downstream generative tasks, we follow the \tulu recipe~\citep{lambert2024tulu3} with an increased focus on permissive licenses and suitable adjustments to hyperparameters.
The \tulu approach involves three phases of training: supervised finetuning (SFT), preference tuning with Direct Preference Optimization (DPO; \citealp{rafailov2024direct}) and on-policy preference data, and finally Reinforcement Learning with Verifiable Rewards (RLVR).
We find that all of the stages in the \tulu Recipe easily translate to the OLMo 2 models.
This section focuses on the development of our 7B and 13B models, where the 1B and 32B models followed very similar recipes.

\begin{table}[h]
\centering
\setlength\tabcolsep{5pt}
\begin{small}
\adjustbox{max width=\linewidth}{
\begin{tabular}{{@{}ll C{0.7cm} R{1.4cm} C{1cm} C{2cm} c@{}}}
\toprule
 \textbf{Category} & \textbf{Benchmark} & \textbf{CoT} & \textbf{\# Shots} & \textbf{Chat} & \textbf{Multiturn ICL} & \textbf{Metric} \\\midrule
\rowcolor{ai2offwhite}  Knowledge Recall & MMLU & \cmark & 0 & \cmark & \xmark & EM \\
\rowcolor{ai2offwhite} & PopQA & \xmark & 15 & \cmark & \cmark & EM \\
\rowcolor{ai2offwhite} & TruthfulQA & \xmark & 6 & \cmark & \xmark & MC2 \\
  Reasoning & BigBenchHard & \cmark & 3 & \cmark & \cmark & EM \\
  & DROP & \xmark & 3 & \xmark & N/A & F1 \\
\rowcolor{ai2offwhite}  Math & GSM8K & \cmark & 8 & \cmark & \cmark & EM \\
\rowcolor{ai2offwhite}   & MATH & \cmark & 4 & \cmark & \cmark & Flex EM \\
   Instruction Following & IFEval & \xmark & 0 & \cmark & N/A & Pass@1 (prompt; loose) \\
  & AlpacaEval 2 & \xmark & 0 & \cmark & N/A & LC Winrate \\
 \rowcolor{ai2offwhite}  Safety & \tulu Safety & \xmark & 0 & \cmark & N/A & Average$^*$ \\
\bottomrule
\end{tabular}}
\end{small}
\vspace{3pt}
\caption{
The \olmotoo Instruct Evaluation Regime (Adapted from \citet{lambert2024tulu3}): settings for development ({\bf{top}}) and unseen ({\bf{bottom}}) portions of the evaluation suite. {\bf{CoT}} are evaluations run with chain of thought prompting~\citep{wei2022chain}.
{\bf{\# shots}} is the number of in-context examples in the evaluation template.
{\bf{Chat}} indicates whether we use a chat template while prompting the model.
{\bf{Multiturn ICL}} indicates that we present each in-context example as a separate turn in a conversation (applicable only when a chat template is used and \# Shots is not 0).
$^*$Average over multiple sub-evaluations---full details of the safety evaluation are in \citet{lambert2024tulu3}.
}
\label{tab:eval-post}
\end{table}

\paragraph{Supervised Finetuning (SFT)} The SFT training of \olmotooinstruct from \tulu relies on selecting the highest-quality, existing instruction datasets and complementing them with scaled synthetic data for Supervised Finetuning based on the PersonaHub method~\citep{chan2024scaling}.
We develop two SFT mixes---\texttt{tulu-3-sft-olmo-2-mixture} which we used for our 7B and 13B models and \texttt{tulu-3-sft-olmo-2-mixture-0225} which includes minor modifications and applied to our 1B and 32B models.

For \texttt{tulu-3-sft-olmo-2-mixture}, given that \olmotoo is not trained for multilingual tasks, we experimented with removing all multilingual data from the SFT stage.
When removing the entire Aya split and the multilingual samples of Wildchat from \tulu, we saw a degradation of $\sim 0.5$ points on average, indicating that the \tulu dataset is balanced and cannot be easily improved by removing irrelevant subsets.
In total, this SFT mix contains 939,104 prompts.

For the 1B and 32B mix, \texttt{tulu-3-sft-olmo-2-mixture-0225}, we further filtered out instructions that included mentions of a date cutoff from the synthetic data generation process as we noticed it was correlated with undesirable behavior like hallucinating date cutoffs and prefacing responses with ``As an AI language model...''.\footnote{These filtering methods were also applied to the chosen samples in the 32B preference data.} 
We also use majority voting to improve the quality of answers to our synthetic math questions, that is, preventing SFT on incorrect math answers. 
For our Persona MATH\footnote{Filtered dataset here: \url{https://huggingface.co/datasets/allenai/tulu-3-sft-personas-math-filtered}} and Grade School Math\footnote{Filtered dataset here: \url{https://huggingface.co/datasets/allenai/tulu-3-sft-personas-math-grade-filtered}} datasets from Tülu 3, we only include prompts and completions where the model reaches a majority vote over 5 completions.
In total, this SFT mix contains 866,138 prompts.

\begin{table}[t]
\centering
\setlength\tabcolsep{3pt}
{\footnotesize
\begin{tabular}{lccccccccccc}
\toprule
    & 
    \textbf{AVG} & 
    \textbf{AE2} & 
    \textbf{BBH} & 
    \textbf{DROP} & 
    \textbf{GSM8K} & 
    \textbf{IFE} & 
    \textbf{MATH} & 
    \textbf{MMLU} & 
    \textbf{Safety} & 
    \textbf{PQA} & 
    \textbf{TQA} \\
\midrule
OLMo 2 1B SFT & 36.9 & 2.4 & 32.8 & 33.8 & 52.1 & 50.5 & 13.2 & 36.4 & \bf{93.2} & 12.7 & 42.1 \\

\rowcolor{ai2offwhite} OLMo 2 1B DPO & 40.6 & \bf{9.5} & 33.0 & 34.5 & 59.0 & 67.1 & 14.1 & 39.9 & 89.9 & 12.3 & 46.4 \\

\rowcolor{ai2lightpink} OLMo 2 1B Instruct & \bf{42.7} & 9.1 & \bf{35.0} & \bf{34.6} & \bf{68.3} & \bf{70.1} & \bf{20.7} & \bf{40.0} & 87.6 & \bf{12.9} & \bf{48.7}
\\\midrule

OLMo 2 7B SFT & 51.4 & 10.2   & 49.6 & 59.6 & 74.6 & 66.9 & 25.3 & 61.1 & \bf{94.6} & \bf{23.6} & 48.6 \\
\rowcolor{ai2offwhite} OLMo 2 7B DPO & 55.9 & 27.9 & 51.1   & 60.2 & 82.6 & \bf{73.0} & 30.3 & 60.8 & 93.7 & 23.5 & 56.0 \\
 \rowcolor{ai2lightpink}   OLMo 2 7B Instruct & \bf{56.5} & \bf{29.1}   & \bf{51.4} & \bf{60.5} & \bf{85.1} & 72.3 & \bf{32.5} & \bf{61.3} & 93.3 & 23.2 & \bf{56.5} \\ \midrule
 OLMo 2 13B SFT & 56.6 & 11.5   & 59.9 & 71.3 & 76.3 & 68.6 & 29.5 & 68.0 & \bf{94.3} & \bf{29.4} & 57.1 \\
 \rowcolor{ai2offwhite}  OLMo 2 13B DPO & 62.0 & \bf{38.3} & 61.4 & 71.5 & 82.3 & \bf{80.2} & 35.2 & 67.9 & 90.3 & 29.0 & \bf{63.9} \\
 \rowcolor{ai2lightpink}  OLMo 2 13B Instruct & \bf{63.4} & 39.5 & \bf{63.0} & \bf{71.5} & \bf{87.4} & 82.6 & \bf{39.2} & \bf{68.5} & 89.7 & 28.8 & 64.3 \\ \midrule
 OLMo 2 32B SFT & 61.7 & 16.9 & 69.7 & 77.2 & 78.4 & 72.4 & 35.9 & 76.1 & \bf{93.8} & 35.4 & 61.3 \\
 \rowcolor{ai2offwhite} OLMo 2 32B DPO & \bf{68.8} & \bf{44.1} & 70.2 & 77.5 & 85.7 & 83.8 & 46.8 & \bf{78.0} & 91.9 & 36.4 & \bf{73.5} \\
 \rowcolor{ai2lightpink} OLMo 2 32B Instruct & \bf{68.8} & 42.8 & \bf{70.6} & \bf{78.0} & \bf{87.6} & \bf{85.6} & \bf{49.7} & 77.3 & 85.9 & \bf{37.5} & 73.2 \\
\bottomrule
\end{tabular}
}
\caption{Comparison of performance for OLMo 2 Instruct after different training stages. The final Instruct model is from the RLVR stage.
The following evaluation names are abbreviated: AVG -- Average, AE2 -- AlpacaEval 2, BBH -- BigBenchHard, IFE -- IFEval, PQA -- PopQA, TQA -- TruthfulQA.
}
\label{tab:instruct_stages}
\end{table}

\begin{figure}[h]
\centering
\begin{minipage}{0.45\textwidth}
\centering
\footnotesize
\setlength{\tabcolsep}{3pt}
\renewcommand{\arraystretch}{1.08}
\begin{tabular}{cccc}
\toprule
Epochs & L.R. & Loss & Avg. Perf. \\
\midrule
2 & 1e-5 & sum & \bf{49.97} \\
3 & 4e-6 & sum & 49.76 \\
2 & 1e-5 & sum & 49.74 \\
2 & 1e-5 & sum & 49.59 \\
3 & 4e-6 & mean & 48.25 \\
2 & 2e-6 & mean & 48.18 \\
\bottomrule
\end{tabular}
\vspace{3pt}
\captionof{table}{Hyperparameter configurations tried for the 7B SFT checkpoint, all on the same dataset used in the final model. SFT models are trained with an effective batch size of 128, a linear learning rate schedule and a warmup up ratio of 0.3.}
\label{tab:sft_hypers}
\end{minipage}
\hfill
\begin{minipage}{0.45\textwidth}
\centering
\includegraphics[width=0.8\textwidth]{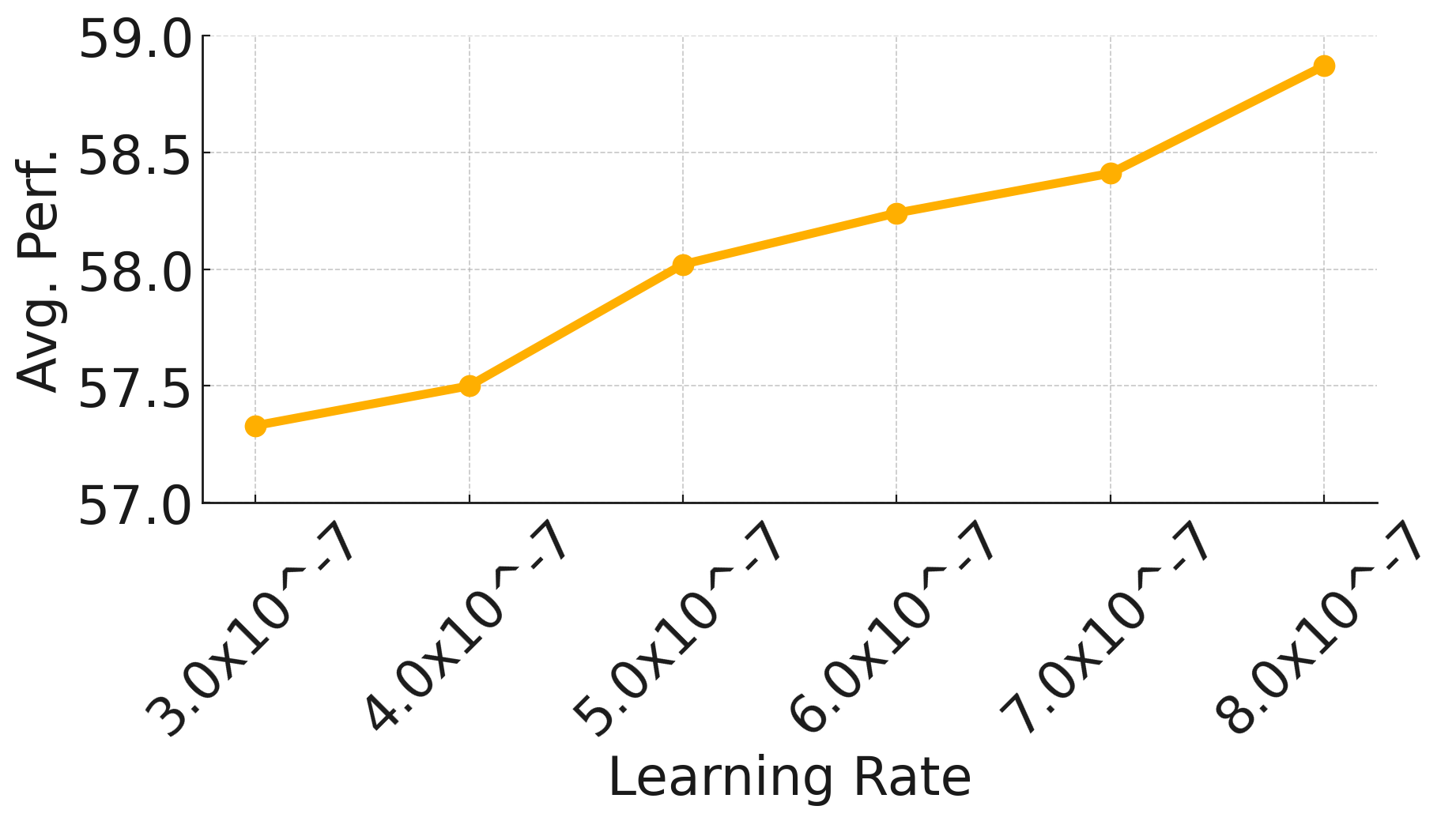}
\captionof{figure}{The average score for DPO checkpoints trained on a development SFT checkpoint on different learning rates. Avg does not include Safety.}
\label{fig:dpo_lrs_plot}
\end{minipage}
\end{figure}

\paragraph{Preference Finetuning (PreFT) with DPO} 
The core strategy of the \tulu pipeline for PreFT is building upon and scaling the UltraFeedback pipeline \citep{cui2023ultrafeedback} for generating synthetic preferences across data for our target domains.
We include on-policy data by sampling responses from some development \olmotoo SFT models at both 7B and 13B, with independent datasets for each.

From \tulu, we updated our model pool to only include models with permissible licenses as shown in Table~\ref{tab:pref-data} in the Appendix.
We made a minor shift from \tulu on the exact prompts used for DPO -- we obtain our prompts from several sources listed in Table~\ref{tab:prompt_sources}, resulting in datasets of 366.7k prompts for 7B and 377.7k prompts for 13B. 
Given this set of prompts, we generate responses from a pool of 20 models of different families and sizes.

To create synthetic preference data we use \texttt{GPT-4o-2024-08-06} as an LM judge \citep{zheng2023judging} and prompted it to rate completions based on helpfulness, truthfulness, honesty, and instruction-following aspects.
We then binarize the ratings across aspects by following Argilla's method\footnote{See \url{https://huggingface.co/datasets/argilla/ultrafeedback-binarized-preferences}.}: we get the average rating across all aspects, take the highest-rated completion as the chosen response, and sample from the remaining completions for the rejected response.

The 1B and 32B DPO models were trained with the same on-policy methodology.

\paragraph{Reinforcement Learning with Verifiable Rewards (RLVR)} 
RLVR is a novel finetuning technique used to target specific domains where prompts with verifiable answers can be constructed.
For example, with a math problem, the RL algorithm Proximal Policy Optimization (PPO)~\citep{schulman2017proximal} only receives a reward if the answer is correct.
For more details, see \citet{lambert2024tulu3}.

Following preference tuning, we trained 7B and 13B reward models using the on-policy 7B and 13B preference dataset. 
Next, we applied RLVR to the highest-performing 7B and 13B DPO checkpoints with a combined dataset comprising GSM8K, MATH training sets, and prompts with constraints from \citet{lambert2024tulu3}. For RLVR, we initialize PPO's value function from the corresponding RMs, which is shown to help improve average scores across evaluations~\citep{lambert2024tulu3}. 
After the initial RLVR training pass on the 13B model, we observe that its performance on GSM8K and MATH was lower than a previous development instruct model. 
Consequently, we perform two additional RLVR training iterations: first on the GSM8K training set, followed by the MATH training set.  
The models selected at the end of the RLVR stage constitute the final OLMo 2 Instruct models. 

\begin{figure}[h]
    \centering
    \begin{minipage}{0.8\textwidth}
        \centering
        \includegraphics[width=0.95\linewidth]{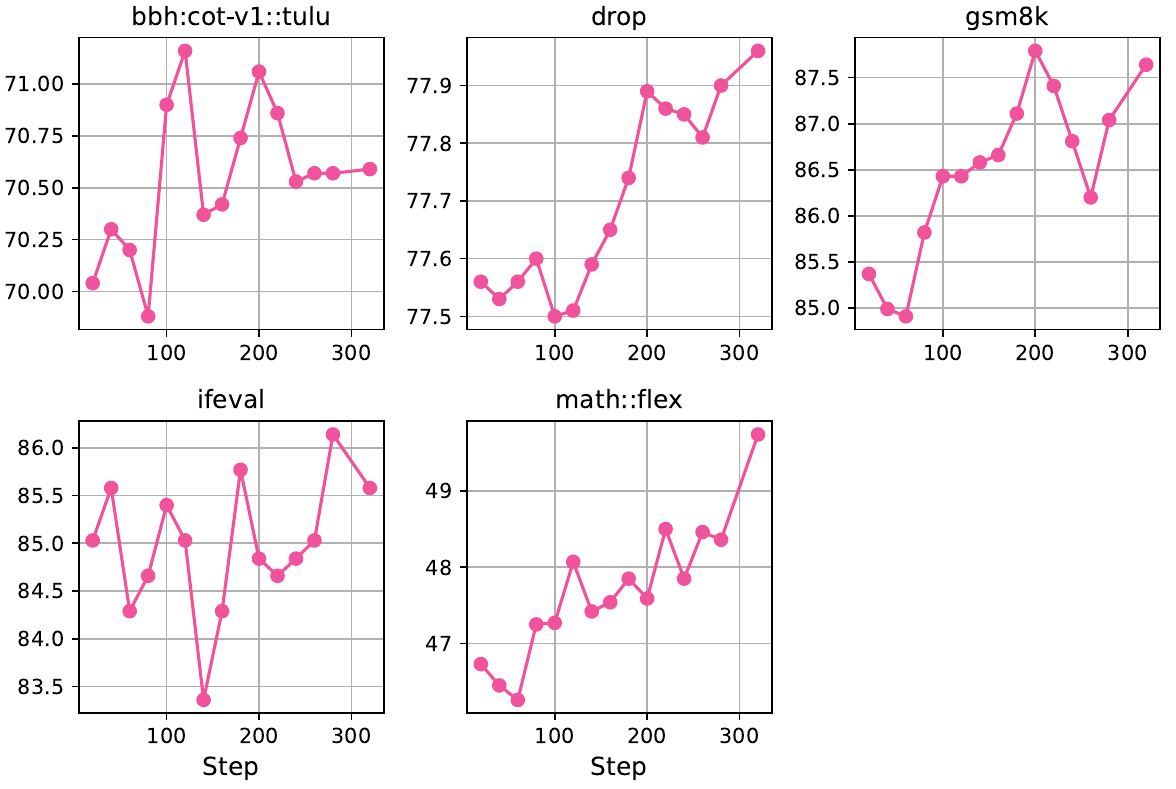}
    \end{minipage}
    \caption{
    The scores from core metrics our evaluation suites for OLMo-2-0325-32B-Instruct trained with RLVR. 
    We train OLMo-2-0325-32B-Instruct on the GSM8K, MATH, and prompts with constraints dataset mix to improve these scores. }
    \label{fig:the-rl-chart-32b}
\end{figure}

For the 1B and 32B model, we performed RLVR with Group Relative Policy Optimization (GRPO)~\citep{shao2024deepseekmath}, which forgoes the need for a reward model.
The evaluation metrics for this 32B model are shown in Fig.~\ref{fig:the-rl-chart-32b}.

\begin{figure}[h]
    \centering
    \begin{minipage}{0.98\textwidth}
        \centering
        \includegraphics[width=0.9\linewidth]{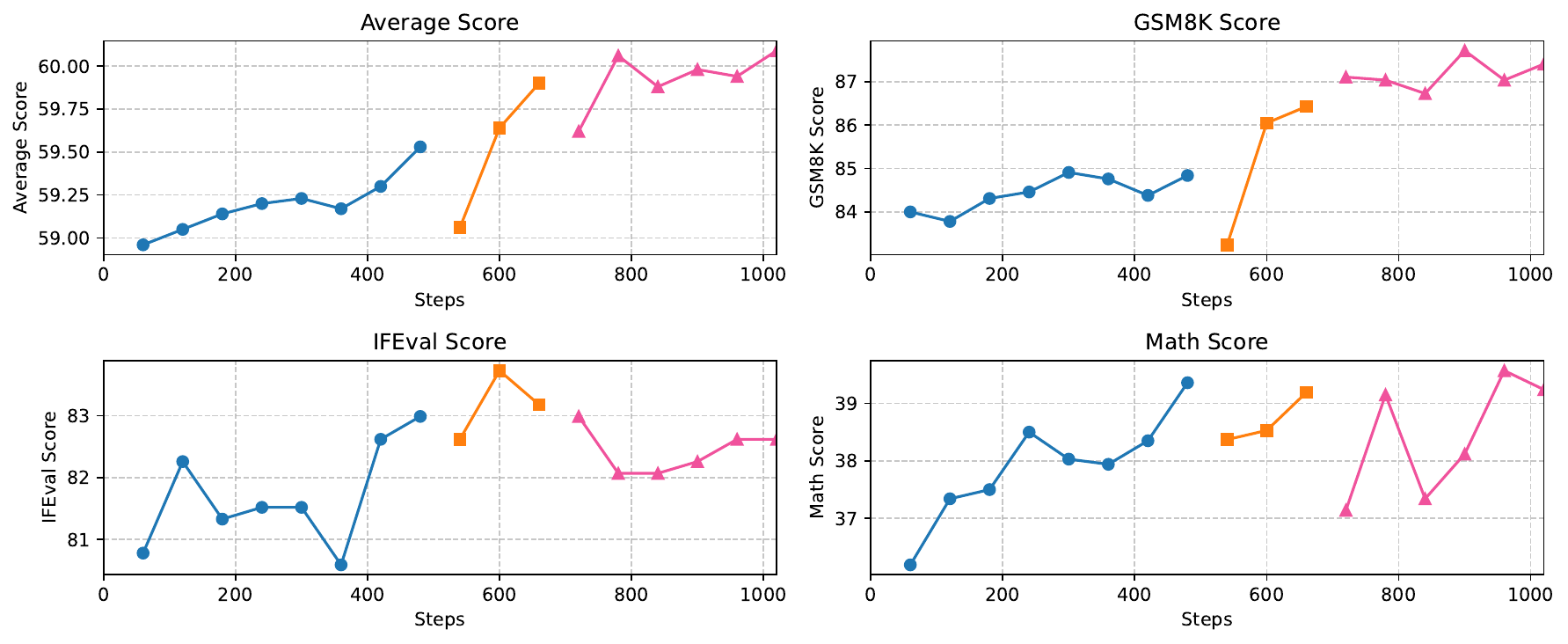}
    \end{minipage}
    {\cblock{31}{119}{180}} OLMo-2-1124-13B-RLVR1
    {\cblock{255}{127}{14}} OLMo-2-1124-13B-RLVR2
    {\cblock{240}{82}{156}} OLMo-2-1124-13B-Instruct (Final RLVR)
    \caption{
    The scores from our evaluation suites for OLMo-2-1124-13B-Instruct trained with RLVR. 
    We train OLMo-2-1124-13B-RLVR1 on the GSM8K, MATH, and prompts with constraints dataset mix, but noticed the GSM8K score was lower than expected. 
    We proceed with training OLMo-2-1124-13B-RLVR2 on GSM8K and observed higher GSM8K score. 
    Finally, we train OLMo-2-1124-13B-Instruct on just MATH and observe even higher GSM8K and MATH scores. 
    Note that the value function was re-initialized from the reward model in each RLVR run. 
    The full learning curves of each RLVR run can be found in Appendix~\ref{appendix:rlvr-13b}. }
    \label{fig:the-rl-chart-13b}
\end{figure}

\begin{figure}[h]
    \centering
    \begin{minipage}{0.98\textwidth}
        \centering
        \includegraphics[width=0.9\linewidth]{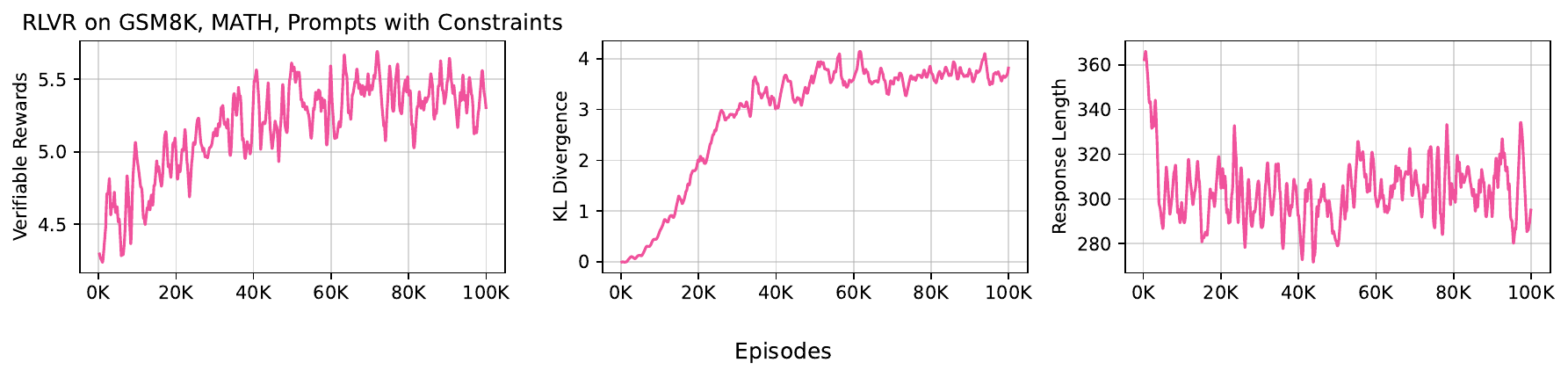}
    \end{minipage}
    \hfill
    \begin{minipage}{0.98\textwidth}
        \centering
        \includegraphics[width=0.9\linewidth]{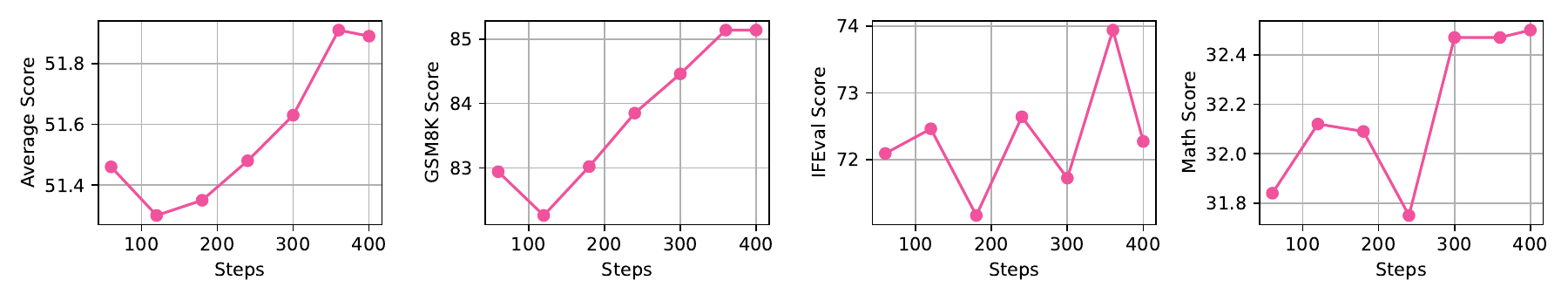}
    \end{minipage}
    {\cblock{240}{82}{156}} OLMo-2-1124-7B-Instruct 
    \caption{The top row shows the training curves of OLMo-2-1124-7B-Instruct on verifiable rewards, KL divergence, and response lengths. In the bottom row, the y-axes show the average scores across our evaluation suites and GSM8K, IFEval, and MATH Flex scores, respectively. Overall, we find RLVR increases not only the training rewards of our 7B models but also the downstream evaluations such as GSM8K.}
    \label{fig:the-rl-chart-7b}
\end{figure}

\paragraph{Hyperparameter selection} 
We perform the following hyperparameter tuning for the 7 and 13B models. 
At each stage we experiment with 1 random seed initially to arrive on a configuration and up to 4 with final hyperparameters. The final hyperparameters are marked with (${\textcolor{ai2pink}{\varheartsuit}}$):
\begin{enumerate}
    \item \textbf{SFT:} We sweep over learning rates $1 \times 10^{-5}$, $2 \times 10^{-5}$ (${\textcolor{ai2pink}{\varheartsuit}}$), $3 \times 10^{-5}$ for the 7B model and $1 \times 10^{-6}$, $4 \times 10^{-6}$, $5 \times 10^{-6}$ (${\textcolor{ai2pink}{\varheartsuit}}$), $7.5 \times 10^{-6}$, $8 \times 10^{-6}$ for the 13B model.
    \item \textbf{DPO:} We sweep over learning rates $5 \times 10^{-7}$, $6 \times 10^{-7}$, $7 \times 10^{-7}$, $8 \times 10^{-7}$ (${\textcolor{ai2pink}{\varheartsuit}}$ - 13B), and $1 \times 10^{-6}$ (${\textcolor{ai2pink}{\varheartsuit}}$ - 7B) for both the 7B model and 13B model.
    \item \textbf{RM:} We train with $3 \times 10^{-6}$ learning rate and 1 random seed for the 7B and 13B models, respectively.
    \item \textbf{RLVR:} We sweep over beta values 0.03, 0.05, 0.07 (${\textcolor{ai2pink}{\varheartsuit}}$ - 7B), and 0.1 (${\textcolor{ai2pink}{\varheartsuit}}$ - 13B). For 13B model, we also sweep over learning rates $3 \times 10^{-7}$ (${\textcolor{ai2pink}{\varheartsuit}}$ - 13B), $4 \times 10^{-7}$ (${\textcolor{ai2pink}{\varheartsuit}}$ - 7B). For 13B, we run this sweep on the best model at each RLVR stage.
\end{enumerate}

We conducted a hyperparameter sweep for SFT and DPO, using earlier development checkpoints, with results detailed in Table~\ref{tab:sft_hypers} and Figure~\ref{fig:dpo_lrs_plot}. A key finding was that OLMo 2 required significantly higher learning rates compared to the Llama 3.1 training recipe described by \citet{lambert2024tulu3}. Finally, the optimized hyperparameters for our final model are presented in Table~\ref{tab:sft_hypers} and Table~\ref{tab:hypers_rlvr}.

The post-training for the 32B model occurred after the release of the 7 and 13B models, so the hyperparameter selection proceeded independently. 
For SFT, we swept over a learning rate of $1\times10^{-6}$,$2\times10^{-6}$,$3\times10^{-6}$,$4\times10^{-6}$,$5\times10^{-6}$, with the best performance as $4\times10^{-6}$ where we ran one additional seed to compare performance.
For DPO, we swept over learning rates again, from $8 \times 10^{-7}$, $1 \times 10^{-6}$, $1.5 \times 10^{-6}$, $2 \times 10^{-6}$, $2.5 \times 10^{-6}$, and the best performance was $2 \times 10^{-6}$.
For RLVR, the 32B does not need a reward model due to the change to GRPO.
Beyond that, the final model was trained with a learning rate of $5 \times 10^{-7}$, with a KL beta of 0.1, and 16 samples  per prompt.

\paragraph{Evaluation of \olmotooinstruct} 
Following \tulu~\citep{lambert2024tulu3}, we evaluate \olmotooinstruct on five categories listed in Table~\ref{tab:eval-post}. 
Although \tulu uses six categories including code-related tasks, we exclude this category since code was not a target skill during the development of \olmotoo. 
For each of the remaining categories, we use the same evaluations as those used for developing the \tulu recipe.
Table~\ref{tab:eval-post} also shows the settings and metrics used for each of the evaluations. These match those recommended in~\citet{lambert2024tulu3} for the non-code categories.

Table~\ref{tab:instruct_stages} presents the performance of OLMo 2 Instruct variants across different training stages. 
A comparative analysis of \olmotooinstruct's performance against similarly-sized open models can be found in Table~\ref{tab:instruct_results}. Furthermore, Figures~\ref{fig:the-rl-chart-13b} and~\ref{fig:the-rl-chart-7b} present the training trajectories and key performance metrics for the 13B and 7B models, respectively. 

The \olmotooinstruct models demonstrate comparable performance to leading open-weight models in the field. Specifically, OLMo 2 13B Instruct achieves results approaching those of Qwen 2.5 14B Instruct while surpassing both \tulu 8B and Llama 3.1 8B Instruct in performance benchmarks. The RLVR stage also demonstrated consistent effectiveness across both model scales, leading to notable improvements in evaluation metrics in tandem with increasing the training reward signal.

Finally, we evaluate OLMo 2-Instruct on the unseen evaluation suite from \citet{lambert2024tulu3} without the code evaluation tasks. The Instruct scores on the unseen evaluation suite are shown in Table~\ref{tab:instruct-unseen}.

\begin{table}[h]
\centering
\begin{minipage}{0.48\textwidth}
    \footnotesize
    \centering
    \setlength\tabcolsep{3pt}
    \begin{tabular}{>{\raggedright}p{0.35\linewidth}p{0.55\linewidth}}
        \toprule
        \textbf{Hyperparameter} & \textbf{RLVR value}  \\
        \midrule
        \rowcolor{ai2offwhite} Learning rate & $3 \cdot 10^{-7}$ for 13B; $4 \cdot 10^{-7}$ for 7B \\
        \rowcolor{ai2offwhite} Effective batch size & 248 for 13B; 224 for 7B \\
        \rowcolor{ai2offwhite} 
        KL penalty coef.  ($\beta$) &
        {0.1 for first and final 13B; 0.03 for second 13B; 0.05 for 7B} \\
        \rowcolor{ai2offwhite} Max total episodes & 200,000 for 13B; 100,000 for 7B \\
        Discount factor $\gamma$ & 1.0 \\
        General advantage estimation $\lambda$ & 0.95 \\
        Mini-batches $N_\text{mb}$ & 1 \\
        PPO update iterations $K$ & 4 \\
        \bottomrule
    \end{tabular}
\end{minipage}
\hfill
\begin{minipage}{0.48\textwidth}
    \footnotesize
    \centering
    \setlength\tabcolsep{3pt}
    \begin{tabular}{>{\raggedright}p{0.60\linewidth}p{0.3\linewidth}}
        \toprule
        \textbf{Hyperparameter} & \textbf{RLVR value}  \\
        \midrule
        PPO's clipping coefficient $\varepsilon$  & 0.2 \\
        Value function coefficient $c_1$ & 0.1 \\
        Gradient norm threshold & 1.0 \\
        Learning rate schedule & \textit{linear} \\
        Generation temperature & 1.0 \\
        Max token length & 2,048 \\
        Max prompt token length & 2,048 \\
        {Penalty reward for no EOS token} & $-10.0$ \\
        Response length & 2,048 \\
        Warm up ratio ($\omega$) & 0.0 \\
        \bottomrule
    \end{tabular}
\end{minipage}
\sethlcolor{ai2offwhite} %
\caption{The hyperparameters of PPO used for optimizing against the verifiable reward function with RLVR.
Hyparameters with different settings for the 7B and 13B parameter models are \hl{highlighted}.}
\label{tab:hypers_rlvr}
\end{table}

\section{Deep Dive: \diveInfra}
\label{sec:diveInfra}

LM training is famously compute intensive.
Training large models requires state-of-the-art hardware, and a lot of work goes into making it run efficiently.
Gains in efficiency can be translated into higher token counts or more parameters, directly affecting the quality of the final model.
GPUs are at the core of this infrastructure, investment in other processes and systems is required to make them perform at peak efficiency.
Data centers need high-speed interconnect between compute nodes to make sure expensive GPUs never have to wait for data to arrive.
Training jobs need access to large amounts of fast, reliable storage for access to training data.
GPUs have higher failure rates than most other hardware, and a single training run might require thousands of them at the same time, making effective monitoring and replacement policies a necessity.
This section provides details about our hardware and software investments to support \olmotoo workloads.

\subsection{Clusters}
\label{sec:clusters}

\olmotoo is trained on two Ai2 clusters, Jupiter and Augusta. Despite hardware and architectural differences, both clusters provided sufficient training throughput. Beaker, Ai2's workload management system, allows researchers to migrate workloads from one cluster to another, and both the 7B and 13B variants were trained partially on both clusters, with the bulk of the 7B training on Jupiter, and the bulk of 13B training on Augusta.

\subsubsection{Jupiter}

Jupiter is a 128-node GPU cluster located in Austin, Texas. 
It is operated by Cirrascale Cloud Services\footnote{\href{https://www.cirrascale.com/}{\path{cirrascale.com}}}.

\paragraph{Compute} It consists of 1,024 NVIDIA H100 GPUs, each with 80GB HBM3 running at 700W. The GPUs are spread across 128 servers with 2x Intel Xeon Platinum 8468 CPUs, 2 TB of DDR5 system memory, and 18 TB of local NVMe storage.

\paragraph{Storage} The servers are connected via a 800 Gbps local network to a WEKA high performance storage cluster\footnote{\href{https://www.weka.io/}{\path{weka.io}}}. This cluster has 1 PB of NVMe SSD storage with 11 physical servers, and 5 PB of HDD storage spread across 12 hosts. The Jupiter GPU servers have two bonded 25 Gbps Mellanox ethernet cards each, providing a total of 50 Gbps of throughput per host. In benchmarks, we reach 761 Gbps of read/write throughput using 64 client machines. 

\paragraph{Interconnect} Cross-node GPU communication is provided via RDMA over InfiniBand and a 2-Tier Rail Optimized \citep{wang2023rail}, balanced, full-bisected network. Each physical server has eight 400 Gbps InfiniBand cards, providing a maximum total throughput per host of 3200 Gbps. This setup allows Ai2 to run dozens of distributed training workloads simultaneously on the same cluster without topological scheduling.

\paragraph{Cooling} The Jupiter servers are racked in \textit{Dynamic Density Cabinets}\footnote{\href{https://www.cirrascale.com/products-and-services/cabinet-technologies}{\path{cirrascale.com/products-and-services/cabinet-technologies}}}. Each cabinet includes 5 servers with dedicated cooling and power. Each cabinet is a closed system, circulating air through an overhead compartment where it is cooled via heat transfer to water. This approach allows the datacenter to achieve a power usage efficiency (PUE) of 1.2. 
Under heavy utilization, our H100 GPUs reach a peak temperature of 75°C; average GPU temperatures are between 60°C and 65°C.

\subsubsection{Augusta Cluster}

The Augusta cluster is a 160-node GPU cluster provided by Google Cloud. The physical servers are located in Council Bluffs, Iowa.

\paragraph{Compute} The cluster is made up of A3 Mega virtual machines\footnote{\texttt{a3-megagpu-8g}, more information at \href{https://cloud.google.com/compute/docs/accelerator-optimized-machines}{\path{cloud.google.com/compute/docs/accelerator-optimized-machines}}}, each with 8 NVIDIA H100 GPUs.

\paragraph{Storage} Augusta workloads use Google Cloud Storage for speeds up to 1 GB/s per VM. We ensure portability by abstracting storage interactions into common libraries supporting both file- and object-based APIs.

\paragraph{Interconnect} Each GPU has a dedicated Ethernet NIC. Fast cross-node GPU communication is achieved using GPUDirect-TCPXO, gVNIC, and compact node placement. This arrangement takes advantage of Google's \href{https://cloud.google.com/blog/topics/systems/the-evolution-of-googles-jupiter-data-center-network}{Jupiter data center network technology} and the \href{https://cloud.google.com/blog/products/compute/titanium-underpins-googles-workload-optimized-infrastructure}{Titanium system} with tiered offloading and full-bandwidth reconfigurable optical links. This provides bandwidth similar to non-blocking network fabrics.

\paragraph{Cooling} The Augusta servers are air-cooled and \href{https://www.google.com/about/datacenters/efficiency/}{the Iowa campus in which they are located reported} a trailing twelve-month power usage efficiency (PUE) of 1.12.

\subsection{Beaker}

\olmotoo workloads were scheduled using Beaker~\citep{beaker2022}, a custom workload management system. 
Beaker benefited \olmotoo in two key ways:

\paragraph{Portability} Beaker's architecture can take advantage of GPUs across 3 different data centers with minimal code changes.  
It can be run anywhere running a single Linux daemon that is packaged as a statically linked binary. 
Typically, workloads can be moved from one location to another by changing a single line of code.

\paragraph{Isolation} Beaker workloads are containerized, providing some isolation guarantees. 
This allows \olmotoo workloads to run simultaneously with other jobs on the same cluster, each with unique environments and dependencies, with minimal conflicts. Notably, the Beaker executor allocates host resources in a fashion that minimizes (but doesn't completely avoid) performance problems caused by noisy-neighbors.
Containers further capture software dependencies and the runtime details of workloads. 
This helps run repeatable experiments, and makes it possible to replay old results even months after they happened. 
This stands in contrast to the more common Slurm-based setup where all workloads, whether they relate to \olmotoo or not, share the same underlying operating system, CUDA libraries, and environment resulting in instability that makes experiments unreproducible after system changes.

Beaker also made it possible for us to take advantage of new compute sources that became available throughout the evolution of the project. 
Its operational simplicity made it possible for a small team of operators to quickly onboard new sources of compute.

\subsection{Stability and Operations}

Both clusters required an initial testing and burn-in period, during which we discovered and remedied problems ranging from ill-seated cables to an improper ordering of the compute nodes in the NCCL library.
These periods required close collaboration with the respective hardware vendors, and both were indispensable during this process.
After this period, both clusters operate approximately at the same level of reliability.

\paragraph{GPU health checks} Beaker executes a simple program prior to running workloads on the assigned GPUs. The program attempts to multiply two tensors. When failures occur, Beaker cordons the associated host and reschedules the workload, quarantining the errant node before introducing instability. This helped reduce interruptions requiring manual attention, and made it viable to configure training jobs to simply restart themselves when encountering an error, safe in the knowledge that they would be moved to the working set of compute nodes.

\paragraph{Cordoning} Beaker supports cordoning nodes as an override for the automatic health checks. A cordoned node is removed from scheduling and gets flagged for repair. In this way, Beaker effectively crowdsources the identification of bad nodes among all the users of the cluster.

\paragraph{Active monitoring} Beyond these two methods, Beaker performs industry-standard monitoring and automatic alerting based on cluster telemetry. The team has operational processes in place for responding to system issues, enabling it to resolve issues promptly.

\subsection{Maximizing hardware utilization}
Ai2's hardware infrastructure (\S\ref{sec:clusters}) has to be complemented by good model training software that gets the most out of the available resources.
Increased efficiency not only lets us train larger models for more tokens, but it also improves the environmental impact of model training~(\S\ref{sec:env-impact}), and raises experimental velocity.
Further, \olmo is not the only Ai2 project, and being responsible with our resource use minimizes the disruption that large model training causes for other teams.

Below we describe several PyTorch optimizations\footnote{All of these techniques and more are implemented in the open-source OLMo-core library, the 2nd generation of the OLMo training codebase. OLMo-core is available at \href{https://github.com/allenai/OLMo-core}{\github\path{allenai/OLMo-core}}.} that had a big impact towards reducing training time of LMs on our infrastructure without any apparent loss in the speed of convergence.

\paragraph{Taking advantage of compilation}
\texttt{torch.compile()} is a function in PyTorch\footnote{\href{https://pytorch.org/tutorials/intermediate/torch_compile_tutorial.html}{\path{pytorch.org/tutorials/intermediate/torch_compile_tutorial}}} that will compile native PyTorch modules and functions into optimized kernels, resulting in significant throughput improvements and GPU memory savings by avoiding the Python overhead associated with calling individual PyTorch operations in sequence, and by reducing the number of reads and writes that must occur on the GPU.
As such, when \texttt{torch.compile()} is used properly, it can effectively match the performance of hand-crafted kernels in many cases without the additional complexity and engineering effort \citep{PyTorch2}.

\paragraph{Minimizing host-device syncs}

\defcitealias{pytorch2024cuda}{PyTorch: CUDA Semantics}
\begin{displayquote}
    \textit{By default, GPU operations are asynchronous. A function call that uses the GPU is enqueued to a particular device, but not necessarily executed until later. This allows the system to execute more computations in parallel, including operations on CPU or other GPUs...} \\
    -- \citetalias{pytorch2024cuda}
\end{displayquote}

Any time the training code forces a \textit{host-device sync}, no more operations can be enqueued until all operations currently in the queue complete.
These synchronization points will hinder performance, and it is easy to unintentionally introduce them.

A surprising number of operations cause host-device syncs:\nopagebreak
\begin{enumerate}
    \item 
    {\bf{Synchronously copying a tensor from CPU to GPU}}
    (e.g., with \texttt{tensor.to}\texttt{(device="cuda")}) will force a host-device sync.
    This can be avoided by copying the tensor \textit{asynchronously}
    (e.g., with \texttt{tensor.to}\texttt{(device="cuda",}~\texttt{non\_blocking=True)}).
    
    \item
    {\bf{Copying a tensor from GPU to CPU}} cannot safely be done asynchronously, so GPU~$\rightarrow$~CPU data transfer should be avoided whenever possible.
    Seemingly innocuous code can cause GPU~$\rightarrow$~CPU data transfer, and therefore host-device syncs, such as \texttt{print}-ing a CUDA tensor or an ``\texttt{if ...:}" block that depends on how a CUDA tensor resolves to a boolean value.
    
    \item
    {\bf{Specific PyTorch operations}} like \texttt{masked\_select()} may unexpectedly cause a host-device sync\footnote{\href{https://github.com/pytorch/pytorch/issues/12461}{\path{github.com/pytorch/pytorch/issues/12461}}}.
\end{enumerate}

Host-device syncs can be detected by calling \texttt{torch.cuda.set\_sync\_debug\_mode("warn")} before starting the training loop.
This will cause PyTorch to emit a warning whenever a host-device sync occurs.
This happens on a best-effort basis.
Some syncs may still be missed.

\paragraph{Asynchronous bookkeeping with a separate backend}
A typical training loop involves periodic ``bookkeeping'' operations like logging metrics and saving checkpoints.
While these operations may be relatively fast, their aggregate cost over the course of a training run can be significant.
These operations also usually involve host-device syncs. 
For example, a training metric like cross-entropy loss is the result of computations that occur on the GPU, and it is materialized first as a CUDA tensor; 
therefore, logging that metric to the console forces a synchronization point.

Many of these operations are essential and cannot be avoided, but it is possible to minimize the time they spend blocking the training loop by performing most of this bookkeeping work asynchronously, in a separate thread.
However, the PyTorch NCCL backend is not thread safe.
To work around this problem, we set up a separate backend that does not rely on NCCL (like GLOO), and use it exclusively for bookkeeping operations.
The bookkeeping workflows could then look like this:

\begin{enumerate}
    \item \textit{For metric collection and logging:}
        Decide on the interval in which to log metrics.
        Since this involves a host-device sync, it should not be done on every training step.
        More commonly, metrics are logged every 10 or every 50 steps.
        During every step, metrics are computed and stored in a GPU tensor on their original devices.
        Only when it is time to log metrics do we copy them to the CPU (causing a host-device sync), and then pass them to the bookkeeping thread, which uses its own PyTorch backend to aggregate the metrics and log them.
    \item \textit{For checkpointing:}
        A similar workflow can be used for checkpointing.
        When it is time to save a checkpoint, the trainer makes a copy of the model and optimizer state in CPU memory (causing a host-device sync). Then it passes the copy to the bookkeeping thread, which assembles the model from the model shards that are stored on each compute node, and saves it to disk, while the main thread can continue training\footnote{See \href{https://pytorch.org/blog/reducing-checkpointing-times/}{\path{pytorch.org/blog/reducing-checkpointing-times}} for a concrete example.}.
\end{enumerate}

In both cases, the only impact on training time is one host-device sync, and the time it takes to copy data from the GPU to the CPU.
The frequency of each event can be configured, and the overall impact on training time is negligible.

\begin{figure*}[h]
\vspace{-0.5em}
    \begin{minipage}[c]{0.45\textwidth} \centering
        \caption{The training throughput in tokens per second (TPS) per device over the course of 1000 steps for two OLMo-1B models on 8 nodes, one with automatic garbage collection (GC), the other with collection done \textit{manually}
        at intervals set within our training codebase.
        With automatic garbage collection, training throughput is slower and less stable, often becoming worse as the run progresses.}
        \label{fig:gc-perf}
    \end{minipage} \hfill
    \begin{minipage}[c]{0.45\textwidth}\centering
        \includegraphics[width=\textwidth, trim={0 0 0 35px}, clip]{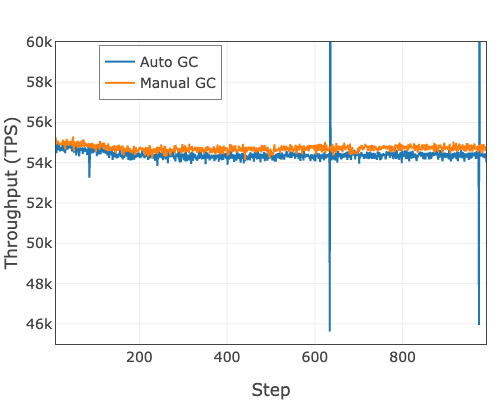}
    \end{minipage}
\vspace{-1em}
\end{figure*}

\paragraph{Explicit Python garbage collection} 
During training, the default Python garbage collector periodically runs a collection.
In a distributed setting, with thousands of training processes that are expected to run in lock-step with each other, nothing enforces that these garbage collections happen at the same time on every process.
Since distributed training can only proceed as fast as the slowest process, this causes a noticeable decrease in average training time per step as well as an increase in variability (Figure~\ref{fig:gc-perf}).
Both worsen as the number of processes increases.

To work around this problem, the \olmotoo trainer disables automatic garbage collection (e.g., by calling \texttt{gc.disable()}\footnote{\href{https://docs.python.org/3/library/gc.html\#gc.disable}{\texttt{docs.python.org/3/library/gc\#gc.disable}}}).
Then, it runs garbage collection explicitly at regular intervals, triggered at the same time in each process (e.g. by calling \texttt{gc.collect(1)}\footnote{\href{https://docs.python.org/3/library/gc.html\#gc.collect}{\path{docs.python.org/3/library/gc\#gc.collect}}}).

\begin{table}[h]
\centering
\adjustbox{max width=\linewidth}{
\begin{tabular}{l c c c c c c c c c}
\toprule
\textbf{Model} & \textbf{\makecell{Total GPU\\Power (MWh)}} & \textbf{\makecell{Power Usage\\Effect.}} & \textbf{\makecell{Carbon\\Intensity}} & \textbf{\makecell{Carbon\\Emissions}} & \textbf{\makecell{Water Usage\\Effect.}} & \textbf{\makecell{Total Water\\Usage (kL)}} \\
\midrule
Llama 2 7B & 74 & 1.1 & - & 31 & \underline{1.29 - 4.26} & \underline{105 - 347} \\
Llama 3.1 8B & 1,022 & 1.1 & - & 420 & \underline{1.29 - 4.26} & \underline{1,450 - 4,823} \\
\rowcolor{ai2offwhite}OLMo 7B & 104 & 1.1 & 0.610 & 70 & 4.26 & 487 \\
\rowcolor{ai2lightpink}\olmotoo 7B & 131 & 1.2 & 0.332 & 52 & 1.29 & 202 \\
\rowcolor{ai2lightpink}\olmotoo 13B & 257 & 1.12 & 0.351 & 101 & 3.10 & 892 \\
\bottomrule
\end{tabular}
}
\vspace{3pt}
\caption{CO\textsubscript{2} emissions and water consumption during pretraining. We estimate the total carbon emissions and water consumption for our new models using PUE information from our data center providers, carbon intensity data and WUE from the local grid for each data center, and total power consumption from time series data logged throughout training. Numbers for Llama 2 \citep{touvron2023llama}, Llama 3 \citep{dubey2024llama}, and the original OLMo \citep{Groeneveld2024OLMoAT} are taken from their respective papers. We also show {\bf simulated} water consumption for Llama 2 and 3, showing a range of water usage numbers using the lowest and highest WUE values for OLMo models.}
\label{tab:environmental_impact}
\end{table}

\subsection{Environmental Impact}
\label{sec:env-impact}

Following our analysis in \citet{Groeneveld2024OLMoAT} and previous literature \citep{patterson2021carbonemissionslargeneural,dodge2022measuringcarbonintensityai,luccioni2022estimatingcarbonfootprintbloom,li2023makingaithirstyuncovering}, we estimate the environmental impact of training our final models by first calculating the total energy consumed during pretraining, and multiplying it by the carbon intensity of the local grid to estimate the amount of carbon released. We additionally extend our previous analysis to also estimate water consumption, calculated by multiplying the power consumed by the water usage efficiency of both the power generation and the cooling hardware. As in \citet{Groeneveld2024OLMoAT}, we emphasize that while our reporting is standard practice, it does not account for other environmental impacts such as embodied emissions and water consumption of the hardware during manufacturing, transportation, and eventual disposal, and other lifetime operational impacts such as deployment and inference, and thus our estimates should be viewed as lower bounds. We report detailed results for our models in Table~\ref{tab:environmental_impact}.

As in \citet{Groeneveld2024OLMoAT}, we calculate the total power consumption for each model by measuring the power consumption of an individual node every 25ms, calculating the average consumption throughout training, and multiplying by the total number of nodes. 
We then multiply this quantity by the power usage effectiveness (PUE) factor for the data center we use to train a model to account for the overall energy efficiency of the data center. 
As the majority of training for \olmotoo 7B is done on the Jupiter cluster, we use Jupiter's efficiency metrics for our analysis of the 7B model.
\olmotoo 13B is trained on Augusta; therefore, we use its efficiency metrics instead.
We estimate consumption at about \textbf{391 MWh of energy} by pretraining \olmotoo 7B and 13B.

To calculate carbon emissions, we multiply the total power consumption by a carbon intensity factor based on the physical location of each data center, measured in kg CO\textsubscript{2} per kWh. 
The Jupiter cluster is powered by Austin Energy, which most recently reported a carbon intensity of 0.332 kg CO\textsubscript{2} per kWh.\footnote{\href{https://web.archive.org/web/20230802220143/https://austinenergy.com/-/media/project/websites/austinenergy/commercial/carbonemissionscalculator.pdf}{\path{austinenergy.com/-/media/project/websites/austinenergy/commercial/carbonemissionscalculator.pdf}}} 
The Augusta cluster is located in Iowa, and the state of Iowa has an average carbon intensity of 0.352 kg CO\textsubscript{2} per kWh\footnote{\href{https://web.archive.org/web/20241127181309/https://www.eia.gov/electricity/state/iowa/}{\path{www.eia.gov/electricity/state/iowa}}}, which we use for our calculations. We estimate that training our latest models emitted about \textbf{154~tCO\textsubscript{2}eq}.
\begin{equation*}
    CO_2\hspace{3pt}\text{Emissions} = P_{\text{GPU}} \cdot \text{PUE} \cdot \text{Carbon Intensity}
\end{equation*}
To calculate water consumption, we multiply the total power consumption by the water usage effectiveness (WUE) of both the offsite power generation as well as the onsite cooling hardware. 
Both clusters use highly efficient, closed-loop cooling hardware, so we assume a WUE\textsubscript{onsite} of 0 liters per kWh. 
Following~\citet{wriestimatingwater}, we assume a WUE\textsubscript{offsite} of 1.29 L per kWh for our Jupiter cluster and 3.10 L per kWh for our Augusta cluster. 
We estimate that training our latest models consumed about \textbf{1.1 million liters of water}.
\begin{equation*}
    \text{Water Consumption} = P_{\text{GPU}} \cdot \text{PUE} \cdot (\text{WUE}_{\text{onsite}} + \text{WUE}_{\text{offsite}})
\end{equation*}
Though we aim to report a comprehensive analysis of the environmental impact of training our models, we emphasize that this is a lower bound on the total cost of developing large models. 
In an upcoming paper \citep{olmo-co2}, we will provide more comprehensive analysis covering energy, emissions, and water consumption throughout model development, pretraining, and deployment.

\section*{Conclusion}
We introduce \olmotoo and \olmotooinstruct, a family of fully open 7B, 13B and 32B parameter language models trained on up to 6T tokens.
Both the base and instruct models are competitive with other open-weight models in their size categories such as Qwen 2.5, Gemma 2, and Llama 3.1.
We detail the substantial contributions required to build competitive language models---many of which are different from the original \olmo---including stable infrastructure, architecture improvements for stability, innovations in late-stage training data, the latest post-training techniques, and many more details.
We release all training and evaluation code, datasets, checkpoints, and logs required to reproduce and expand on the models.
\olmotoo marks continued progress in open-source language models, building a new ecosystem for research, one where new training methods and techniques need to be understood and shared.

\phantomsection
\section*{Author Contributions}
\label{sec:contrib}

A successful team project like \olmo{} would not be possible without the fluid contributions of many teammates across formal team boundaries. 
As not all of these can be captured, we indicate each authors' primary contributing role in \olmotoo.
Authors are listed in alphabetical order: 

\begin{itemize}
    \item For base model development, including training and data curation: Shane Arora, Akshita Bhagia, Christopher Clark, Allyson Ettinger, Dirk Groeneveld, Yuling Gu, David Heineman, Matt Jordan, Jiacheng Liu, Kyle Lo, William Merrill, Tyler Murray, Jake Poznanski, Dustin Schwenck, Luca Soldaini, Oyvind Tafjord, David Wadden, and Pete Walsh.

    \item For instruct model development, including training and data curation: Faeze Brahman, Pradeep Dasigi, Nouha Dziri, Yuling Gu, Shengyi Huang, Hamish Ivison, Nathan Lambert, Saumya Malik, Lester James V. Miranda, Jacob Morrison, Valentina Pyatkin, Oyvind Tafjord, and Christopher Wilhelm.

    \item For operational support, including program management, legal guidance, release process, and more: Taira Anderson, David Atkinson, Crystal Nam, and Aman Rangapur.

    \item For Ai2 cluster setup and support: Michal Guerquin, Michael Schmitz, Sam Skjonsberg, and Michael Wilson

    \item For mentorship and advising: Ali Farhadi, Hannaneh Hajishirzi, Pang Wei Koh, Noah A. Smith, and Luke Zettlemoyer.
\end{itemize}

Authorship for this work was determined by those making direct contributions to the \olmotoo models, related artifacts, and their release. Core contributors are recognized for their sustained, significant contributions critical to the success of the \olmotoo project.

\section*{Acknowledgments}
\label{sec:acks}

This work would not be possible without the support of our colleagues at Ai2:

\begin{itemize}
    \item We thank Ben Bogin, Tim Dettmers, Ananya Harsh Jha, Ani Kembhavi, Matt Deitke, Ian Magnusson, Sewon Min, Niklas Muennighoff, Yizhong Wang, Alexander Wettig, and Valentin Hofmann for helpful research discussions and sharing of relevant findings across related projects.
    \item We thank Taylor Blanton, Byron Bischoff, Yen-Sung Chen, Arnavi Chheda, Jesse Dodge, Karen Farley, Huy Tran, Eric Marsh, Chris Newell, and Aaron Sarnat for building the Ai2 Playground for model demos.
    \item We thank Yoganand Chandrasekhar, Johann Dahm, Fangzhou Hu, and Caroline Wu for their work on the Ai2 cluster.
    \item We also thank others at Ai2 for many indirect contributions to the project: Robert Berry, Alex Buraczynski, Jennifer Dumas, Jason Dunkelberger, Rob Evans, David Graham, Regan Huff, Jenna James, Rodney Kinney, Bailey Kuehl, Sophie Lebrecht, Jaron Lochner, Carissa Schoenick, Will Smith, Sruthi Sreeram, Brooke Vlahos, Alice Wang, Caitlin Wittlif, Jiangjiang Yang.
\end{itemize}

We also appreciate conversations with and feedback from Cody Blakeney, Mansheej Paul, Jonathan Frankle, Armen Aghajanyan, Akshat Shrivastava, Mike Lewis, and John Schulman.

\olmotoo would not have been possible without the support of many other institutions. 
In particular, we thank Google for their support in setting up the training environment for \olmotoo and to Cirrascale for their on-going support of Ai2's cluster.
We also acknowledge the National Artificial Intelligence Research Resource (NAIRR) Pilot and Microsoft Azure for providing inference credits in support of this project.

\begin{textblock*}{100mm}(\dimexpr\paperwidth-100mm+6cm,\dimexpr\paperheight-1.92cm)
    \nointerlineskip
    \includegraphics[height=1.50em]{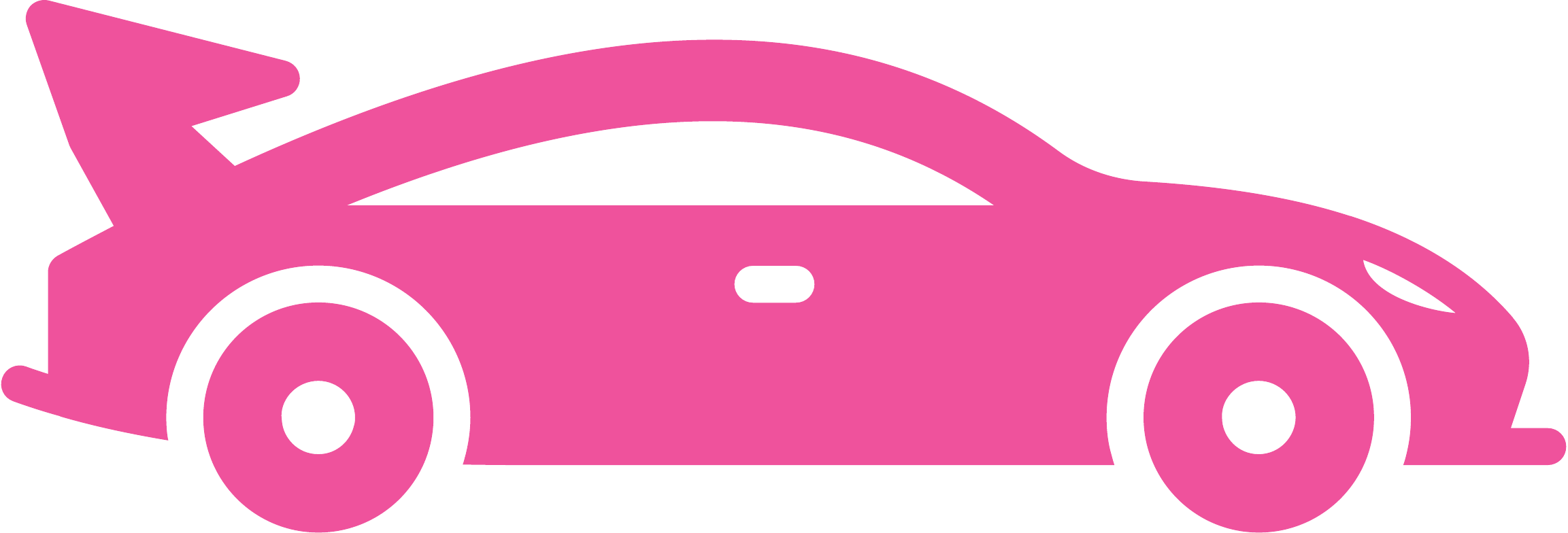}
\end{textblock*}

\clearpage
\bibliographystyle{abbrvnat}
\bibliography{neurips_2023}

\clearpage

\appendix

\section{\olmotoo Evaluation Framework}

We evaluate \olmotoo using OLMES, a unified, standardized evaluation suite and toolkit\footnote{The OLMES (Open Language Model Evaluation System) framework can be found at \href{https://github.com/allenai/olmes}{\path{github.com/allenai/olmes}}} to guide the development and assess performance of language models. 

\subsection{Base Model Eval}
\label{app:eval-base}

\olmo base models are evaluated on 11 tasks, consisting of 5 multiple-choice tasks, 2 generative tasks, and 4 additional held-out tasks not utilized during model development. 
See Table~\ref{tab:task-details} for the list of tasks along with details of the task formulations following the principles of the OLMES standard \citep{olmes}, described further below.

\begin{table*}[h]
  \centering

\begin{small}
\begin{tabular}{llllll}
\toprule
\bf{task} & \bf{split} & \bf{\# inst (total)} & \bf{\# shots} & \bf{metric} & \bf{reference}\\
\midrule
\rowcolor{midgrey}\multicolumn{6}{c}{\textbf{\textit{Multiple-choice tasks}}} \\
ARC-Challenge (ARC\_C) & Test &  1172 & 5 & pmi & \citep{clark2018think} \\
BoolQ & Val &  1000 (3270) & 5  & none & \citep{clark-etal-2019-boolq}\\
HellaSwag (HSwag) & Val &  1000 (10042) & 5 & char & \citep{zellers-etal-2019-hellaswag}\\
MMLU$^\dagger$ & Test &  14042 & 5 & char & \citep{hendryckstest2021}\\
WinoGrande (WinoG) & Val &  1267 &  5 & none & \citep{Sakaguchi_Le_Bras_Bhagavatula_Choi_2020}\\
\rowcolor{midgrey}\multicolumn{6}{c}{\textbf{\textit{Generative tasks}}} \\
DROP & Val &  1000 (9536) & 5 & F1 & \citep{dua-etal-2019-drop}\\
Natural Questions (NatQs) & Val &  1000 (3610) & 5 & F1 & \citep{kwiatkowski-etal-2019-natural} \\
\rowcolor{midgrey}\multicolumn{6}{c}{\textbf{\textit{Held-out tasks}}} \\
AGIEval English & Test &  2646 & 1 & MCF & \citep{zhong-etal-2024-agieval} \\
GSM8K & Test & 1319  & 8 (CoT) & EM & \citep{cobbe2021trainingverifierssolvemath}\\
MMLU-Pro & Test &  12032 & 5 & MCF & \citep{wang2024mmlu} \\
TriviaQA & Val &  7993 & 5 & F1 & \citep{joshi-etal-2017-triviaqa}\\
\bottomrule
\end{tabular}
\end{small}
  \caption{Details of OLMES benchmarks used in \olmotoo evaluation, with standardized choices of dataset split, number of instances to use, along with total number if sampling was used. For multiple-choice tasks, when using the Cloze/Completion Formulation (CF), the ``metric'' column specifies which normalization scheme to use. Following the OLMES standard, we evaluate each model using both the MCF (Multiple-Choice Formulation) and CF formulations, and the best performing one is used. For efficiency reasons, we limit MMLU and held-out multiple-choice evaluations to MCF only as all the relevant models strongly prefer that format for these tasks.}
  \label{tab:task-details}
\end{table*}

\paragraph{Multiple-choice tasks} We use the formulation of the 10 multiple-choice tasks defined in the OLMES evaluation standard~\citep{olmes}. OLMES (Open Language Model Evaluation Standard) is a set of principles and associated standard (with a reference implementation in the OLMES system framework) for reproducible LM evaluations that is open, practical, and documented, providing recommendations guided by experiments and results from the literature~\citep{biderman2024lessons,eval-harness}. For multiple-choice tasks it is designed to support comparisons between smaller base models that require the cloze/completion formulation of multiple-choice questions (score each answer completion separately) against larger models that can utilize the multiple-choice formulation. To make our evaluations reproducible, we follow the OLMES standard in prompt formatting, choice of in-context examples, probability normalization, and all other details. See Table~\ref{tab:task-details} and see \citet{olmes} for more details.

\paragraph{Generative tasks} Following the principles of OLMES \citep{olmes}, such as prompt formatting and having 5-shot curated in-context examples, we also evaluated on a suite of generative tasks, OLMES-Gen. This suite covers factual knowledge tasks (Natural Questions \citep{kwiatkowski-etal-2019-natural} and Jeopardy \citep{mosaic-jeopardy}) and tasks testing reading comprehension (SQuAD \citep{rajpurkar-etal-2016-squad}, DROP \citep{dua-etal-2019-drop}, and CoQA \citep{reddy-etal-2019-coqa}). For CoQA, the task comprises presenting a passage followed by a conversation so far, where each turn in the conversation contains a question and an answer. In this case, the previous question and answer pairs serve to guide the model in terms of the output format, and we do not include additional few-shot examples. For all other tasks, we follow OLMES in using 5-shot curated in-context examples. As the list of gold answers for these tasks are often incomplete, we use F1 as the primary metric to give partial credit when models produce answers that partially match.  The task details of OLMES-Gen are summarized in Table~\ref{tab:task-details}.

\paragraph{Held-out tasks} We also evaluate on a held-out suite of tasks that were not used when making decisions during model development. 
This suite includes advanced admission and qualification exams (AGIEval English\footnote{Specifically these 8 tasks: aqua-rat, logiqa-en, lsat-ar, lsat-lr, lsat-rc, sat-en, sat-math, gaokao-english} \citep{zhong-etal-2024-agieval}), tasks believed to be challenging to LMs (BigBenchHard, BBH; \citealp{suzgun2022challengingbigbenchtaskschainofthought}), math reasoning (GSM8K; \citealp{cobbe2021trainingverifierssolvemath}), a more challenging and reasoning-focused extension of MMLU (MMLU Pro; \citealp{wang2024mmlu}), and an unseen factual knowledge task (TriviaQA; \citealp{joshi-etal-2017-triviaqa}). We use existing in-context examples where available - for GSM8K, we use the 8-shot CoT examples from \citet{wei2023chainofthoughtpromptingelicitsreasoning}; for BBH we use the 3-shot CoT prompts from the original dataset; in evaluating MMLU-Pro, we used 5-shot examples from the original dataset. We use a 1-shot (with passage context, no CoT) prompt for AGIEval English, and a manually curated 5-shot examples from the train set for TriviaQA. Note that for the case of GSM8K, we never evaluated our models on the entire test set during the development stage, instead we use 200 examples to inform choices during development (e.g., choices of annealing mixtures); in Section \ref{sec:diveAnnealing} we refer to this 200-example subset as GSM*. 

We make all implementations publicly available at \href{https://github.com/allenai/olmes}{\path{github.com/allenai/olmes}}.

\subsection{Instruct Model Eval}
\label{app:eval-post}

\paragraph{Instruct tasks} 
We perform instruct model evaluation based on existing practices in current literature using the OLMES benchmark suite \citep{olmes} using the configuration reported in \citet{lambert2024tulu3}. 

See Table~\ref{tab:instruct-task-details} for a list of instruct tasks along with their configurations. 
These tasks include chat variations of our held-out tasks (GSM8k; \citealp{cobbe2021trainingverifierssolvemath}, BBH; \citealp{suzgun2022challengingbigbenchtaskschainofthought}), additional long-tail knowledge (PopQA; \citealp{mallen2023llm_memorization}), misconception (TruthfulQA; \citealp{lin2021truthfulqa}) and instruction-following tasks (IFEval; \citealp{zhou2023instructionfollowingevaluationlargelanguage}, AlpacaEval 2; \citealp{dubois2024length}). 
For our MMLU instruct evaluation, we use the CoT version from \citet{lambert2024tulu3} using their prompt asking the model to ``summarize'' its reasoning before answering the question. 
We evaluate Python code completion (HumanEval; \citealp{chen2021codex}, HumanEval+; \citealp{evalplus}) and competition MATH~\citep{hendrycksmath2021} with the same setup and answer extraction in OLMES.

\begin{table}[h]
\centering
\small
\begin{small}
\begin{tabular}{{L{1.7cm} L{2.5cm} C{1cm} C{1cm} C{1cm} C{1.5cm} C{1.5cm}}}
\toprule
\textbf{Category} & \textbf{Task} & \textbf{CoT} & \textbf{\# shots} & \textbf{Chat} & \textbf{Multiturn ICL} & \textbf{Metric} \\\midrule
\multicolumn{7}{c}{\textbf{\textit{Instruct tasks}}} \\
\rowcolor{ai2offwhite}\cellcolor{white}Knowledge Recall & MMLU & \cmark & 0 & \cmark & \xmark & EM \\
\rowcolor{ai2offwhite} \cellcolor{white} & PopQA & \xmark & 15 & \cmark & \cmark & EM \\
\rowcolor{ai2offwhite} \cellcolor{white} & TruthfulQA & \xmark & 6 & \cmark & \xmark & MC2 \\
\cellcolor{white}Reasoning & BigBenchHard & \cmark & 3 & \cmark & \cmark & EM \\
& DROP & \xmark & 3 & \xmark & N/A & F1 \\
\rowcolor{ai2offwhite}\cellcolor{white}Math & GSM8K & \cmark & 8 & \cmark & \cmark & EM \\
\rowcolor{ai2offwhite}\cellcolor{white} & MATH & \cmark & 4 & \cmark & \cmark & Flex EM \\
\cellcolor{white}Coding & HumanEval & \xmark & 0 & \cmark & N/A & Pass@10 \\
& HumanEval+ & \xmark & 0 & \cmark & N/A & Pass@10 \\
\rowcolor{ai2offwhite}\cellcolor{white}Instruction Following & IFEval & \xmark & 0 & \cmark & N/A & Pass@1 (prompt; loose) \\
& AlpacaEval 2 & \xmark & 0 & \cmark & N/A & LC Winrate \\
\rowcolor{ai2offwhite}\cellcolor{white}Safety & T\"ulu 3 Safety & \xmark & 0 & \cmark & N/A & Average$^*$ \\
\bottomrule
\end{tabular}
\end{small}
\vspace{3pt}
\caption{Details of OLMES benchmarks used for to evaluate \olmotooinstruct. \textbf{CoT} are evaluations run with chain of thought prompting~\citep{wei2022chain}.
\textbf{\#Shots} is the number of in-context examples in the evaluation template.
\textbf{Chat} refers to whether we use a chat template while prompting the model.
\textbf{Multiturn ICL} refers to a setting where we present each in-context example as a separate turn in a conversation (applicable only when a chat template is used and \# Shots is not 0). 
$^*$Average over multiple sub-evaluations
}
\label{tab:instruct-task-details}
\end{table}

\section{\olmotoo 1B}
\label{app:1b}

While the goal of this work is to develop development recipes for our target 7B, 13B and 32B sizes, often it is useful to perform experimentation at the 1B model size. We define \olmotoo 1B similar to \olmotoo 7B, but with the following departures:
\begin{itemize}
    \item \textbf{Layers:} 16 instead of 32
    \item \textbf{Hidden Size ($d_{model}$: } 2048 instead of 4096
    \item \textbf{Attention Heads (Q/KV): } 16/16 (MHA) instead of 32/32 (MHA)
    \item \textbf{Batch Size: } 512 instead of 1024
    \item \textbf{Peak LR: } $4.0 \cdot 10\text{E}{-4}$ instead of $3.0 \cdot 10\text{E}{-4}$
\end{itemize}

\subsection{Difficulties with \olmotoo 1B}
\label{app:1b-results}

We developed our \olmotoo recipe developed using the \olmotoo 1B model (Appendix~\ref{app:1b}) and have found findings to generalize well to the 7B, 13B and 32B scales, as seen by our competitive results in Table~\ref{tab:evals_overview}.
Yet, we have found scaling the number of training tokens for \olmotoo 1B to be difficult.

\paragraph{Training} We pretrain \olmotoo 1B to 4 trillion tokens on \olmomix and perform a single 50B token anneal on \dolminos. 
Similar to \olmotoo 7B, we use 2000 steps of warmup, set the schedule to 5 trillion tokens but truncate at the 4 trillion mark. We use a higher peak learning rate of $4.0 \cdot 10\text{E}{-4}$.

\paragraph{Base Results} Table~\ref{tab:1b_results} presents experimental results on our main base model evaluation suite. We find that while \olmotoo remains competitive with other similarly-sized models like SmolLM 2, it lags behind the smaller Gemma 2 and Qwen 2.5 base models.

\begin{table}[h]
\setlength\tabcolsep{2pt} 
\renewcommand{\arraystretch}{1.1}
\begin{center}
\begin{small}
\begin{tabular}{lcc|cccccc|cccc}
\toprule
    &&\multicolumn{7}{c}{\quad \quad \quad \quad \textbf{\texttt{Dev Benchmarks}}} & \multicolumn{4}{c}{\textbf{\texttt{Held-out Evals}}} \\
    {\textbf{Model}} &
    {\textbf{Avg}} &
    {\textbf{FLOPs}} & 
    {\textbf{MMLU}} & 
    {$\textbf{ARC$_C$}$} & 
    {\textbf{HS}} & 
    {\textbf{WG}} & 
    {\textbf{NQ}} & 
    {\textbf{DROP}} & 
    {\textbf{AGI}} & 
    {\textbf{GSM}} & 
    {\textbf{MMLU$_P$}} &
    {\textbf{TQA}} 
    \\
\midrule
\rowcolor{ai2offwhite}
\multicolumn{13}{c}{\textbf{Open-weights models 1-2B Parameters}} \\
Qwen 2.5 1.5B & 51.5 & 1.7 & 61.4 & 77.3 & 67.0 & 65.4 & 17.7 & 36.4 & 47.9 & 63.2 & 29.9 & 49.1 \\
Gemma 2 2B & 47.9 & 0.2 & 53.1 & 67.4 & 74.4 & 70.8 & 24.1 & 36.9 & 38.4 & 26.8 & 22.2 & 65.2 \\
\rowcolor{ai2offwhite}
\multicolumn{13}{c}{\textbf{Fully-open models}} \\
SmolLM 2 1.7B & 44.7 & 1.1 & 50.9 & 62.0 & 73.3 & 66.9 & 19.1 & 26.5 & 35.3 & 30.3 & 22.0 & 60.6 \\
\rowcolor{ai2lightpink} \olmotoo 1B & 43.7 & 0.4 & 44.3 & 51.3 & 69.5 & 66.5 & 20.8 & 34.0 & 36.3 & 43.8 & 16.1 & 54.7 \\
\bottomrule
\end{tabular}
\end{small}
  \caption{
  \olmotoo 1B vs.~comparable models (size, architecture) with known pretraining FLOPs (relative to $10\text{E}{23}$).
  }
\label{tab:1b_results}
\end{center}
\end{table}

\paragraph{Analysis} We postulate that our \olmotoo 1B may struggle with pretraining token efficiency due to model capacity.
\olmotoo is smaller than the smallest variants of other competitive model families like Qwen 2.5 or Gemma 2. 
We hypothesize that below a certain model size, the optimal pretraining recipe may require the inclusion of task-specific data, such as that seen in supervised fine-tuning (SFT) to achieve non-random performance over more challenging tasks in our evaluation suite.
Better performance could also be achieved by distilling from a more powerful model, a strategy used by the smaller Gemma 2 models.

For example, Table~\ref{tab:evals-pre-post-anneal} shows the benefit of \dolminos is higher with smaller base models:
+37.0\% for the 1B model, +18.7\% for the 7B model, +15.9\% for the 13B model, and +12.3\% for the 32B model.
These results also show that \olmotoo 1B with only Stage 1 pretraining struggles to break out of random performance for multiple-choice formatted tasks (25\% for MMLU and ARC Challenge, 10\% for MMLU Pro). 

As further evidence of this, Table~\ref{tab:instruct_results_1b} shows that applying our same \olmotooinstruct post-training recipe to \olmotoo 1B results in \olmotooinstruct 1B with highly competitive performance to even Qwen 2.5 and even Gemma 3.

\begin{table}[t]
\centering
\setlength\tabcolsep{3pt}
{\small
\begin{tabular}{lccccccccccc}
\toprule
    \textbf{Model} & 
    \textbf{Avg} & 
    \textbf{AE2} & 
    \textbf{BBH} & 
    \textbf{DROP} & 
    \textbf{GSM} & 
    \textbf{IFE} & 
    \textbf{MATH} & 
    \textbf{MMLU} & 
    \textbf{Safety} & 
    \textbf{PQA} & 
    \textbf{TQA}
\\
\midrule
\rowcolor{ai2offwhite}\multicolumn{12}{c}{\textbf{Open weights models 1–2B Parameters}} \\
Gemma 3 1B & 38.3 & 20.4 & 39.4 & 25.1 & 35.0 & 60.6 & 40.3 & 38.9 & 70.2 & 9.6 & 43.8 \\
Llama 3.2 1B & 39.3 & 10.1 & 40.2 & 32.2 & 45.4 & 54.0 & 21.6 & 46.7 & 87.2 & 13.8 & 41.5 \\
Qwen 2.5 1.5B & 41.7  & 7.4 & 45.8 & 13.4 & 66.2 & 44.2 & 40.6 & 59.7 & 77.6 & 15.5 & 46.5 \\
\rowcolor{ai2offwhite}
\multicolumn{12}{c}{\textbf{Fully-open models}} \\
SmolLM2 1.7B & 34.2  & 5.8 & 39.8 & 30.9 & 45.3 & 51.6 & 20.3 & 34.3 & 52.4 & 16.4 & 45.3 \\
\rowcolor{ai2lightpink} \olmotoo 1B     & 42.7    & 9.1       & 35.0   & 34.6   & 68.3   & 70.1   & 20.7   & 40.0   & 87.6        & 12.9   & 48.7 \\
\bottomrule
\end{tabular}
}
\caption{
\olmotooinstruct 1B's performance vs open-weights models of comparable size.
}
\label{tab:instruct_results_1b}
\end{table}

\clearpage
\section{Additional Instruct Details}

\subsection{Additional Hyperparameters}
All of the models used to generate preference data for \olmotooinstruct are listed in Table~\ref{tab:pref-data}.
The prompt sources for the preference datasets are listed in Table~\ref{tab:prompt_sources} -- for more information on their contents, refer to~\citet{lambert2024tulu3}.
The hyperparameters used to train the reward models for RLVR value network initialization are shown in Table~\ref{tab:rm-hypers}.

\subsection{Additional RLVR Learning Curves}
\label{appendix:rlvr-13b}
\begin{figure}[t]
    \centering
    \begin{minipage}{0.98\textwidth}
        \centering
        \includegraphics[width=\linewidth]{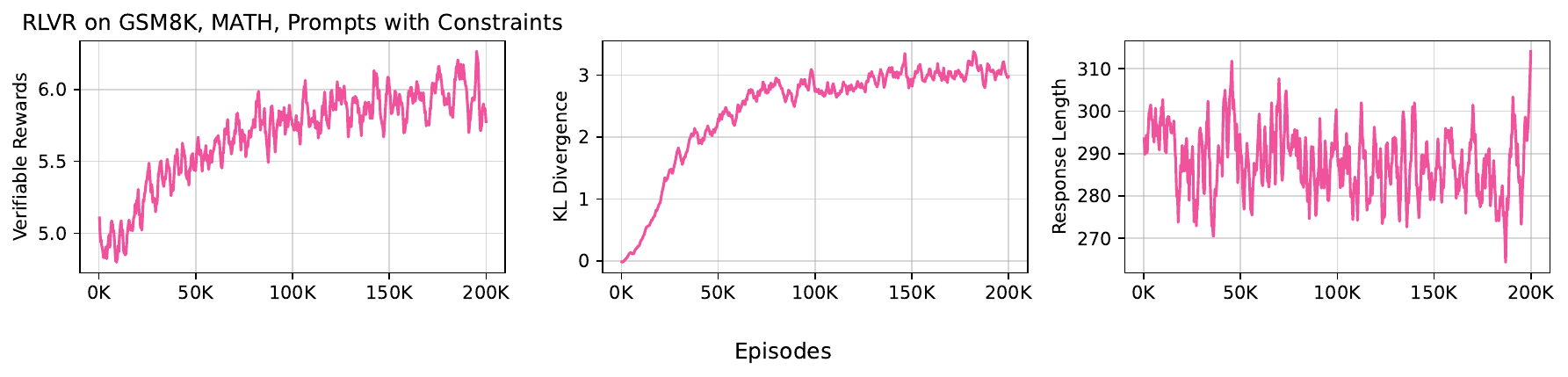}
    \end{minipage}
    \hfill
    \begin{minipage}{0.98\textwidth}
        \centering
        \includegraphics[width=\linewidth]{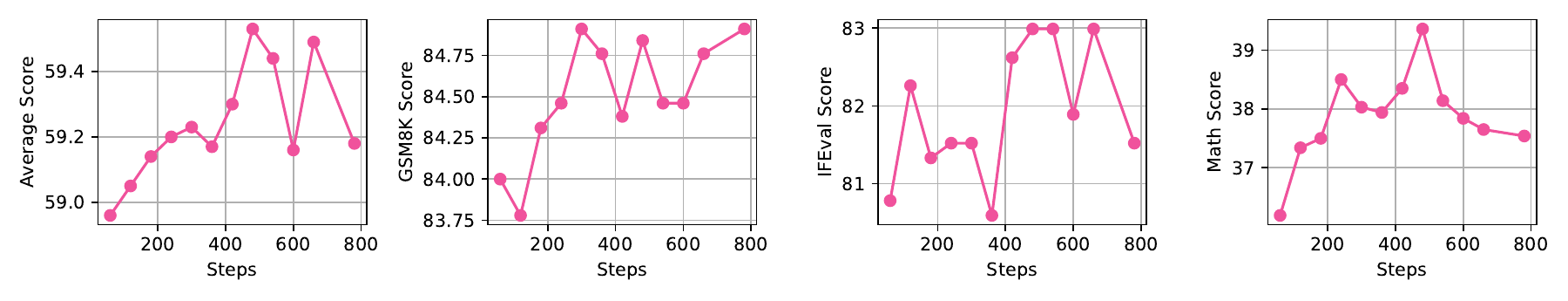}
    \end{minipage}
    {\cblock{240}{82}{156}} OLMo-2-1124-13B-RLVR1
    \caption{The top row shows the training curves of OLMo-2-1124-13B-RLVR1 showing verifiable rewards, KL divergence, and response lengths. The bottom row shows the corresponding downstream evaluations and the average scores across our evaluation suites.}
    \label{fig:the-rl-chart-13b2-1}
\end{figure}

\begin{figure}[t]
    \centering
    \begin{minipage}{0.98\textwidth}
        \centering
        \includegraphics[width=\linewidth]{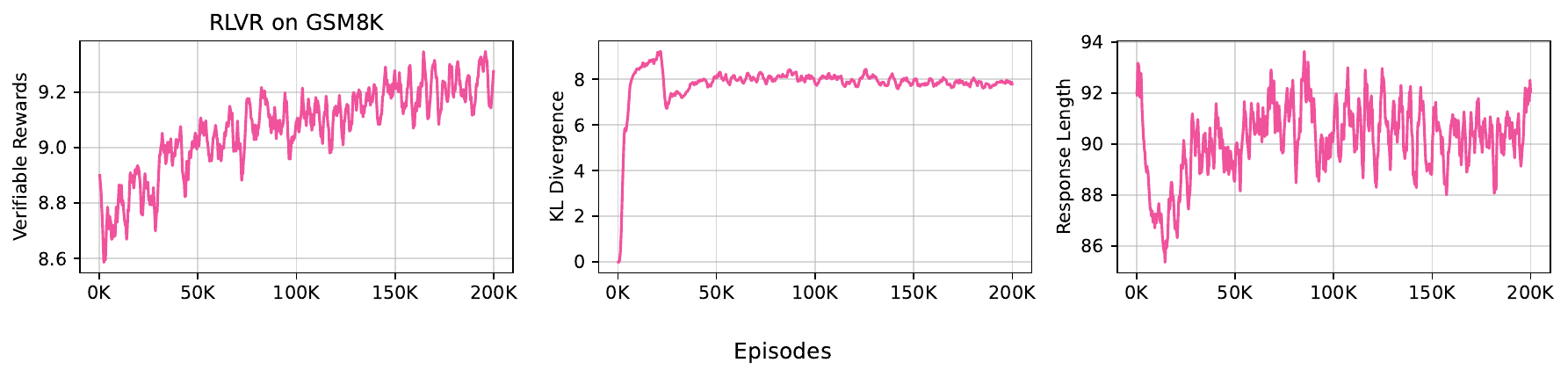}
    \end{minipage}
    \hfill
    \begin{minipage}{0.98\textwidth}
        \centering
        \includegraphics[width=\linewidth]{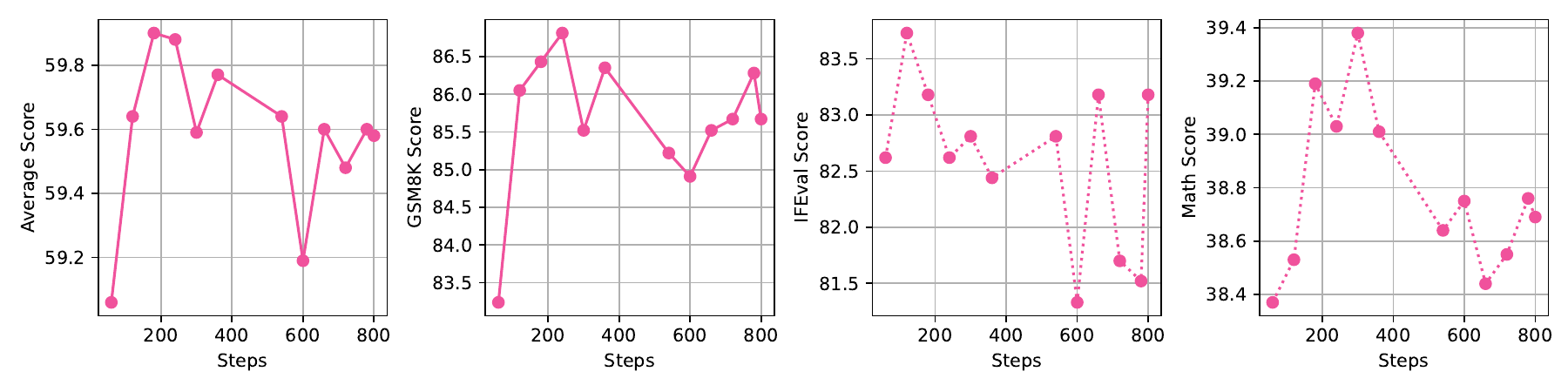}
    \end{minipage}
    {\cblock{240}{82}{156}} OLMo-2-1124-13B-RLVR2
    \caption{The top row shows the training curves of OLMo-2-1124-13B-RLVR2 showing verifiable rewards, KL divergence, and response lengths. The solid lines in the bottom row show the corresponding downstream evaluation and the average scores across our evaluation suites. }
    \label{fig:the-rl-chart-13b2-2}
\end{figure}

\begin{figure}[t]
    \centering
    \begin{minipage}{0.98\textwidth}
        \centering
        \includegraphics[width=\linewidth]{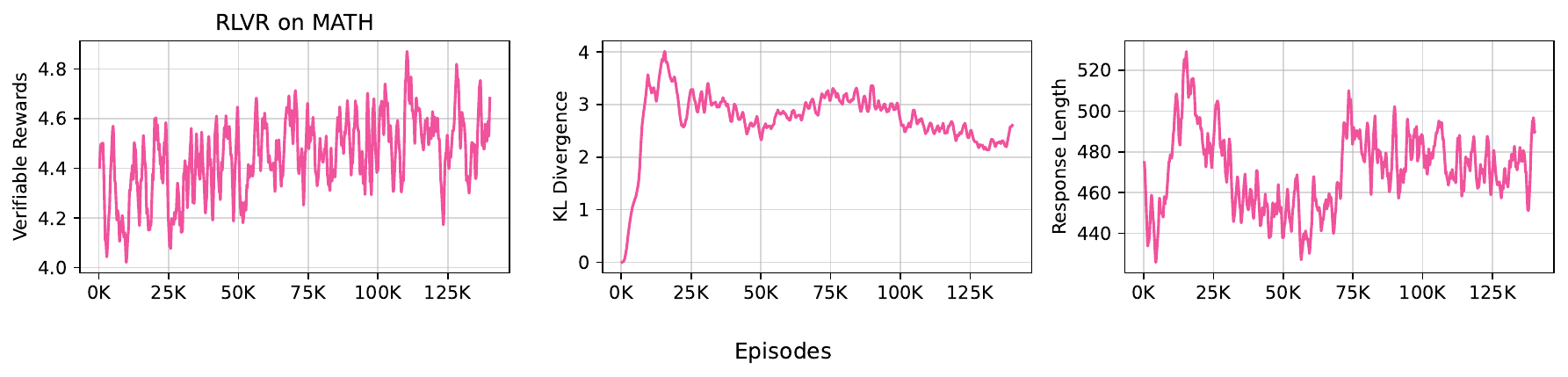}
    \end{minipage}
    \hfill
    \begin{minipage}{0.98\textwidth}
        \centering
        \includegraphics[width=\linewidth]{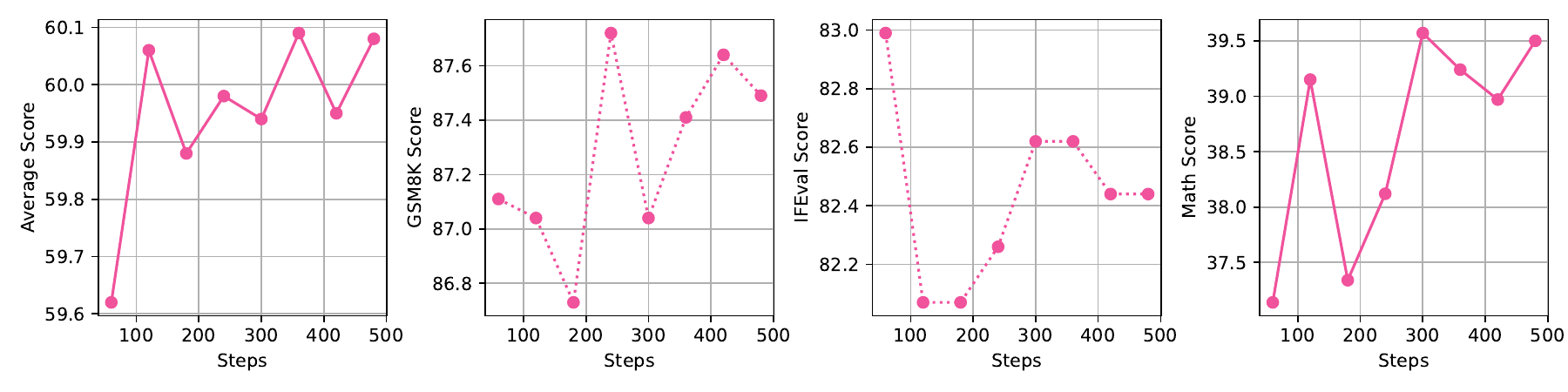}
    \end{minipage}
    {\cblock{240}{82}{156}} OLMo-2-1124-13B-Instruct
    \caption{The top row shows the training curves of OLMo-2-1124-13B-Instruct showing verifiable rewards, KL divergence, and response lengths. The solid lines in the bottom row show the corresponding downstream evaluation and the average scores across our evaluation suites. }
    \label{fig:the-rl-chart-13b2-3}
\end{figure}

The additional 13B RLVR learning curves of can be found at Figure~\ref{fig:the-rl-chart-13b2-1}, Figure~\ref{fig:the-rl-chart-13b2-2}, and Figure~\ref{fig:the-rl-chart-13b2-3}.

\subsection{\olmotooinstruct Preview Models}
We made an initial release\footnote{\url{https://allenai.org/blog/olmo2}} prior to this report. However, soon after the release, a tokenizer issue came to our attention: \textbf{ Our base model’s pre-tokenization logic differs from our instruct model’s tokenizer.}

\begin{figure}[h]
    \centering
    \begin{minipage}{0.98\textwidth}
        \centering
        \includegraphics[width=\linewidth]{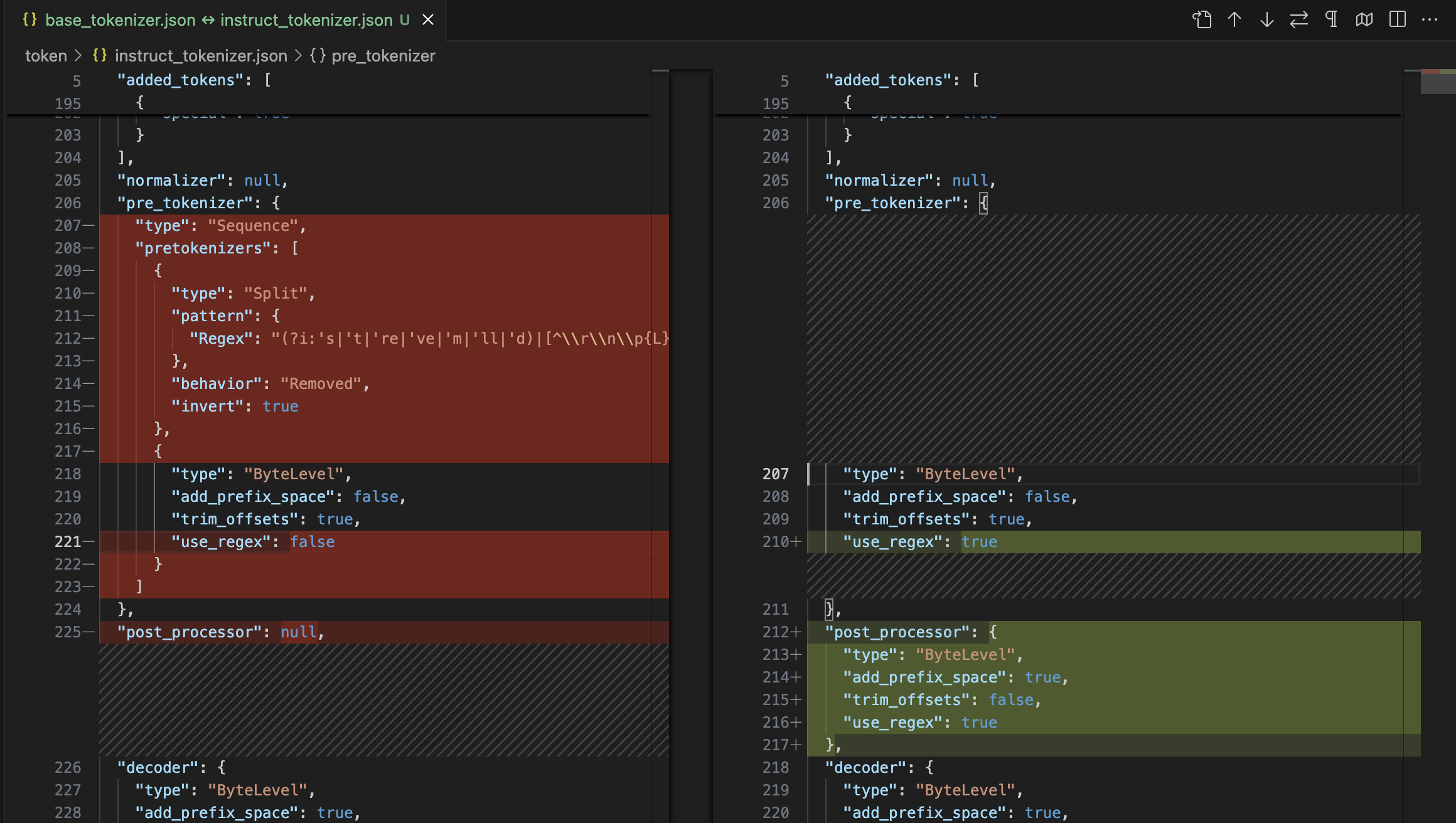}
    \end{minipage}
    \caption{The file diff between the \olmotooinstruct and \olmotooinstruct Preview's \texttt{tokenizer.json}: the pre-tokenization logic is lost during \olmotooinstruct Preview's training, so we have decided to re-train \olmotooinstruct models.}
    \label{fig:borked}
\end{figure}

Specifically, the OLMo-2 base models utilized the \texttt{GPT2Tokenizer} tokenizer class, with custom pre-tokenization logic (e.g., on splitting or truncating sequences), which is lost during the instruct model’s training. Figure~\ref{fig:borked} shows the filediff between the base model’s \texttt{tokenizer.json} and the instruct model’s \texttt{tokenizer.json}.

Because of this, we have decided to retrain our \olmotooinstruct models to be consistent with our base models and mark the existing post-trained models as \emph{preview} models. 

\begin{figure}[t]
    \centering
    \begin{minipage}{0.98\textwidth}
        \centering
        \includegraphics[width=\linewidth]{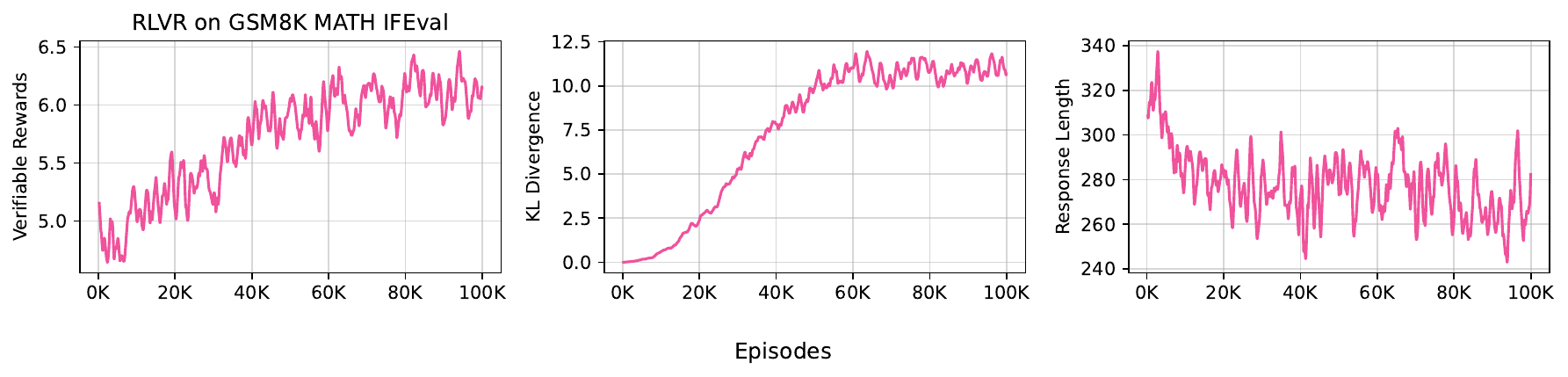}
    \end{minipage}
    \hfill
    \begin{minipage}{0.98\textwidth}
        \centering
        \includegraphics[width=\linewidth]{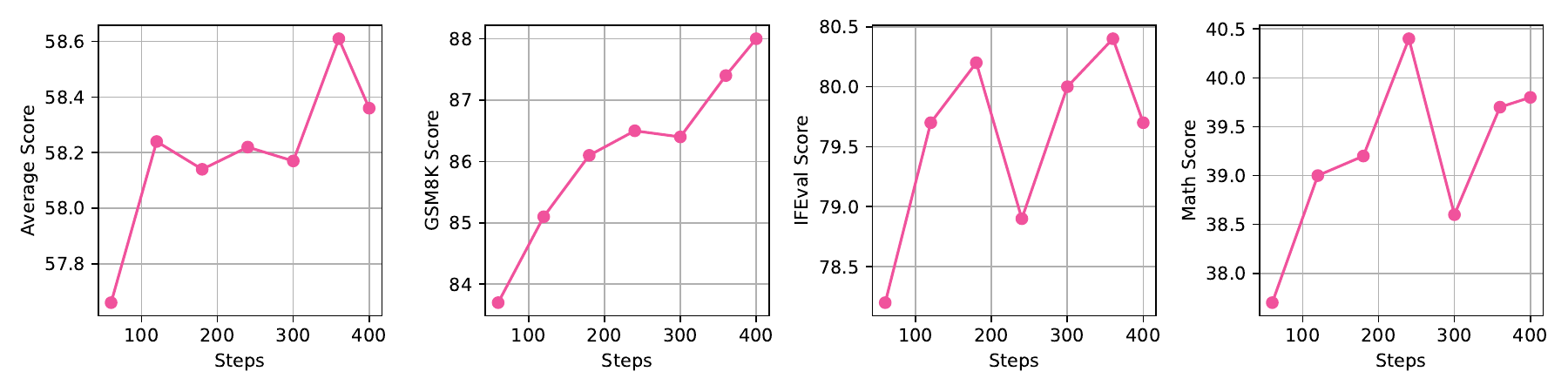}
    \end{minipage}
    {\cblock{240}{82}{156}} OLMo-2-1124-13B-Instruct-Preview 
    \caption{The OLMo-2-1124-13B-Instruct-Preview results. The top row shows the training curves of OLMo-2-1124-7B-Instruct on verifiable rewards, KL divergence, and response lengths. In the bottom row, the y-axes show the average scores across our evaluation suites and GSM8K scores. Overall, RLVR increases both training rewards and evaluation scores.}
    \label{fig:the-rl-chart-13b-legacy}
\end{figure}

\begin{figure}[t]
    \centering
    \begin{minipage}{0.98\textwidth}
        \centering
        \includegraphics[width=\linewidth]{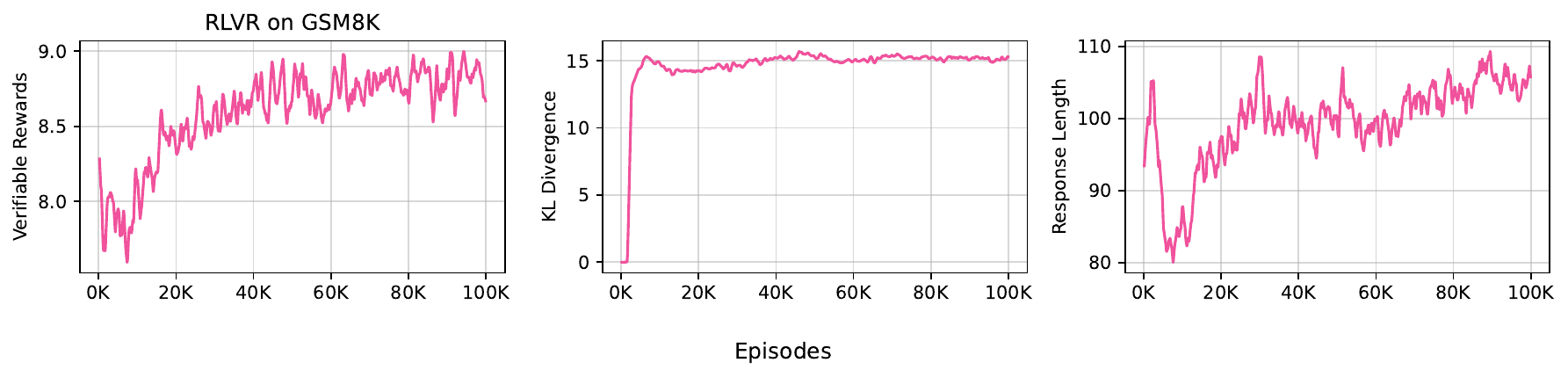}
    \end{minipage}
    \hfill
    \begin{minipage}{0.98\textwidth}
        \centering
        \includegraphics[width=\linewidth]{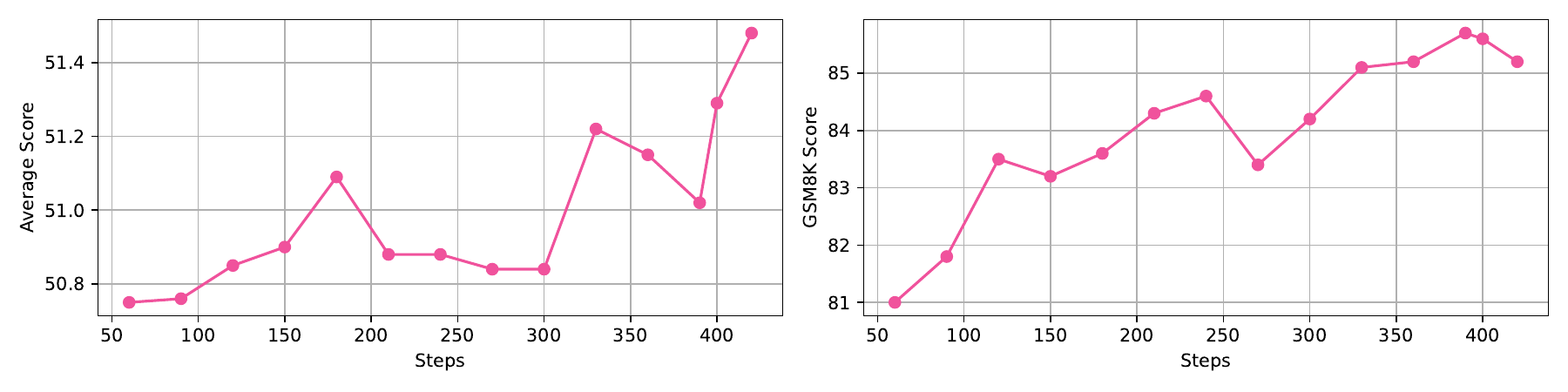}
    \end{minipage}
    {\cblock{240}{82}{156}} OLMo-2-1124-7B-Instruct-Preview
    \caption{The top row shows the training curves of OLMo-2-1124-13B-Instruct-Preview on verifiable rewards, KL divergence, and response lengths. In the bottom row, the y-axes show the average scores across our evaluation suites and GSM8K, IFEval, and MATH Flex scores, respectively. Overall, we found RLVR increases not only the training rewards of our 13B models but also the downstream evaluations such as GSM8K.}
    \label{fig:the-rl-chart-7b-legacy}
\end{figure}

Nevertheless, the \olmotooinstruct Preview learning curves can be found at Figure~\ref{fig:the-rl-chart-13b-legacy} and Figure~\ref{fig:the-rl-chart-7b-legacy}.

\begin{table}[t]
\centering
\setlength\tabcolsep{3pt}
{\small
\begin{tabular}{l c c c c c c}
\hline
 &  & AGI Eval & DeepMind  &  &  & \\ 
Model & Average &  English &  Math & GPQA & IFEval OOD & MMLU Pro \\ \hline
OLMo 2 32B Instruct & 44.9 & 68.3 & 34.7 & 35.9 & 33.1 & 52.7 \\
OLMo 2 32B DPO & 43.8 & 68.6 & 34.5 & 35.7 & 26.8 & 53.3 \\
OLMo 2 32B SFT & 39.3 & 63.9 & 33.4 & 32.6 & 20.4 & 46.3 \\
OLMo 2 1124 13B Inst. & 35.2 & 60.5 & 26.8 & 28.8 & 18.7 & 41.4 \\
OLMo 2 1124 13B DPO & 35.5 & 60.1 & 25.4 & 32.1 & 18.0 & 41.8 \\
OLMo 2 1124 13B SFT & 33.0 & 56.0 & 27.1 & 27.0 & 16.6 & 38.2 \\
OLMo 2 1124 7B Inst. & 32.2 & 57.2 & 19.1 & 30.1 & 18.7 & 36.0 \\
OLMo 2 1124 7B DPO & 31.8 & 56.7 & 17.7 & 30.6 & 17.3 & 36.6 \\
OLMo 2 1124 7B SFT & 29.8 & 52.7 & 19.0 & 27.7 & 16.2 & 33.2 \\
OLMo 7B 0724 Inst. & 22.9 & 43.6 & 5.8 & 27.9 & 14.4 & 22.9 \\
OLMoE 1B 7B 0924 Inst. & 20.5 & 39.1 & 4.2 & 27.5 & 11.3 & 20.6 \\ \hline
\end{tabular}
}
\caption{Evaluation results for OLMo Instruct models on the unseen suite from~\citet{lambert2024tulu3}.
Note that IFEval OOD has been improved via minor bug fixes for OLMo 2 7B and 13B, so the numbers are not exactly comparable to those in~\citet{lambert2024tulu3}. }
\label{tab:instruct-unseen}
\end{table}

\clearpage
\section{Additional Hyperparameters}
The models used for the on-policy preference data generation are listed in Table~\ref{tab:pref-data}.

\begin{table}[h!]
\centering
\begin{tabular}{ll}
\toprule
\textbf{Model Name} & \textbf{Reference} \\ \midrule
\href{https://huggingface.co/01-ai/Yi-34B-Chat}{Yi-34B-Chat} & \citep{young2024yi} \\ 
\href{https://huggingface.co/01-ai/Yi-6B-Chat}{Yi-6B-Chat} & \citep{young2024yi} \\ 
\href{https://huggingface.co/allenai/tulu-2-7b}{T\"ulu 2 7B} & \citep{ivison2023camels}\\ 
\href{https://huggingface.co/allenai/tulu-2-13b}{T\"ulu 2 13B} & \citep{ivison2023camels} \\ 
\href{https://huggingface.co/google/gemma-2-27b-it}{Google Gemma 2 27B it} & \citep{gemma2} \\ 
\href{https://huggingface.co/google/gemma-2-9b-it}{Google Gemma 2 9B it} & \citep{gemma2} \\ 
\href{https://platform.openai.com/docs/models#gpt-4o}{GPT-4o} & \citep{hurst2024gpt} \\ 
\href{https://huggingface.co/mosaicml/mpt-30b-chat}{MPT 30B Chat} & \citep{MosaicML2023Introducing} \\ 
\href{https://huggingface.co/mosaicml/mpt-7b-8k-chat}{MPT 7B 8k Chat} & \citep{MosaicML2023Introducing} \\ 
\href{https://huggingface.co/mistralai/Mistral-7B-Instruct-v0.2}{Mistral 7B Instruct v0.2} & \citep{jiang2023mistral} \\ 
\href{https://huggingface.co/mistralai/Mistral-Nemo-Instruct-2407}{Mistral Nemo Instruct 2407} & \citep{mistralnemo} \\ 
\href{https://huggingface.co/Qwen/Qwen2.5-32B-Instruct}{Qwen2.5 32B Instruct} & \citep{qwen2.5} \\ 
\href{https://huggingface.co/Qwen/Qwen2.5-14B-Instruct}{Qwen2.5 14B Instruct} & \citep{qwen2.5} \\ 
\href{https://huggingface.co/Qwen/Qwen2.5-7B-Instruct}{Qwen 2.5 7B Instruct} & \citep{qwen2.5} \\ 
\href{https://huggingface.co/tiiuae/falcon-7b-instruct}{Falcon 7B} & \citep{falcon40b} \\ 
\href{https://huggingface.co/HuggingFaceTB/SmolLM2-1.7B-Instruct}{SmolLM2 1.7B Instruct} & \citep{allal2024SmolLM2} \\
\href{https://huggingface.co/microsoft/Phi-3-mini-128k-instruct}{Phi 3 Mini 128k Instruct} & \citep{abdin2024phi} \\
\href{https://huggingface.co/microsoft/Phi-3.5-mini-instruct}{Phi 3.5 Mini Instruct} & \citep{abdin2024phi} \\
\href{https://huggingface.co/numind/NuExtract-1.5}{NuExtract-1.5} & \citep{NuExtract15} \\
\bottomrule
\end{tabular}
\vspace{3pt}
\caption{External models used to sample off-policy data in the synthetic preference pipeline. 
These are in addition to the on-policy samples from the SFT checkpoints.
\label{tab:pref-data}
}
\end{table}

\begin{table}[h!]
\centering
\begin{tabular}{@{}ll@{}}
\toprule
\textbf{Hyperparameter} & \textbf{Value} \\
\midrule
Learning Rate & $3 \cdot 10^{-6}$ \\
Gradient Norm Threshold & 1.0 \\
Learning Rate Schedule & Linear \\
Batch Size (effective) & 256 \\
Max Token Length &2,048 \\
Number of Epochs & 1 \\
\bottomrule
\end{tabular}
\vspace{3pt}
\caption{This table shows the hyperparameters used to train the reward model for RLVR value network initialization.}
\label{tab:rm-hypers}
\end{table}

\begin{table}[h!]
    \centering
    \begin{tabular}{llll}
    \toprule
    \textbf{Dataset} & \textbf{Counts} & \textbf{7B DPO} & \textbf{13B DPO} \\
    \midrule
       SFT Reused  & 117,025 & $\checkmark$ & $\checkmark$   \\
       SFT IF & 65,792 & $\checkmark$ & $\checkmark$  \\
       WildChat Unused & 84,105 & $\checkmark$ & $\checkmark$ \\
       WildChat Reused & 17,703 &  $\checkmark$ & $\checkmark$ \\
       WildChat IF & 10,794 &  & $\checkmark$ \\
       Ultrafeedback (cleaned) & 60,816 & $\checkmark$ & $\checkmark$ \\
       DaringAnteater IF & 1,618 & $\checkmark$ & $\checkmark$ \\
       T\"ulu 3 Personas IF & 19,890 & $\checkmark$ & $\checkmark$ \\
       \midrule
       \textit{Total} & 377,743 & & \\
    \bottomrule
    \end{tabular}
    \caption{Prompt sources for preference finetuning datasets.}
    \label{tab:prompt_sources}
\end{table}

\begin{figure}[h]
    \centering
    \begin{minipage}{0.98\textwidth}
        \centering
        \includegraphics[width=\linewidth]{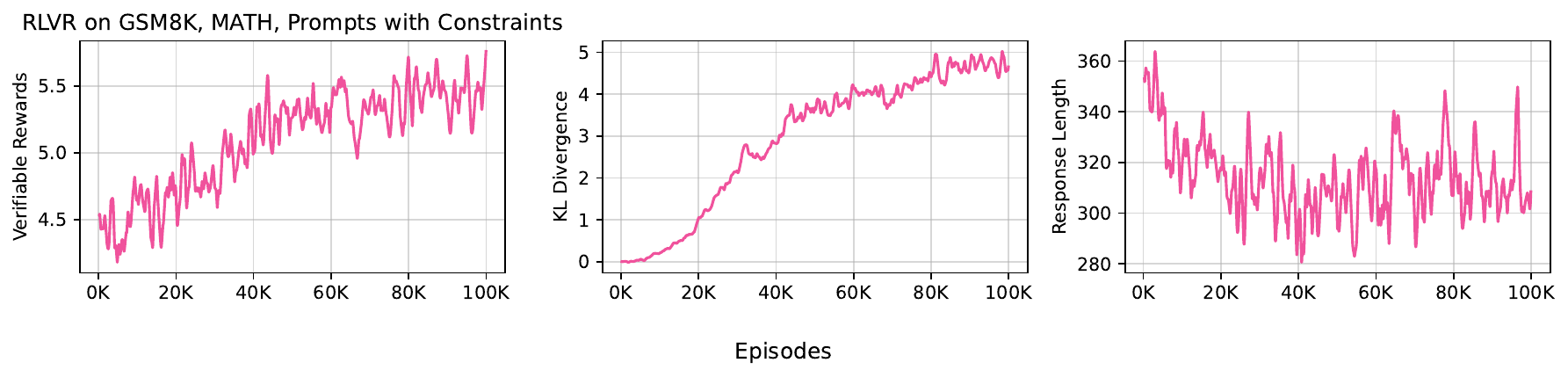}
    \end{minipage}
    \hfill
    \begin{minipage}{0.98\textwidth}
        \centering
        \includegraphics[width=\linewidth]{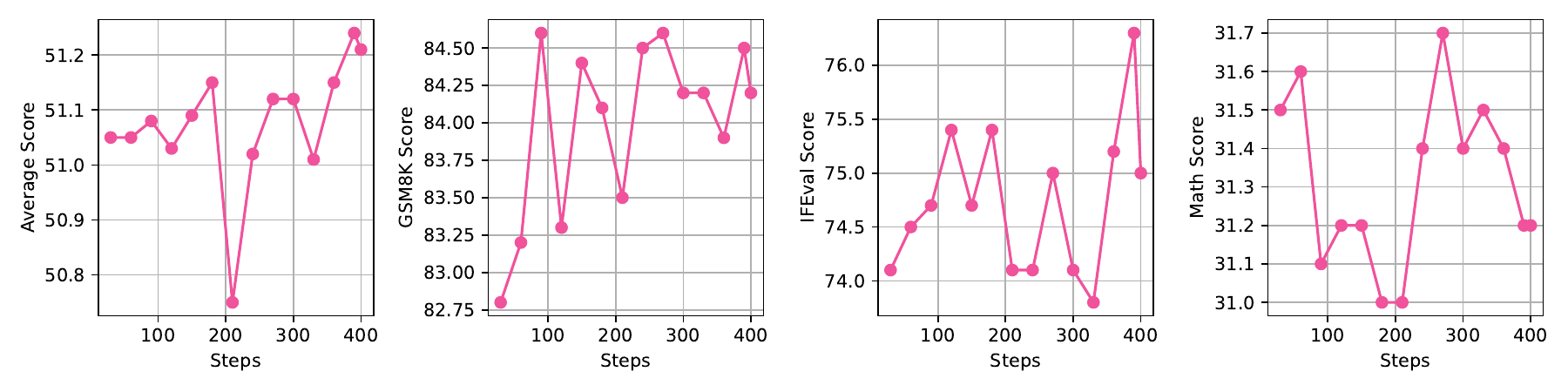}
    \end{minipage}
    {\cblock{240}{82}{156}} OLMo 2 7B Instruct Preview (alternative) 
    \caption{The top row shows the training curves of OLMo-2-1124-13B-Instruct on verifiable rewards, KL divergence, and response lengths. In the bottom row, the y-axes show the average scores across our evaluation suites and GSM8K, IFEval, and MATH Flex scores, respectively. Overall, we found RLVR increases not only the training rewards of our 13B models but also the downstream evaluations such as GSM8K.}
    \label{fig:the-rl-chart-7b-alt}
\end{figure}

\clearpage

\clearpage

\section{Annealing Data Details}
\label{app:data_details}

\begin{figure}[h]
    \centering
    \begin{tcolorbox}[colframe=black!80!white, colback=black!2!white, boxrule=0.5mm, width=\textwidth, arc=2mm, auto outer arc, title=Hard Math Problems (prompt), fonttitle=\color{white}\bfseries]
    \setstretch{1.2}
Create a math problem related to the following persona:\\\\
\texttt{\{persona\}}\\\\
Note:\\\\
1. The math problem should be challenging and involve advanced mathematical skills and knowledge. Only top talents can solve it correctly.\\
2. You should make full use of the persona description to create the math problem to ensure that the math problem is unique and specific to the persona.\\
3. Your response should always start with "Math problem:". Your response should not include a solution to the created math problem.\\
4. Your created math problem should include no more than 2 sub-problems.
    \end{tcolorbox}
    \caption{Prompt used to generate hard math word problems. \texttt{\{persona\}} are borrowed from \citet{chan2024scaling}.}
    \label{fig:persona-prompt-sft-math}
\end{figure}

\begin{figure}[h]
    \centering
    \begin{tcolorbox}[colframe=black!80!white, colback=black!2!white, boxrule=0.5mm, width=\textwidth, arc=2mm, auto outer arc, title=Hard Math Problems (response), fonttitle=\color{white}\bfseries]
    \setstretch{1.2}
    Provide solution to the given math problem.\\\\
    Problem: \texttt{\{generated\_math\_problem\}}\\\\
    Note: Provide your solution step-by-step, and end your solution in a new line in the following format: \\ Final Answer: The final answer is \$final\_answer\$. I hope it is correct.
    \end{tcolorbox}
    \caption{Prompt used to generate solutions for hard math word problems.}
    \label{fig:persona-prompt-sft-math-res}
\end{figure}

\end{document}